\documentclass[10pt,twocolumn,letterpaper]{article}

\usepackage[pagenumbers]{cvpr} 

\usepackage{array}
\usepackage{makecell}
\usepackage{booktabs}
\usepackage{multirow}
\usepackage{amsmath,amssymb}
\usepackage{adjustbox}
\usepackage{float}

\newcolumntype{P}[1]{>{\raggedright\arraybackslash}p{#1}}

\definecolor{cvprblue}{rgb}{0.21,0.49,0.74}
\usepackage[pagebackref,breaklinks,colorlinks,allcolors=cvprblue]{hyperref}
\usepackage{multirow}

\usepackage[utf8]{inputenc}
\usepackage[T1]{fontenc}

\def\confName{CVPR}
\def\confYear{2027}

\title{Learned but Not Read: A Representation–Readout Mismatch in Parallel Scanpath Prediction}

\author{
Dongyang Guo\\
Technical University of Munich, Germany\\
Munich Center for Machine Learning (MCML), Germany\\
{\tt\small dongyang.guo@tum.de}
\and
Yiming Du\\
Ludwig-Maximilians-Universität München, Germany\\
Munich Center for Machine Learning (MCML), Germany\\
{\tt\small yiming.du@campus.lmu.de}
\and
Yasmeen Abdrabou\\
Technical University of Munich, Germany\\
Munich Center for Machine Learning (MCML), Germany\\
{\tt\small yasmeen.abdrabou@tum.de}
\and
Guorong Feng\\
Technical University of Munich, Germany\\
{\tt\small guorongfeng466@gmail.com}
\and
Enkelejda Kasneci\\
Technical University of Munich, Germany\\
Munich Center for Machine Learning (MCML), Germany\\
{\tt\small enkelejda.kasneci@tum.de}
}

\begin{document}
\maketitle
\begin{abstract}
We identify a mismatch between representation and readout in parallel scanpath models that supervise per-fixation spatial distributions but output coordinates through a separate regression head. In a controlled model family, replacing the regression output with the spatial map expectation on the same checkpoint significantly improves all six metrics on COCO-Search18 target-present, and sharpened readout still helps when the regression head reads the full spatial map. Building on this finding, we propose calibrated inference readout. It combines temperature-sharpened spatial readout (SR) with development-set termination calibration (TC). For checkpoints with a supervised spatial head, it requires no retraining, introduces no new learnable parameters, and adds no extra forward passes. After locking the readout rule and calibrating termination on each development set, SR raises ScanMatch by 0.045 at fixed termination on target-absent and AiR-D, and SR + TC improves all six metrics over native predictions. In an exploratory target-present evaluation, ScanMatch rises from 0.364 to 0.519. SR alone also improves all six metrics on Gazeformer equipped with an auxiliary spatial head. Under matched readout and termination, paired comparisons detect no additional gain from end-to-end training through the spatial map. HAT-SF, our shared-feature adaptation of the autoregressive HAT, remains more accurate on several metrics. These findings suggest that spatial supervision should be evaluated together with the readout that translates the learned representation into fixations.
\end{abstract}    
\section{Introduction}
\label{sec:Introduction}

When humans perform visual tasks, the temporal order of their fixations reflects how the task guides attention \cite{yarbus2013eye,wolfe2021guided}. Predicting such fixation sequences (i.e., scanpaths) helps understand goal-directed attentional control and supports applications such as human–computer interaction \cite{jiang2024eyeformer}. With the emergence of datasets such as COCO-Search18 \cite{chen2021coco,yang2020predicting} and AiR-D \cite{chen2020air}, task-conditioned scanpath prediction has become a quantitatively comparable problem \cite{yang2020predicting,chen2021predicting,mondal2023gazeformer,yang2022target,yang2024unifying,chen2024gazexplain,cartella2025modeling}.

Scanpath models must convert learned representations into fixation coordinates, and existing methods do so differently. Some select the next location from step-wise spatial maps~\cite{yang2020predicting,yang2022target,yang2024unifying}, some regress all coordinates in parallel~\cite{mondal2023gazeformer}, and some sample from predicted spatial distributions~\cite{chen2024gazexplain}. Because these methods also differ in architecture and training objective, existing comparisons cannot tell how much of a performance gap comes from the learned spatial representation and how much from how locations are read out from it. Even same-architecture comparisons change the output form and the training objective together~\cite{yang2024unifying}, rather than isolating readout on the same weights.

We study this question in a controlled family of parallel models that share all modules except the location pathway: Reg has only a regression head. Reg+Map adds auxiliary spatial-map supervision but still outputs coordinates through the regression head. Task-Conditioned Spatial Decoding (TCSD) removes the regression head and decodes coordinates from the spatial map during both training and inference. On COCO-Search18, replacing only the coordinate source of the same Reg+Map weights, from the regression head to the expectation over the spatial map, significantly improves all six evaluation metrics. The supervised spatial map thus contains localization information that the regression output does not fully exploit, which we call a mismatch between representation and readout.

Based on this observation, we propose calibrated inference readout, which operates only at inference and does not update model weights. Spatial location readout (SR) takes the expectation over the temperature-sharpened spatial map as the fixation coordinate, and termination calibration (TC) selects a stopping threshold on the development set so that the mean predicted length matches the human mean. To test whether decoding through the spatial map during training brings further gains, we use TCSD as a control. Under matched readout and termination, we detect no additional gain of TCSD over post-hoc readout from Reg+Map.

On COCO-Search18 target-present, SR + TC raises the ScanMatch of Reg+Map from 0.364 to 0.519 (10 seeds). We compare against HAT-SF, a shared-feature adaptation of the autoregressive HAT~\cite{yang2024unifying} retrained under the same protocol. On their three common seeds, with both stopping thresholds calibrated by the same rule, we are higher on ScanMatch, while HAT-SF remains better on KLD, CC, and NSS. Because the readout scheme was formed after observing these test results, we report them as exploratory. After locking the rules, SR + TC improves all six metrics on target-absent search~\cite{yang2022target} and AiR-D~\cite{chen2020air}. Furthermore, SR improves all six metrics on Gazeformer~\cite{mondal2023gazeformer} equipped with an auxiliary spatial head.

Our main contributions are threefold:
\begin{itemize}
    \item \textbf{Diagnosis of a representation-readout mismatch.} Through controlled same-checkpoint comparisons in a family of parallel scanpath models, we show that supervised spatial maps contain localization information left underutilized by native regression outputs.
    \item \textbf{Calibrated inference readout.} We combine temperature-sharpened spatial readout with development-set termination calibration and separate their contributions: SR helps under every tested control, while TC mainly corrects lengths of existing checkpoints. The scheme adds no learnable parameters and applies to checkpoints that already provide per-fixation spatial distributions and an adjustable stopping threshold.
    \item \textbf{Training control and transfer scope.} Under matched readout and termination protocols, we find no detectable benefit from end-to-end spatial decoding training. 
\end{itemize}
\section{Related Work}
\label{sec:related work}

\paragraph{Task-conditioned scanpath prediction.}
Scanpath prediction was initially studied mainly under free viewing~\cite{assens2018pathgan,kummerer2022deepgaze}. Datasets such as COCO-Search18~\cite{chen2021coco,yang2020predicting} and AiR-D~\cite{chen2020air} later extended the field to task-conditioned prediction. For instance, IRL~\cite{yang2020predicting} and FFM~\cite{yang2022target} address target-present and target-absent search, respectively, while Chen~\etal~\cite{chen2021predicting} address visual question answering. Subsequent work emphasizes generality. Examples include supporting unseen search targets~\cite{mondal2023gazeformer}, handling multiple fixation tasks within a single framework~\cite{yang2024unifying,cartella2025modeling}, and generating natural language explanations alongside scanpaths~\cite{chen2024gazexplain}. These methods differ in task representation, architecture, and location readout. 

\paragraph{Output forms of fixation locations.}
Existing methods use either coordinate or spatial-distribution parameterizations. Gazeformer~\cite{mondal2023gazeformer} decodes ordered fixation slots in parallel and regresses the parameters of a Gaussian over each fixation's location and duration. Spatial-distribution methods predict probabilities over candidate locations, either on a coarse grid~\cite{yang2020predicting,yang2022target,chen2021predicting}, which limits location precision, or as per-step spatial maps from which HAT~\cite{yang2024unifying} selects and GazeXplain~\cite{chen2024gazexplain} samples locations step by step. In these works, the readout is fixed together with the architecture rather than studied as a separate variable. HAT compared dense prediction and regression within one architecture and found higher sequence scores for the former, especially under target absence~\cite{yang2024unifying}. This comparison, however, changes output form and training objective together. Our models have both a supervised spatial head and a regression head, so the readout can be changed with all weights fixed.

\paragraph{Heatmaps, coordinate regression, and decoding.}
In keypoint localization, the relation between heatmaps and coordinate regression has been widely studied. Heatmap methods~\cite{tompson2014joint} take the maximum of per-location confidence. Integral regression~\cite{sun2018integral} and differentiable spatial-to-numerical transforms~\cite{nibali2018numerical} take the expectation over normalized heatmaps for end-to-end training, and likelihood-based regression narrows the gap to heatmap methods~\cite{li2021human}. DARK~\cite{zhang2020distribution} shows that the heatmap-to-coordinate decoding step itself strongly affects accuracy, and Gu~\etal~\cite{gu2021removing} analyze the bias of expectations over softmax heatmaps. Our temperature-sharpened expectation belongs to this class of operations and is not a new operator. Unlike these settings, however, our spatial map is only an auxiliary target and the native coordinates come from a separate regression head, so the question is not how to decode a heatmap more accurately but which output the location should be read from.

\paragraph{Termination and evaluation.}
Scanpath length depends on termination prediction under each model's inference rules. Gazeformer labels each parallel slot as a valid fixation or padding~\cite{mondal2023gazeformer}, and HAT predicts a step-wise termination probability~\cite{yang2024unifying}. Sequence metrics such as ScanMatch~\cite{cristino2010scanmatch}, MultiMatch~\cite{dewhurst2012depends}, and sequence edit distance, as well as saliency metrics built from predicted fixations~\cite{bylinskii2018different}, all depend on the number of predicted fixations, so termination settings can change model comparisons. We therefore calibrate the termination threshold separately on the development set and report its effect apart from location readout.
\section{Method}
\label{sec:method}

\begin{figure*}[t]
    \centering
    \includegraphics[width=\textwidth]{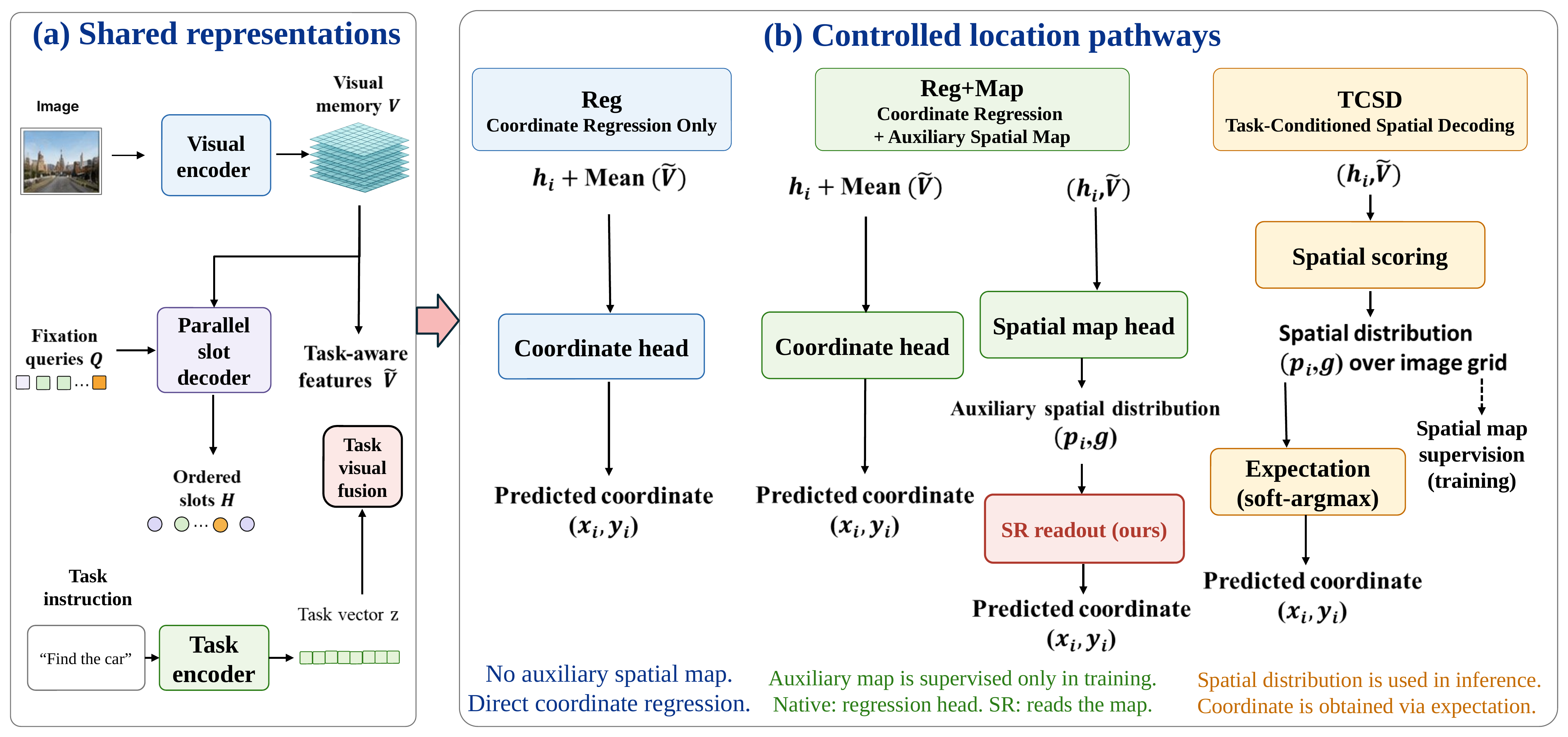}
    \caption{Overall framework. (a) Shared visual encoder, task encoding and fusion, and parallel slot decoder. (b) Controlled location pathways: Reg regresses coordinates; Reg+Map adds an auxiliary spatial map supervised only in training, and SR (red box, ours) reads coordinates from it at inference without updating weights; TCSD takes the expectation over its spatial map in both training and inference.}
    \label{fig:method}
\end{figure*}

Our method, calibrated inference readout, applies two components to a trained parallel scanpath model: spatial location readout (SR) decodes fixation locations from the learned spatial distribution (Section~\ref{sec:sr}), and termination calibration (TC) sets the stopping threshold on the development set (Section~\ref{sec:tc}). Neither adds learnable parameters or forward passes. Figure~\ref{fig:method} shows the controlled model family, which shares a visual encoder, a task encoder, and a parallel slot decoder and differs only in the location pathway (Section~\ref{sec:family}).

\subsection{Problem Setting and Shared Architecture}
\label{sec:arch}

Given an image $I$ and a task condition $c$ (a search target or a question), the model predicts an ordered scanpath $S = \{(\mathbf{x}_i, d_i)\}_{i=1}^{T}$, where $\mathbf{x}_i \in [0,1]^2$ is the fixation location normalized by image size, $d_i$ is its duration, and the length $T$ is determined by termination prediction.

\textbf{Visual and task encoding.} The input is a set of pre-extracted features $F \in \mathbb{R}^{G \times D_{\mathrm{img}}}$ on an $H_g \times W_g$ grid ($G = H_g W_g$). A visual Transformer yields $V = [v_1, \dots, v_G] \in \mathbb{R}^{G \times D}$ with hidden dimension $D$. The task condition is encoded by a frozen RoBERTa~\cite{liu2019roberta}, passed through a trainable pooler and projection to obtain $z$, and fused with each visual feature: $\tilde{v}_g = F_{\mathrm{fuse}}([v_g; z])$, where $F_{\mathrm{fuse}}$ is a linear layer with a nonlinearity.

\textbf{Parallel slots.} $L$ learnable ordered queries are decoded by a Transformer decoder into slot states $[h_1, \dots, h_L]$, $h_i \in \mathbb{R}^{D}$, in a single forward pass. Slot $i$ predicts the location, duration, and termination probability $\pi_i$ of the $i$-th fixation. We use $L = 16$ and a $24 \times 32$ grid ($G = 768$). The decoder cross-attends to $V$; task information enters the location pathway only through the fused features, identically in the three core configurations. Visual encoding, task fusion, and query-based parallel decoding follow Gazeformer and GazeXplain~\cite{mondal2023gazeformer,chen2024gazexplain} (Figure~\ref{fig:method}a); the location pathways below are constructed for controlled comparison.

\subsection{Controlled Model Family and Training Objectives}
\label{sec:family}

\textbf{Coordinate regression.} Reg and Reg+Map add the mean fused feature $\bar{v} = \frac{1}{G}\sum_g \tilde{v}_g$ to the slot state and regress $\hat{\mathbf{x}}^{\mathrm{reg}}_i = \mathrm{sigmoid}\big(f_x(h_i + \bar{v})\big)$
with a linear projection $f_x$. The slot states $h_i$ are computed from visual features before task fusion, so task conditioning enters the regression head only through $\bar{v}$. For a fixed image, changing the task therefore adds the same offset to the pre-sigmoid coordinate logits of every slot.

\textbf{Spatial distribution.} Each slot state is matched with every grid feature by a scaled dot product of linear projections $f_q$ and $f_k$, giving a distribution over the grid:
\begin{equation}
    s_{i,g} = \frac{f_q(h_i)^\top f_k(\tilde{v}_g)}{\sqrt{D}}, \qquad p_{i,g} = \frac{\exp(s_{i,g})}{\sum_{j=1}^{G} \exp(s_{i,j})}.
\end{equation}
TCSD has no coordinate head and takes the expectation over the normalized grid-cell centers $r_g \in [0,1]^2$: $\hat{\mathbf{x}}^{\mathrm{TCSD}}_i = \sum_{g} p_{i,g}\, r_g$.

\textbf{Training objectives.} Let $m_i$ indicate whether slot $i$ is a valid fixation and $N_{\mathrm{fix}} = \sum_i m_i$. The coordinate loss is Smooth-L1 ($\beta = 1$), averaged over the two coordinate dimensions, to the center of the grid cell $g_i^*$ containing the $i$-th ground-truth fixation; the map loss is the cross-entropy to $q_i$, an isotropic Gaussian ($\sigma = 1$ cell) centered at that fixation and normalized over the grid:
\begin{align}
    \mathcal{L}_{\mathrm{coord}} &= \frac{1}{N_{\mathrm{fix}}} \sum_{i=1}^{L} m_i\, \ell_{\mathrm{SL1}}\big(\hat{\mathbf{x}}_i, r_{g_i^*}\big), \\
    \mathcal{L}_{\mathrm{map}} &= -\frac{1}{N_{\mathrm{fix}}} \sum_{i=1}^{L} m_i \sum_{g=1}^{G} q_{i,g} \log p_{i,g},
\end{align}
where $\hat{\mathbf{x}}_i$ is the coordinate output of the model ($\hat{\mathbf{x}}^{\mathrm{reg}}_i$ for Reg and Reg+Map, $\hat{\mathbf{x}}^{\mathrm{TCSD}}_i$ for TCSD). The total loss is $\mathcal{L} = \mathcal{L}_{\mathrm{stop}} + \lambda_d \mathcal{L}_{\mathrm{dur}} + \alpha_m \mathcal{L}_{\mathrm{map}} + \lambda_x \mathcal{L}_{\mathrm{coord}}$ with $\lambda_d = 1$; Reg uses $(\alpha_m, \lambda_x) = (0, 1)$, and Reg+Map and TCSD use $(1, 1)$. $\mathcal{L}_{\mathrm{dur}}$ is the negative log-likelihood of the duration under a predicted log-normal distribution, whose median is used at inference. $\mathcal{L}_{\mathrm{stop}}$ is a binary cross-entropy on $\pi_i$ that labels the valid fixation slots as continue (0) and the first slot after them as stop (1), ignoring later slots, and scanpaths longer than $L$ are truncated. Further details are given in the supplementary material.

\subsection{Spatial Location Readout}
\label{sec:sr}

For each slot, SR sharpens the spatial distribution with a temperature $\kappa$ and takes the expectation over the grid-cell centers:
\begin{equation}
    \tilde{p}_{i,g}(\kappa) = \frac{p_{i,g}^{1/\kappa}}{\sum_{j} p_{i,j}^{1/\kappa}}, \qquad \hat{\mathbf{x}}^{\mathrm{SR}}_i(\kappa) = \sum_{g=1}^{G} \tilde{p}_{i,g}(\kappa)\, r_g.
\end{equation}
$\kappa = 1$ gives the plain expectation, denoted SR ($\kappa = 1$), which can fall between the peaks of a multimodal map; $\kappa \to 0^+$ approaches the argmax (for a unique maximum), which is restricted to grid-cell centers, and intermediate values suppress secondary peaks while keeping sub-grid coordinates. For TCSD, the native output equals SR ($\kappa = 1$). SR with $\kappa = 0.5$ is applied to it at test time without retraining.

\textbf{Choice of \(\kappa\).} We evaluated 14 candidate readouts on the COCO-Search18 development set, including plain and temperature-scaled expectations, argmax, top-5 expectation, and entropy-gated combinations (see the supplementary material). Selection used the mean rank across MultiMatch, ScanMatch, and SED, with ties resolved in that metric order. The selected temperature, \(\kappa = 0.5\), was then fixed for all evaluated parallel models with auxiliary spatial maps. The candidate set was formed after examining decoding comparisons on the target-present test set, whereas the choice among candidates and all termination thresholds use only development data. Results on that split are therefore exploratory. Evaluations on models, datasets, and task conditions outside the selection are termed transfer evaluations after rule locking.

\subsection{Termination Calibration}
\label{sec:tc}

By default, decoding stops before the first slot with $\pi_i \geq 0.5$, which does not guarantee human-like scanpath lengths. TC treats the stopping threshold $\tau$ as a calibration step independent of SR. For a given $\tau$, the number of decoded fixations is $\hat{N}(\tau) = \min\{i : \pi_i \geq \tau\} - 1$ (or $L$ if no such slot exists). We select the threshold for each trained checkpoint on the development set of the corresponding dataset:
\begin{equation}
    \hat{\tau} = \arg\min_{\tau \in \{0.05, 0.06, \dots, 0.95\}} \Big| \overline{\hat{N}(\tau)}_{\mathrm{dev}} - \overline{T}_{\mathrm{dev}} \Big|,
\end{equation}
where the bars denote means over development-set scanpaths, and $T$ is the human scanpath length and ties are broken toward 0.5. Matched termination thus means the same rule, not the same threshold. This rule relies only on human scanpath lengths rather than evaluation metrics. Each $\hat{\tau}$ is locked before decoding the corresponding test set. Because the target-present test set was already evaluated at $\tau = 0.5$ when TC was introduced, TC results on this condition are also exploratory.

\subsection{Scope and Transfer}
\label{sec:scope}

SR only requires a per-slot spatial distribution over a grid, allowing us to test it on two external models.

\textbf{Gazeformer} predicts a Gaussian over each fixation's coordinates but no per-slot distribution over the grid. We add an auxiliary spatial map head supervised as in Reg+Map (Gazeformer-Aux) and apply SR ($\kappa = 0.5$) without TC.

\textbf{HAT-SF} generates a spatial map at each step, takes the argmax after inhibition of return, and feeds that point back as history. We replace only this argmax with SR while keeping its native stopping rule. The readout is selected on the development set from $\{$argmax, $\kappa = 1, 0.5, 0.25\}$ using the same average-rank criterion, with ties instead resolved in favor of the native argmax. Since HAT-SF already decodes from its spatial map, this tests whether readout changes still help without an auxiliary spatial head. Models lacking per-fixation spatial distributions, such as ScanDiff, are out of scope, and IRL is not tested.
\section{Results}
\label{sec:result}

\subsection{Experimental Setup}
\textbf{Datasets and splits.} The main experiments use the target-present (TP) condition of COCO-Search18~\cite{chen2021coco,yang2020predicting}. We use the validation set of the official split1 as the held-out test set (3,258 scanpaths, 315 images). Both the training and development sets are drawn exclusively from the official training set. We stratify by target category at the image level, assigning approximately 10\% of the images in each category stratum to the development set (2,658 scanpaths, 219 images) and the remaining 18,885 scanpaths (1,707 images) to the training set. We remove the 8 images shared with the test set and their 79 training scanpaths. The three splits share no images. Because the TP test results were involved in designing the readout scheme, we report readout results on TP only as an exploratory analysis.

Transfer evaluations after rule locking use two settings. (i) COCO-Search18 target-absent (TA): We similarly use the official validation set as the test set (3,259 scanpaths, 309 images). From the official training set, we stratify by the category combination of the query target in each image, assigning approximately 10\% of the images as the development set (2,540 scanpaths, 198 images) and the remaining 18,957 scanpaths (1,587 images) for training. We remove the 12 images that appear in both official splits from the training set. All models, including HAT-SF, are trained and evaluated only on TA data, and our task text states that the target is absent. This setting therefore tests readout under known target absence rather than the detection of target presence. (ii) AiR-D~\cite{chen2020air}: This dataset uses visual question answering as the task condition, taking an image and the full question text as input. We re-split the dataset into training, development, and test sets at the image level with a ratio of approximately 80\% / 10\% / 10\% (11,121 / 1,399 / 1,383 scanpaths across 987 images). All questions associated with the same image are assigned to the same split.

\textbf{Evaluation metrics.} We use MultiMatch (MM)~\cite{dewhurst2012depends}, ScanMatch (SM)~\cite{cristino2010scanmatch}, and sequence edit distance (SED) for sequence similarity, and KLD, CC, and NSS~\cite{bylinskii2018different} to compare saliency maps from predicted and human fixations per ``task \(\times\) image'' group. Each map accumulates the group's fixations and is smoothed with a Gaussian (\(\sigma=22.4\) px); for deterministic models, the predicted map thus reflects a single scanpath. Values in the tables are averaged over random seeds. On TP and AiR-D, sequence metrics are averaged over all scanpaths, and saliency metrics over all groups. On TA, both are first averaged within each image and then across images. The human consistency reference uses leave-one-subject-out (Human LOO), and all methods use the same evaluator.

\textbf{Models and training.} Visual features are extracted from layer4 of a frozen Mask R-CNN~\cite{he2017mask} ResNet-50-FPN. Reg+Map and TCSD have 4.6M trainable parameters. We use AdamW~\cite{loshchilov2019decoupled} (learning rate \(1\times 10^{-4}\), weight decay \(5\times 10^{-5}\), gradient clipping 1.0, mixed precision). On COCO, the batch size is 16, with a maximum of 30 epochs and a patience of 8. On AiR-D, the batch size is 8, with a maximum of 12 epochs and a patience of 5. Early stopping and checkpoint selection are based on the combined loss on the development set (coordinate error, termination, and duration terms). Random seeds: On TP, Reg+Map and TCSD use 10 seeds each (3-12, where 6-12 are trained after \(\kappa\) is locked), and Reg uses 5 seeds (3-7). TA uses 5 seeds (3-7). AiR-D uses 3 seeds (3-5). The readout rule selection uses seeds 3-5 on the TP development set.

\textbf{External baselines.} All baselines are retrained on the same splits, selected by their native development-set losses, and evaluated by the same evaluator using seeds 3-5. Gazeformer~\cite{mondal2023gazeformer} uses its official architecture and losses. The visual input is a shared feature grid, and the task text is encoded by the CLS representation of a frozen RoBERTa~\cite{liu2019roberta}. It is trained for 30 epochs. Gazeformer-Aux adds auxiliary spatial map supervision on top of it (seeds 1-5). HAT-SF is a shared-feature adaptation of HAT~\cite{yang2024unifying}. The visual backbone and pixel decoder of the original implementation are replaced with a shared feature grid, while the rest (sequential task queries, fixation history working memory, dense map, termination head, maximum of 7 fixations, original loss) remains unchanged. HAT does not predict duration, so we uniformly use the median duration from the training set during export. Results for IRL~\cite{yang2020predicting} and ScanDiff~\cite{cartella2025modeling}, as well as training details for all baselines, are provided in the supplementary material.

\textbf{Statistical analysis.} We report 95\% bootstrap confidence intervals. Metrics are first aggregated within each image. Each resampling iteration samples training seeds and test images with replacement for 20,000 iterations, with shared sampling indices for both compared methods. Paired effects are averaged equally across images and may therefore differ slightly from the differences in table means. Effects are always raw differences (former minus latter). For SED and KLD, negative values indicate that the former is better. An interval containing zero indicates only that no difference was detected, not that the two are equivalent. Comparisons with external baselines use only the common seeds 3-5.

\textbf{Efficiency measurements.} Latency and throughput are measured on one RTX 4060 Laptop GPU at batch sizes 1 and 8 (30 warm-up and 200 timed runs, median of 3 rounds). Timing covers scanpath generation from pre-extracted visual features and task embeddings, including readout and termination, but not feature extraction, text encoding, or data loading; speed is compared only with our HAT-SF implementation.

\subsection{Main Results}
\label{sec:main}
Table~\ref{tab:main_results} shows COCO-Search18 target-present results.

\begin{table*}[t]
\centering
\caption{COCO-Search18 target-present test set. Reg uses 5 seeds, Reg+Map, TCSD, and their readouts use 10, and Gazeformer and HAT-SF use 3. SR rows reuse base checkpoints and are exploratory. Latency: median time per scanpath at batch size 1 (RTX 4060 Laptop). SR + TC adds at most 2.3\,ms. Gazeformer and HAT-SF use their native stopping rules.}
\label{tab:main_results}
\begin{tabular}{lccccccc}
\hline
\textbf{Method} & \textbf{MM \(\uparrow\)} & \textbf{SM \(\uparrow\)} & \textbf{SED \(\downarrow\)} & \textbf{KLD \(\downarrow\)} & \textbf{CC \(\uparrow\)} & \textbf{NSS \(\uparrow\)} & \textbf{Latency (ms)} \\
\hline
Reg & 0.781 & 0.253 & 4.60 & 19.56 & 0.094 & 0.50 & - \\
Reg+Map & 0.816 & 0.364 & 3.64 & 16.66 & 0.137 & 0.74 & 15.2 \\
+ SR (\(\kappa=1\)) & 0.833 & 0.473 & 3.20 & 11.43 & 0.328 & 2.05 & - \\
+ SR & 0.836 & 0.504 & 3.03 & 9.50 & 0.461 & 3.03 & - \\
+ SR + TC & 0.837 & 0.519 & 2.57 & 9.68 & 0.458 & 3.00 & 15.2 \\
TCSD & 0.830 & 0.456 & 3.78 & 11.32 & 0.319 & 1.98 & 14.1 \\
+ SR + TC & 0.837 & 0.518 & 2.55 & 9.63 & 0.455 & 2.97 & 16.4 \\
\hline
Gazeformer & 0.811 & 0.416 & 2.60 & 12.98 & 0.306 & 2.14 & - \\
HAT-SF & 0.832 & 0.515 & 2.12 & 9.40 & 0.530 & 3.68 & 51.5 \\
Human (LOO) & 0.847 & 0.539 & 2.13 & 6.79 & 0.470 & 2.98 & - \\
\hline
\end{tabular}
\end{table*}

\textbf{The auxiliary spatial map of Reg+Map outperforms its native regression output.} For the same set of Reg+Map weights, replacing only the coordinate source from the regression head to the expectation over the spatial map (\(\kappa=1\)) significantly improves all six metrics. Specifically, ScanMatch improves by +0.109 [0.092, 0.127], KLD by \(-\)5.23 [\(-\)6.06, \(-\)4.44], and CC by +0.191 [0.162, 0.220]. This indicates that the supervised spatial map contains localization information that the regression output does not fully utilize. When replacing the linear regression head with an MLP, ScanMatch only rises to 0.394. Readout from the spatial map of the linear-head model remains significantly better (+0.075 [0.055, 0.101], seeds 3-7). This shows that the gap cannot be explained solely by the capacity of the tested linear regression head. Temperature sharpening further improves all six metrics over plain expectation, with ScanMatch gaining +0.031 [0.026, 0.036] and CC gaining +0.133 [0.119, 0.147]. When evaluating only on seeds 6-12 trained after \(\kappa\) locking, the gain magnitude is similar (ScanMatch +0.032 [0.026, 0.038]). Termination calibration further reduces SED from 3.03 to 2.57.

\textbf{No consistent additional gain from end-to-end training.} Under native outputs, TCSD significantly outperforms Reg+Map on five metrics except for SED (ScanMatch +0.091 [0.072, 0.111]). However, the native output of TCSD is itself the expectation over its spatial map, corresponding to Reg+Map + SR (\(\kappa=1\)). Under this identical readout and without termination calibration, TCSD is significantly worse on MultiMatch, ScanMatch, and SED (ScanMatch \(-\)0.018 [\(-\)0.031, \(-\)0.006]). After both models calibrate termination under the same rule, the intervals for all six metrics contain zero (ScanMatch \(-\)0.001 [\(-\)0.008, 0.006]). Therefore, comparing training pathways requires jointly controlling location readout and termination strategy. An exploratory equivalence analysis with bounds set by existing test gains is provided in the supplementary material.

\textbf{The gap to autoregressive models narrows.} On the seeds 3-5, Reg+Map + SR + TC shows no detectable difference from HAT-SF on MultiMatch, ScanMatch, and KLD, though HAT-SF remains better on SED, CC, and NSS. This depends on first-fixation handling. With the first fixation uniformly removed, HAT-SF leads on ScanMatch (\(-\)0.023 [\(-\)0.037, \(-\)0.007]), while we lead on MultiMatch (+0.024 [0.008, 0.040]). When HAT-SF's stopping threshold is calibrated by the same rule (mean length 2.9 to 3.9), we surpass it on ScanMatch (+0.017 [0.006, 0.028]), while HAT-SF remains better on KLD, CC, and NSS. Details are in the supplementary material. The parallel model generates all fixations in one forward pass and is \(3.4\times\) faster than our HAT-SF implementation at batch size 1 (Table~\ref{tab:main_results}).

\subsection{Cross-task Conditions and Datasets}
\label{sec:transfer_tasks}
We apply the locked SR + TC to two settings that did not participate in the selection: the target-absent condition of the same dataset, and AiR-D with a different task formulation (Table~\ref{tab:transfer}). Table~\ref{tab:transfer} reflects the joint effect of both components. Their individual contributions are detailed in Section~\ref{sec:ablation}.

\begin{table}[t]
\centering
\caption{Calibrated inference readout (SR + TC) in the two transfer settings. TA: 5 seeds (HAT-SF: 3), image-level means. AiR-D: 3 seeds. Len: mean scanpath length (development-set human mean: 5.6 / 10.1). HAT-SF uses its native stopping rule}
\label{tab:transfer}
\setlength{\tabcolsep}{2.5pt}
\resizebox{\columnwidth}{!}{%
\begin{tabular}{llccccccc}
\hline
\textbf{Setting} & \textbf{Method} & \textbf{MM \(\uparrow\)} & \textbf{SM \(\uparrow\)} & \textbf{SED \(\downarrow\)} & \textbf{KLD \(\downarrow\)} & \textbf{CC \(\uparrow\)} & \textbf{NSS \(\uparrow\)} & \textbf{Len} \\
\hline
\multirow{4}{*}{TA} & Reg+Map & 0.799 & 0.137 & 13.93 & 16.86 & 0.155 & 0.56 & 15.5 \\
 & + SR + TC & 0.815 & 0.320 & 6.14 & 15.09 & 0.262 & 1.09 & 6.1 \\
 & TCSD + SR + TC & 0.815 & 0.317 & 6.19 & 15.03 & 0.266 & 1.11 & 6.2 \\
 & HAT-SF & 0.821 & 0.328 & 5.41 & 11.94 & 0.362 & 1.61 & 6.5 \\
\hline
\multirow{3}{*}{AiR-D} & Reg+Map & 0.786 & 0.252 & 10.13 & 10.18 & 0.366 & 0.96 & 12.5 \\
 & + SR + TC & 0.793 & 0.327 & 7.97 & 7.70 & 0.456 & 1.27 & 10.1 \\
 & TCSD + SR + TC & 0.792 & 0.323 & 8.00 & 7.66 & 0.443 & 1.23 & 10.1 \\
\hline
\end{tabular}}
\end{table}

For Reg+Map, switching from native regression to SR ($\kappa=0.5$)
with the checkpoint and calibrated stopping threshold held fixed
leaves every predicted scanpath length unchanged. ScanMatch improves
by $0.045$ on target-absent search (95\% paired CI $[0.038, 0.053]$)
and by $0.045$ on AiR-D (95\% paired CI $[0.033, 0.057]$).
The paired intervals favor SR on all six metrics for target-absent
search and on five of six for AiR-D. Relative to uncalibrated native
predictions, SR + TC improves all six metrics in both settings, with
consistent directions across seeds. With the same spatial readout
and termination calibration procedure, the paired intervals for
TCSD versus Reg+Map contain zero for all six metrics in both settings,
consistent with Section~\ref{sec:main}.

\textbf{Largest improvement on target-absent.} The average length of uncalibrated predictions is 15.5 fixations, approaching the upper limit of 16 slots. In contrast, the human average length on the development set is 5.6. After calibration, the average predicted length on the test set drops to 6.1. Most of the gain in sequence metrics comes from this step (Section~\ref{sec:ablation}). When HAT-SF’s stopping threshold is calibrated by the same rule, HAT-SF is better on all six metrics for target-absent (ScanMatch \(-\)0.026 [\(-\)0.039, \(-\)0.013]).

\subsection{Transfer to Other Architectures}
\label{sec:transfer_arch}

\textbf{Gazeformer.} We add a supervised auxiliary spatial head to Gazeformer and compare readouts on the same five checkpoints, retaining the native stopping rule. With the locked temperature \(\kappa=0.5\), SR improves all six metrics over native Gaussian sampling: ScanMatch rises from \(0.416\) to \(0.456\), and CC from \(0.308\) to \(0.405\). Gaussian-mean decoding also yields a ScanMatch of \(0.416\); SR improves it by \(0.040\) (\(95\%\) CI [\(0.030\), \(0.050\)]). Plain expectation (\(\kappa=1\)) improves only MultiMatch significantly and degrades KLD by \(0.62\) ([\(0.09\), \(1.18\)]). Sharpening significantly improves five metrics over plain expectation: ScanMatch, SED, KLD, CC, and NSS. Full results are in the supplementary material.


\textbf{HAT-SF.} HAT-SF takes the argmax from its spatial map at each step. Under the rules in Section \ref{sec:scope}, development-set selection retains this native argmax. Development-set ScanMatch scores are 0.505 (argmax), 0.495 (\(\kappa=0.25\)), 0.398 (\(\kappa=0.5\)), and 0.214 (\(\kappa=1\)). Forcing \(\kappa=0.5\) on target-absent significantly degrades all six metrics. Because HAT-SF's current position affects subsequent generation, this boundary test differs from replacing coordinates after fixed parallel outputs.

\subsection{Ablations and Analysis}
\label{sec:ablation}

\textbf{Contributions of the two components vary by task and metric.} Table~\ref{tab:ablation} isolates the effects of spatial readout and termination calibration. SR consistently improves ScanMatch, SED, and KLD across settings, while TC further boosts ScanMatch and SED. Although TC slightly degrades KLD on COCO, it improves it on AiR-D. TC mainly shortens overly long sequences. On TP, the 90th percentile length drops from 16 (the slot limit) to 4, close to humans (5) and HAT-SF (4), and the average length error falls from 2.02 to 1.45. This effect is most prominent on target-absent search, where TC alone raises ScanMatch from 0.137 to 0.275 and cuts mean length from 15.46 to 6.13. There, a fixed length chosen on the development set performs comparably on ScanMatch (calibrated minus fixed: \(+0.007\) \([-0.003, 0.017]\)). In contrast, permuting calibrated lengths across images lowers it by \(0.015\) \([0.006, 0.023]\). TC thus mainly corrects the mean length. Labelling all post-end slots in the stop loss (seeds 3-5) cuts native length from 15.7 to 4.8. With this loss, TC reduces ScanMatch by 0.019 [0.015, 0.023], while SR still adds 0.055 [0.048, 0.062].

\begin{table}[t]
\centering
\caption{Contributions of the two components (Reg+Map). ``+ SR'' compares SR with native regression at the default threshold (\(\tau=0.5\)), and ``+ TC'' compares SR + TC with SR. Cells show raw differences. \(\checkmark\) and \(\times\) mark significant improvement and degradation (95\% paired interval excludes zero). On TP, ``+ SR'' uses seeds 3-5 and ``+ TC'' uses 10 seeds, so the two are not additive. TA uses 5 seeds and AiR-D uses 3. Section~\ref{sec:transfer_tasks} reports SR at the calibrated \(\tau\).}
\label{tab:ablation}
\setlength{\tabcolsep}{2.5pt}
\resizebox{\columnwidth}{!}{%
\begin{tabular}{llcccc}
\hline
\textbf{Setting} & \textbf{Step} & \textbf{SM} & \textbf{SED} & \textbf{KLD} & \textbf{CC} \\
\hline
\multirow{2}{*}{Target-present} & + SR & +0.136 \(\checkmark\) & \(-\)0.63 \(\checkmark\) & \(-\)7.28 \(\checkmark\) & +0.329 \(\checkmark\) \\
 & + TC & +0.015 \(\checkmark\) & \(-\)0.46 \(\checkmark\) & +0.18 \(\times\) & \(-\)0.002 \\
\hline
\multirow{2}{*}{Target-absent} & + SR & +0.029 \(\checkmark\) & \(-\)0.58 \(\checkmark\) & \(-\)2.51 \(\checkmark\) & +0.008 \\
 & + TC & +0.155 \(\checkmark\) & \(-\)7.21 \(\checkmark\) & +0.74 \(\times\) & +0.099 \(\checkmark\) \\
\hline
\multirow{2}{*}{AiR-D} & + SR & +0.037 \(\checkmark\) & \(-\)0.52 \(\checkmark\) & \(-\)2.50 \(\checkmark\) & +0.076 \(\checkmark\) \\
 & + TC & +0.034 \(\checkmark\) & \(-\)1.61 \(\checkmark\) & \(-\)0.03 \(\checkmark\) & +0.009 \(\checkmark\) \\
\hline
\end{tabular}}
\end{table}

\textbf{Robustness to temperature.} A sweep on the COCO development set reveals a plateau for $\kappa \in [0.25,0.7]$ (ScanMatch 0.471-0.482). The tested readouts with \(\kappa \le 1\), as well as argmax (0.477), all outperform native regression (0.353). At \(\kappa=1.5\), ScanMatch drops to 0.416.

\textbf{Readout increases early target hits.} We measure whether any of the first k fixations falls inside the target bounding box (Hit@k, seeds 3-5, without termination calibration). After uniformly removing the first fixation, Hit@3 rises from 0.239 (native regression) to 0.574 (plain expectation). Sharpened readout further raises it to 0.778 (human 0.832, HAT-SF 0.913).

\textbf{Robust across target categories.} Comparing Reg+Map + SR + TC with native regression across target categories (seeds 3-5), all ScanMatch improvement intervals are above zero across the 18 categories (ranging from +0.085 for keyboard to +0.234 for fork). On seed 3, grouped by ``task \(\times\) image'', 92\% of the groups show improvement.

\textbf{Other controls.} Items (i) and (ii) use seeds 3-5. (i) \textit{Coordinate supervision target.} Under exact-coordinate supervision, native Reg+Map achieves a ScanMatch of \(0.348\). Plain spatial readout improves all six metrics (ScanMatch \(+0.129\), 95\% CI \([0.101, 0.153]\)), while sharpened readout reaches \(0.503\). With identical exact-coordinate supervision, \(\kappa=1\), and the default stopping rule, all six paired intervals between TCSD and Reg+Map contain zero. (ii) \textit{Task conditioning and spatial access.} A pure regression model, whose decoder directly attends to task-fused features, lags behind native TCSD by \(0.208\) in ScanMatch (\([0.192, 0.223]\), image-only bootstrap). Giving Reg+Map the same decoder access raises its native ScanMatch from \(0.365\) to \(0.408\). Yet, reading this checkpoint's map still improves all six metrics, with a gain of \(+0.071\) (95\% CI \([0.049, 0.098]\)) for \(\kappa=1\) and \(+0.103\) (\([0.082, 0.130]\)) for \(\kappa=0.5\). For Spatial Matched Regression, which pools features with its spatial map for coordinate regression, reading coordinates from this map with checkpoint, stopping rule, and durations fixed raises ScanMatch from \(0.431\) to \(0.459\). The gain is \(+0.028\) \([0.019, 0.038]\) for \(\kappa=1\) and \(+0.055\) \([0.043, 0.069]\) for \(\kappa=0.5\). Paired intervals favor spatial readout on all six metrics. Additionally feeding the slot's full spatial map to this head raises native ScanMatch to \(0.436\), yet reading the map still adds \(+0.016\) \([0.003, 0.029]\) at \(\kappa=1\) and \(+0.049\) \([0.025, 0.068]\) at \(\kappa=0.5\). (iii) \textit{Center bias reference.} The image-center baseline achieves 0.246 ScanMatch, 0.341 CC, and 2.70 NSS. This illustrates the strong influence of center bias on saliency metrics. A map-only variant (TCSD without coordinate loss) reaches \(0.461\) with \(\kappa=1\) and \(0.487\) with \(\kappa=0.5\) (\(+0.026\), \([0.019, 0.033]\)).

\textbf{Qualitative results and failure modes.} For cases selected by a predefined rule (see supplementary material), the regression output often stops on the background near the target. In contrast, the spatial map peaks on the target, and the readout follows this peak. In failure cases, human search is dispersed, but the spatial map concentrates at the image center. Readout cannot correct such localization errors inherent in the map.

\section{Discussion}
\label{sec:discussion}

\paragraph{Learning and Use of Representations.}
For models with both a spatial head and a regression head, a same-checkpoint readout comparison is a simple diagnostic of whether supervised representations are used at prediction time. Extracting coordinates from spatial distributions is not new~\cite{sun2018integral,nibali2018numerical}. We instead ask whether a trained model's own map holds information its regression output leaves underused. The gap persists with an MLP head, a map-pooling head, a full-map head, and a decoder attending to task-fused features. With readout and termination matched, we detect no gain from TCSD, without implying that the two training schemes are equivalent. Smooth-L1 with $\beta = 1$ on normalized coordinates is a squared loss whose optimum is the conditional mean, which the plain map expectation also estimates. The plain-expectation gain shrinks as the head gains spatial information (seeds 3-5: 0.106, 0.071, 0.028, and 0.016 for Reg+Map, equal access, Spatial Matched Regression, and full-map regression), whereas sharpening toward the dominant mode adds 0.026--0.032 in every controlled model and transfers to Gazeformer-Aux, paralleling the failure of mean regression on multimodal targets~\cite{bishop1994mixture,chai2019multipath}. Unlike HAT's dense-versus-regression comparison~\cite{yang2024unifying}, these results fix the weights.

\paragraph{Readout rules and task behavior.}
Location readout and termination calibration act on different stages, but their effects on evaluation metrics are interrelated. Under the target-absent condition, the single-positive stop loss leaves uncalibrated lengths near the slot limit, so termination calibration accounts for most of the sequence-metric gains. Because predicted saliency maps are built from decoded fixations, changes in saliency metrics cannot be attributed solely to location accuracy. With termination held fixed, SR increases the rate at which early fixations hit the target. This indicates that the gain also manifests in concrete search behavior. The presence of a spatial map does not guarantee that every readout is effective: HAT-SF~\cite{yang2024unifying} retains its native argmax. We therefore argue for evaluating whether the representation and readout match, rather than assuming that expectation or a fixed temperature is universally optimal.

\paragraph{Comparison with strong baselines.}
Close values on some metrics do not imply comparable overall performance. HAT-SF remains superior on several metrics. Furthermore, the outcome on ScanMatch and MultiMatch changes when the first fixation is removed. The observed speed advantage holds only against our HAT-SF implementation, not the official full HAT. Because capacity, optimization, and implementation are not strictly matched, the remaining accuracy gap cannot be attributed solely to parallel versus autoregressive generation.

\paragraph{Limitations.}
The target-present results are exploratory because the test results informed the readout design. Seeds 6-12, trained after locking \(\kappa\), show similar gains. HAT-SF uses shared features and training-set median durations, while Gazeformer's duration loss is still decreasing at 30 epochs. Richer regression inputs narrow but do not close the readout gap, and TC mainly compensates for the single-positive stop loss. SR requires an existing spatial head, and TC requires a development set per setting. Of the external baselines, only HAT-SF is evaluated with calibrated termination, while AiR-D~\cite{chen2020air} lacks a retrained baseline. TCSD is trained with plain expectation under a single loss setting. Our conclusions apply to the tested architectures and tasks. Sensitivity to question content and free viewing remain untested.

\section{Conclusion}
\label{sec:conclusion}

We investigated how parallel scanpath models map learned spatial representations to fixation coordinates. In a controlled model family, same-checkpoint comparisons reveal that a supervised auxiliary spatial map contains localization information left underused by the native regression output, a mismatch between representation and readout. The gap persists even when the regression head reads the full spatial map. To address this, we propose calibrated inference readout. It combines temperature-sharpened spatial readout with development-set termination calibration without requiring model weight updates. In paired comparisons with readout and termination matched, we detect no additional gain from end-to-end spatial map training. After rule locking, SR + TC improves all six metrics over native predictions on target-absent search and AiR-D. SR alone also improves all six metrics on Gazeformer-Aux. Development-set selection for HAT-SF retains its native argmax, highlighting the need for per-model readout assessment. These findings suggest evaluating not only how spatial representations are learned, but also how they are used, and jointly controlling readout and termination when comparing training pathways. Future work will extend this analysis to free viewing and to baselines using their original visual frontends.
{
    \small
    \bibliographystyle{ieeenat_fullname}
    \bibliography{main}
}

\onecolumn
\appendix
\section*{Appendix}

\section{Model and training details}
\label{sec:supp_training}

This section provides further details on the architecture, supervision, and training protocol from Sections~3.1, 3.2, and~4.1 of the main paper. External-baseline training is covered in Section~\ref{sec:supp_base_training}.

\subsection{Architecture}
\label{sec:supp_train_arch}

Table~\ref{tab:supp_train_arch} lists the components shared by Reg, Reg+Map,
and TCSD.

\begin{table}[H]
\centering
\caption{Shared backbone of the controlled model family.}
\label{tab:supp_train_arch}
\footnotesize
\setlength{\tabcolsep}{4pt}
\begin{tabular}{@{}ll@{}}
\toprule
Item & Value \\
\midrule
Visual features & Frozen Mask R-CNN~\cite{he2017mask} ResNet-50-FPN layer4, 2048-d, $24\times32$ grid ($G=768$) \\
Task encoding & Frozen RoBERTa~\cite{liu2019roberta} (768-d), trainable pooler and projection \\
Hidden size $D$ & 256 \\
Attention heads & 8 \\
Encoder / decoder layers & 2 / 2 \\
Feed-forward size & 256 \\
Activation & ReLU \\
Dropout & Encoder 0.1, decoder 0.2 \\
Slots $L$ & 16 \\
Trainable parameters & 4.6M for Reg+Map and TCSD (128.9M in total, the rest is frozen RoBERTa) \\
\bottomrule
\end{tabular}
\end{table}

\paragraph{Output heads.}
Separate linear heads process each slot state \(h_i\). One head outputs the stop probability \(\pi_i\) through a sigmoid. Another predicts the parameters \(\mu_i\) and \(\log\sigma_i^2\) of a log-normal duration distribution. At inference, the duration is the distribution median, \(\exp(\mu_i)\). The spatial distribution follows Eq.~(1) of the main paper. It is computed as a scaled dot product between the slot state and the task-fused features. In Reg and Reg+Map, the coordinate head is a single linear layer. It takes \(h_i+\bar v\) as input and applies a sigmoid to output normalized coordinates.

\paragraph{Control variants.}
Table~\ref{tab:supp_train_variants} lists how each variant differs from its reference configuration. Most variants use Reg+Map as the reference. The map-only variant uses TCSD, FullMap uses SMR, FullMap-MDN uses FullMap, and Gazeformer-Aux uses Gazeformer. SMR is the Spatial Matched Regression from Section~4.5. FullMap is the full-map regression from Section~5. All unlisted components match the reference configuration. This includes the architecture, loss weights, and training settings.

\begin{table*}[t]
\centering
\caption{Control variants and their changes relative to the reference configuration.}
\label{tab:supp_train_variants}
\footnotesize
\setlength{\tabcolsep}{4pt}
\renewcommand{\arraystretch}{1.15}
\begin{tabular}{@{}P{3.0cm}P{10.2cm}P{3.6cm}@{}}
\toprule
Variant & Change relative to the reference configuration & Where it appears \\
\midrule
Reg & Removes the spatial map head and the map loss ($\alpha_m=0$). & Main paper, Table~1 \\
TCSD & No coordinate head. The coordinate is the expectation over the spatial map, and the coordinate
  loss acts on this expectation. & Main paper, Table~1 and Section~4.2 \\
MLP head & The coordinate head becomes a two-layer MLP ($2D\rightarrow D\rightarrow2$, ReLU). Its input is
  the slot state concatenated with the mean of the task-fused features. & Main paper, Section~4.2 \\
Map-only & TCSD without the coordinate loss. Only the map loss supervises it. & Main paper, Section~4.5 \\
Exact-coordinate supervision & The coordinate target is the continuous human fixation instead of the
  center of its grid cell. & Main paper, Section~4.5 \\
Regression with task access & The decoder of Reg attends directly to the task-fused features (the
  pure regression model of Section~4.5). & Main
  paper, Section~4.5 \\
Equal access & The decoder of Reg+Map attends directly to the task-fused features. & Main paper,
  Section~4.5, and Section~\ref{sec:supp_replication} \\
SMR & Removes the linear coordinate head. A two-layer MLP regresses the coordinate from the slot state
  concatenated with the task-fused features pooled by the slot's spatial map. Smooth-L1 supervision. &
  Main paper, Section~4.5, and Sections~\ref{sec:supp_mechanism} and~\ref{sec:supp_replication} \\
FullMap & Adds the slot's full 768-d spatial distribution to the input of SMR. & Main paper,
  Section~4.5, and Sections~\ref{sec:supp_mechanism} and~\ref{sec:supp_replication} \\
FullMap-MDN & The regression head of FullMap outputs a mixture of $K=4$ diagonal Gaussians and is
  trained by negative log-likelihood. The coordinate is the mean of the component with the largest
  weight. & Section~\ref{sec:supp_mechanism} \\
All-slot stop loss & The stop loss covers every slot, and the end slot and every slot after it have label
  1. The stop term of the development selection metric changes in the same way. & Main paper,
  Section~4.5, and Section~\ref{sec:supp_termination} \\
Gazeformer-Aux & Adds to Gazeformer an auxiliary spatial map head supervised in the same way as in
  Reg+Map. & Main paper, Section~4.4, and Section~\ref{sec:supp_baselines} \\
\bottomrule
\end{tabular}
\end{table*}

\subsection{Supervision targets}
\label{sec:supp_train_targets}

\paragraph{Length and truncation.}
Each human scanpath keeps only its first $L=16$ fixations. The slots of the kept fixations are valid
slots ($m_i=1$).

\paragraph{Stop labels.}
The stop loss is a binary cross-entropy on \(\pi_i\). Valid slots have the label ``continue'' (0). The first slot after the valid slots has the label ``stop'' (1). Later slots are excluded from the stop loss. For human scanpaths with 16 or more fixations, all 16 slots are valid. Therefore, no slot receives a positive stop label. We refer to this setup as the ``original stop loss'' in Section~\ref{sec:supp_termination}.

\paragraph{Coordinate targets.}
Human fixation coordinates are normalized by the image size. By default, the coordinate target is the center of the grid cell containing the fixation ($r_{g^*}$ in Eq.~(2) of the main paper). The exact-coordinate variant instead uses continuous coordinates. The coordinate loss uses Smooth-L1 ($\beta=1$). It is averaged first over the two coordinate dimensions and then over the valid slots.

\paragraph{Spatial map targets.}
For each valid slot, we place an isotropic Gaussian ($\sigma=1$ grid cell) at the continuous human fixation position on the $24\times32$ grid. We then normalize it over the grid. The map loss is the cross-entropy between this target and the predicted distribution (Eq.~(3) of the main paper).

\paragraph{Durations.}
The duration loss is the negative log-likelihood under the log-normal distribution, averaged over valid slots. We divide input durations by 1000 in the data pipeline. Because COCO-Search18 records durations in milliseconds, the model operates in seconds. We apply the same scaling to AiR-D. Its raw MAT files lack a verified time unit in the dataset manifest. Therefore, we make no claims regarding its original unit.

\paragraph{Total loss.}
$\mathcal{L}=\mathcal{L}_{\text{stop}}+\lambda_d\mathcal{L}_{\text{dur}}+\alpha_m\mathcal{L}_{\text{map}}
+\lambda_x\mathcal{L}_{\text{coord}}$ with $\lambda_d=1$. Reg uses $(\alpha_m,\lambda_x)=(0,1)$, Reg+Map and
TCSD use $(1,1)$, and the map-only variant uses $(1,0)$.

\subsection{Training configuration and checkpoint selection}
\label{sec:supp_train_config}

Table~\ref{tab:supp_train_config} lists the training settings by dataset.

\begin{table}[t]
\centering
\caption{Training configuration. COCO-TA keeps the core architecture and optimization hyperparameters of
COCO-TP. It is trained and evaluated on its own TA split with its own data and feature files, and its task
text states explicitly that the target is absent.}
\label{tab:supp_train_config}
\footnotesize
\setlength{\tabcolsep}{4pt}
\begin{tabular}{@{}lll@{}}
\toprule
Item & COCO-TP / COCO-TA & AiR-D \\
\midrule
Optimizer & AdamW~\cite{loshchilov2019decoupled}, learning rate $10^{-4}$, weight decay $5\times10^{-5}$ & Same \\
Gradient clipping & 1.0 & 1.0 \\
Mixed precision & fp16 & fp16 \\
Batch size & 16 & 8 \\
Maximum epochs & 30 & 12 \\
Early-stopping patience & 8 & 5 \\
Image canvas & $1680\times1050$ & $512\times384$ \\
Task text & Task description of the target category & Full question text, at most 64 tokens \\
\bottomrule
\end{tabular}
\end{table}

\paragraph{Checkpoint selection.}
For each scanpath, the combined metric sums three terms. After each epoch, we compute this metric on the development set and retain the checkpoint with the lowest value. Early stopping relies on the same metric.

\begin{itemize}
\item Coordinate error. The mean pixel distance between predicted and target coordinates over valid slots, normalized by the image diagonal.
\item Stop negative log-likelihood. Averaged over slots in the stop loss.
\item Duration negative log-likelihood (weight \(\lambda_d=1\)). Averaged over valid slots.
\end{itemize}

Predicted coordinates are the native output of each model. They come from the regression head in Reg and Reg+Map, and from the spatial map expectation in TCSD. The metric computes the three-term sum per scanpath, averages within each validation batch, and then averages across batches. The map loss is excluded from the selection metric. The all-slot stop variant substitutes the stop term with its all-slot definition (Section~\ref{sec:supp_termination}).

\paragraph{Seeds.}
On COCO-TP, Reg+Map and TCSD use 10 seeds each (3 to 12) and Reg uses 5 seeds (3 to 7). Seeds 6 to 12 were trained after locking \(\kappa\). COCO-TA uses 5 seeds (3 to 7), and AiR-D uses 3 seeds (3 to 5). We report seed counts for control variants in their respective sections.

\subsection{Data splits}
\label{sec:supp_train_splits}

Table~\ref{tab:supp_train_splits} gives the splits for COCO-Search18 target-present
(COCO-TP), COCO-Search18 target-absent
(COCO-TA), and AiR-D, as described in
Section~4.1 of the main paper.

\begin{table*}[t]
\centering
\caption{Data splits (scanpaths / images).}
\label{tab:supp_train_splits}
\footnotesize
\setlength{\tabcolsep}{4pt}
\renewcommand{\arraystretch}{1.15}
\begin{tabular}{@{}lcccP{8.2cm}@{}}
\toprule
Setting & Train & Development & Test & Split rule \\
\midrule
COCO-TP & 18,885 / 1,707 & 2,658 / 219 & 3,258 / 315 & The test set is the official split1 validation
  set. The development set is drawn from the official training set, stratified by target category, with
  about 10\% of the images. The 8 images shared with the test set and their 79 training scanpaths are
  removed. \\
COCO-TA & 18,957 / 1,587 & 2,540 / 198 & 3,259 / 309 & The test set is the official validation set. The
  development set takes about 10\% of the images, stratified by the category combination of the query
  targets in each image. The 12 images shared by both official splits are removed. \\
AiR-D & 11,121 / 793 & 1,399 / 93 & 1,383 / 101 & Re-split at the image level into about
  80\% / 10\% / 10\%, 987 images in total. All questions of one image go to the same split. \\
\bottomrule
\end{tabular}
\end{table*}

Within each setting, splits share no physical images. Test results did not influence gradient updates or checkpoint selection. However, they did inform the COCO-TP readout design (Section~\ref{sec:supp_readout_selection}).

\section{Evaluation protocol, statistics, and efficiency}
\label{sec:supp_evaluation}

All methods use the evaluator from Section~4.1 of the main paper. Below, we detail its metric settings, resampling procedure, and timing protocol.

\subsection{Metrics and aggregation}
\label{sec:supp_eval_metrics}

\paragraph{Sequence metrics.}
Each predicted scanpath is compared against its corresponding human scanpath in image canvas pixel coordinates ($1680\times1050$ for COCO-Search18, $512\times384$ for AiR-D).
\begin{itemize}
\item MultiMatch~\cite{dewhurst2012depends}. We average its five dimensions: vector, direction, length, position, and duration. Details for each dimension are in Section~\ref{sec:supp_readout_variants}.
\item ScanMatch~\cite{cristino2010scanmatch}. We divide the canvas into $16\times12$ regions and set the threshold to 3.5 without using durations.
\item SED. We divide the canvas into a $5\times5$ grid. Both scanpaths are converted into region strings, and the Levenshtein distance between them is computed.
\end{itemize}
The evaluator pads any sequence with fewer than 3 fixations up to 3 fixations before computing these metrics. The added points have pixel coordinates $(1,1)$ and a duration of 1~ms. They are excluded from saliency metrics and length statistics.

\paragraph{Saliency metrics.}
For each task-image group, we accumulate all human fixations and predicted fixations onto two separate pixel maps. We then smooth each map with a Gaussian (\(\sigma=22.4\) pixels) and compute KLD, CC, and NSS~\cite{bylinskii2018different}. NSS uses the normalized predicted map at the human fixation locations. A deterministic model produces one scanpath per group, so its predicted map reflects this single scanpath. We floor and clip coordinates to the canvas. Human coordinates follow their one-based indexing convention.

\paragraph{Aggregation.}
\begin{itemize}
\item COCO-TP and AiR-D. We average sequence metrics over all scanpaths (3,258 for COCO-TP). We average saliency metrics over all task-image groups (326 for COCO-TP).
\item COCO-TA. We first average both metrics within each image, then across images.
\item Table values are means over seeds.
\end{itemize}

\paragraph{First fixation.}
Unless stated otherwise, predicted and human sequences retain their first fixation. Results excluding it are in Section~\ref{sec:supp_base_first}.

\paragraph{Human reference.}
We compute the human consistency reference using leave-one-subject-out (Human LOO) with the same evaluator.

\subsection{Statistical analysis}
\label{sec:supp_eval_stats}

\paragraph{Paired bootstrap.}
For two methods under comparison, we first compute the difference for each seed and test image. We average sequence metrics over the scanpaths of that image, and saliency metrics over its groups. Each replicate draws training seeds and test images with replacement. Both methods use the same drawn indices. We then compute the mean difference with images weighted equally. We run 20,000 replicates and report 95\% percentile intervals (2.5\% and 97.5\% quantiles).

\paragraph{Conventions.}
\begin{itemize}
\item Differences are raw values (former minus latter). For SED and KLD, negative values favor the former.
\item Paired differences weight images equally. Table means on COCO-TP and AiR-D average over scanpaths or groups. Therefore, subtracting two mean columns may not equal the paired difference.
\item An interval containing zero indicates no detected difference, not equivalence. Equivalence analysis is in Section~\ref{sec:supp_equivalence}.
\item Between-model comparisons with external baselines use common seeds 3 to 5. Same-checkpoint Gazeformer-Aux comparisons use seeds 1 to 5.
\item Some tables report the number of seeds favoring the first method to demonstrate cross-seed agreement.
\item \(^{\checkmark}\) marks a 95\% interval excluding zero in favor of the first term. \(^{\times}\) marks one favoring the second term. \(^\dagger\) marks one containing zero.
\item All paired intervals in this supplement follow these conventions. We note deviations in the respective sections, such as replicate counts or resampling units.
\end{itemize}

\paragraph{Multiple comparisons.}
We report separate 95\% intervals for each metric and comparison. We apply no correction for multiple comparisons. Results whose intervals only just exclude zero warrant caution.

\subsection{Efficiency measurement}
\label{sec:supp_eval_efficiency}

\paragraph{Setup.}

We benchmark seed-3 checkpoints on a single RTX 4060 Laptop GPU (PyTorch 2.1.2, CUDA 12.1). Each of the three rounds consists of 30 warm-up iterations and 200 timed iterations. We rotate the model order across rounds and report the median of the three round-wise mean batch latencies. All models start from pre-extracted visual features. The timed call
includes the model forward pass, location readout, and termination decision. For the parallel models, this includes the RoBERTa forward pass on pre-tokenized task text. HAT-SF instead receives a target category index and does not run a text encoder. Tokenization, visual feature extraction, and data loading are outside the timed call. The latency values and speed ratios reported in the main text and Table~\ref{tab:supp_eval_efficiency} apply to this setup; they measure neither decoder-only latency with cached task embeddings nor end-to-end latency from raw inputs. For TC, we use the threshold selected on the development set for each checkpoint: 0.47 for Reg+Map and 0.45 for TCSD.

\begin{table*}[t]
\centering
\caption{Inference efficiency (seed 3). ``Per scanpath'' is the batch latency divided by the batch size.
``vs.\ HAT-SF'' is the latency of HAT-SF divided by the latency of the configuration at the same batch
size.}
\label{tab:supp_eval_efficiency}
\footnotesize
\setlength{\tabcolsep}{5pt}
\begin{tabular}{@{}cl ccccc@{}}
\toprule
Batch & Configuration & \makecell{Batch latency\\(ms)} & \makecell{Per scanpath\\(ms)} &
  \makecell{Throughput\\(scanpaths/s)} & \makecell{Peak memory\\(MiB)} & vs.\ HAT-SF \\
\midrule
1 & Reg+Map (native) & 15.23 & 15.23 & 65.6 & 1109 & $3.38\times$ \\
1 & Reg+Map + SR + TC & 15.18 & 15.18 & 65.9 & 1109 & $3.39\times$ \\
1 & TCSD (native) & 14.11 & 14.11 & 70.9 & 1109 & $3.65\times$ \\
1 & TCSD + SR + TC & 16.38 & 16.38 & 61.0 & 1109 & $3.15\times$ \\
1 & HAT-SF & 51.53 & 51.53 & 19.4 & 1113 & $1.00\times$ \\
\midrule
8 & Reg+Map (native) & 25.15 & 3.14 & 318.1 & 1452 & $4.40\times$ \\
8 & Reg+Map + SR + TC & 27.34 & 3.42 & 292.6 & 1452 & $4.05\times$ \\
8 & TCSD (native) & 24.27 & 3.03 & 329.6 & 1452 & $4.56\times$ \\
8 & TCSD + SR + TC & 25.81 & 3.23 & 309.9 & 1452 & $4.29\times$ \\
8 & HAT-SF & 110.65 & 13.83 & 72.3 & 1316 & $1.00\times$ \\
\bottomrule
\end{tabular}
\end{table*}

\paragraph{Readout overhead.}
Relative to native outputs, SR + TC changes batch latency by \(-0.05\) ms for Reg+Map and \(+2.28\) ms for TCSD at batch size 1. At batch size 8, the latency increase is \(+2.19\) ms for Reg+Map and \(+1.54\) ms for TCSD (Table~\ref{tab:supp_eval_efficiency}). Across all tested configurations, the maximum latency increase is roughly 2.3 ms. This is comparable to the inter-round variation of about 2 ms. Therefore, these measurements do not isolate the overhead of SR alone.

\paragraph{Scope of the speed comparison.}
These speed ratios reflect our specific implementations and hardware. They do not apply to the official HAT or to end-to-end latency from raw images and text.

\section{Readout selection and temperature}
\label{sec:supp_readout_selection}

In Section~3.3 of the main paper, we evaluated 14 candidate readouts on the COCO-Search18 development set. We ranked them by their mean rank over MultiMatch, ScanMatch, and SED, breaking ties in that metric order. We then fixed $\kappa=0.5$ for all subsequent experiments. Section~4.5 reports a temperature sweep on the development set. This section provides full results for both experiments and includes stability checks after locking.

\paragraph{Conventions.}
Unless stated otherwise, results in this section use Reg+Map on the COCO-TP development set with seeds 3 to 5. We use the default termination threshold \(\tau=0.5\) and no TC. Within each comparison, we change only the readout. Checkpoints, sequence lengths, and durations remain fixed. Section~\ref{sec:supp_sel_candidates} presents the locked rule results on the COCO-TP test set. Section~\ref{sec:supp_sel_stability} uses seeds 6 to 12, which were trained after locking. We formed the candidate set after observing decoding comparisons on the COCO-TP test set. However, we used only the development set to choose among the candidates. Therefore, as stated in the main paper, readout results on the COCO-TP test set are exploratory.

\subsection{The 14 candidates and the selection rule}
\label{sec:supp_sel_candidates}

\paragraph{Candidates.}
Let \(p\) denote the spatial distribution of one slot over the \(24\times32\) grid. Every readout outputs a weighted average of grid-cell centers.
\begin{itemize}
\item Plain expectation. We use \(p\) as the weights (SR, \(\kappa=1\)).
\item Argmax. We take the center of the most probable cell.
\item Top-5 expectation. We keep the 5 most probable cells, renormalize their weights, and compute the expectation.
\item Temperature expectation. We use renormalized weights \(p^{1/\kappa}\). The candidates include \(\kappa=0.5\).
\item Entropy-gated combinations. For each scanpath, we compute the mean entropy of its slot distributions normalized by \(\log 768\). Let the threshold be the \(q\) quantile of this value on the development set (\(q\in\{0.2, 0.35, 0.5, 0.65, 0.8\}\)). Scanpaths below the threshold use one readout, and the rest use the other. We test both directions. Low entropy can use plain expectation with high entropy using top-5, or the reverse.
\end{itemize}
This yields 14 candidates in total: four fixed readouts and \(5\times2\) gated combinations.

\paragraph{Selection rule.}
On the development set, we average each candidate over seeds 3 to 5. We compute the rank of each candidate for MultiMatch, ScanMatch, and SED separately, and then average these three ranks. We select the candidate with the lowest mean rank. We break ties by MultiMatch, then ScanMatch, then SED. We record the selected rule and the SHA-256 hashes of the development input files in the lock manifest.

\begin{table}[t]
\centering
\caption{Development results of the 14 candidate readouts (COCO-TP development set, mean over Reg+Map
seeds 3 to 5, $\tau=0.5$). ``Gate $q$, A $\rightarrow$ B'' uses A when the normalized mean entropy is
below the $q$ quantile and B otherwise. Exp.: plain expectation.}
\label{tab:supp_sel_candidates}
\footnotesize
\setlength{\tabcolsep}{4pt}
\begin{tabular}{@{}rlcccc@{}}
\toprule
Order & Candidate & Mean rank & MM $\uparrow$ & SM $\uparrow$ & SED $\downarrow$ \\
\midrule
1 (selected) & Temperature, $\kappa=0.5$ & 2.667 & 0.8337 & 0.4776 & 3.7151 \\
2  & Gate 0.2, Exp. $\rightarrow$ top-5   & 2.667 & 0.8318 & 0.4777 & 3.6906 \\
3  & Top-5 expectation                   & 2.667 & 0.8317 & 0.4810 & 3.6812 \\
4  & Gate 0.35, Exp. $\rightarrow$ top-5  & 4.333 & 0.8317 & 0.4748 & 3.7136 \\
5  & Gate 0.65, Exp. $\rightarrow$ top-5  & 6.667 & 0.8320 & 0.4657 & 3.7701 \\
6  & Gate 0.5, Exp. $\rightarrow$ top-5   & 6.667 & 0.8315 & 0.4711 & 3.7349 \\
7  & Gate 0.8, Exp. $\rightarrow$ top-5   & 7.667 & 0.8322 & 0.4591 & 3.8075 \\
8  & Argmax                              & 7.667 & 0.8224 & 0.4765 & 3.7321 \\
9  & Gate 0.8, top-5 $\rightarrow$ Exp.   & 8.667 & 0.8303 & 0.4728 & 3.7699 \\
10 & Gate 0.5, top-5 $\rightarrow$ Exp.   & 9.667 & 0.8310 & 0.4608 & 3.8425 \\
11 & Gate 0.65, top-5 $\rightarrow$ Exp.  & 9.667 & 0.8305 & 0.4662 & 3.8072 \\
12 & Gate 0.35, top-5 $\rightarrow$ Exp.  & 11.333 & 0.8308 & 0.4571 & 3.8638 \\
13 & Plain expectation ($\kappa=1$)      & 12.333 & 0.8308 & 0.4508 & 3.8962 \\
14 & Gate 0.2, top-5 $\rightarrow$ Exp.   & 12.333 & 0.8307 & 0.4542 & 3.8868 \\
\bottomrule
\end{tabular}
\end{table}

The first three candidates share a mean rank of 2.667 (Table~\ref{tab:supp_sel_candidates}). We break ties by MultiMatch first. The temperature expectation scores highest (0.8337 vs. 0.8318 and 0.8317), so we select it. All three perform similarly across metrics. Top-5 leads slightly on ScanMatch and SED, while the temperature expectation leads slightly on MultiMatch. Therefore, this selection follows the predefined tie-break rather than a clear development-set advantage. After locking, this rule achieves 0.8350 MultiMatch, 0.5013 ScanMatch, and 3.1634 SED on the COCO-TP test set (seeds 3-5, \(\tau=0.5\), exploratory).

\subsection{Temperature sweep}
\label{sec:supp_sel_sweep}

The initial selection compared only two temperatures (\(\kappa=0.5\) and \(\kappa=1\)). To verify whether the selected value lies within a performance plateau, we evaluated additional temperatures on the same development set. We conducted this sweep after locking \(\kappa\), and it did not alter the final selection.

\begin{table}[t]
\centering
\caption{Temperature sweep on the development set (COCO-TP development set, Reg+Map seeds 3 to 5, mean
$\pm$ standard deviation over seeds, $\tau=0.5$).}
\label{tab:supp_sel_sweep}
\footnotesize
\setlength{\tabcolsep}{5pt}
\begin{tabular}{@{}lccc@{}}
\toprule
Readout & MM $\uparrow$ & SM $\uparrow$ & SED $\downarrow$ \\
\midrule
Native regression & $0.8147\pm0.0035$ & $0.3526\pm0.0310$ & $4.307\pm0.543$ \\
Argmax & $0.8224\pm0.0014$ & $0.4765\pm0.0123$ & $3.732\pm0.526$ \\
$\kappa=0.25$ & $0.8328\pm0.0011$ & $0.4820\pm0.0103$ & $3.678\pm0.524$ \\
$\kappa=0.35$ & $0.8335\pm0.0013$ & $0.4811\pm0.0110$ & $3.689\pm0.513$ \\
$\kappa=0.5$ (locked) & $0.8337\pm0.0013$ & $0.4776\pm0.0110$ & $3.715\pm0.522$ \\
$\kappa=0.7$ & $0.8329\pm0.0013$ & $0.4705\pm0.0114$ & $3.785\pm0.521$ \\
$\kappa=1$ (plain expectation) & $0.8308\pm0.0014$ & $0.4508\pm0.0143$ & $3.896\pm0.548$ \\
$\kappa=1.5$ & $0.8266\pm0.0017$ & $0.4155\pm0.0137$ & $4.059\pm0.555$ \\
\bottomrule
\end{tabular}
\end{table}

\begin{figure*}[t]
\centering
\includegraphics[width=0.84\linewidth]{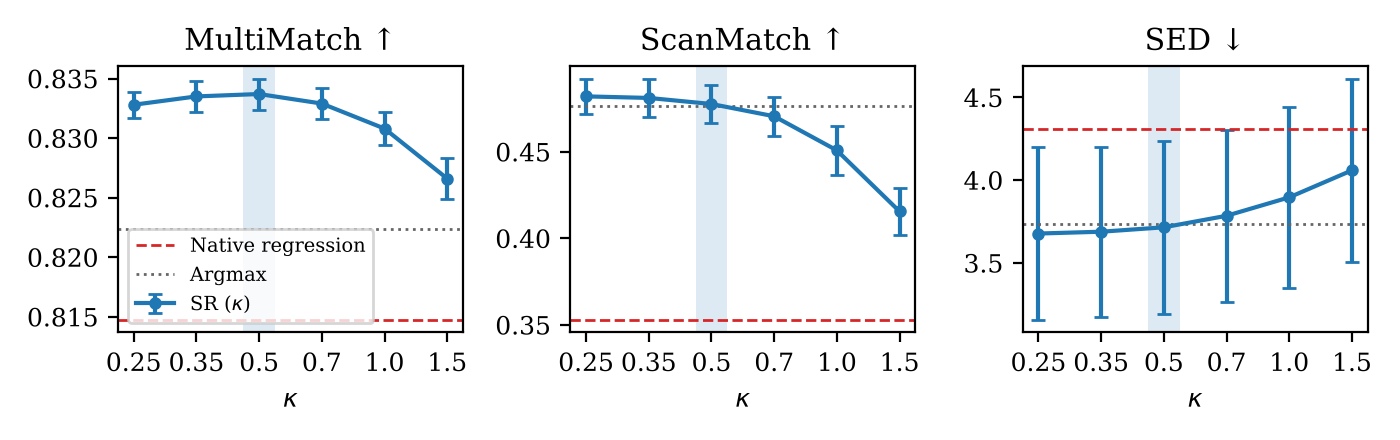}
\caption{Temperature sweep on the COCO-TP development set (Reg+Map, seeds 3 to 5, mean $\pm$ standard
deviation over seeds, $\tau=0.5$). The shaded band marks the locked $\kappa=0.5$. Dashed line: native
regression. Dotted line: argmax. The values are those of Table~\ref{tab:supp_sel_sweep}.}
\label{fig:supp_sel_sweep}
\end{figure*}

Results remain close across the four tested points (\(\kappa=0.25, 0.35, 0.5\), and \(0.7\)) (Table~\ref{tab:supp_sel_sweep} and Figure~\ref{fig:supp_sel_sweep}). ScanMatch ranges from 0.471 to 0.482, and MultiMatch from 0.833 to 0.834. Smaller \(\kappa\) values yield slightly better ScanMatch and SED, while MultiMatch peaks at \(\kappa=0.5\). At \(\kappa=1.5\), ScanMatch drops to 0.416. Every readout with \(\kappa\le1\), including argmax (ScanMatch 0.477), outperforms native regression (0.353) on all three metrics. The locked \(\kappa=0.5\) falls within this plateau, though it is not optimal on every metric. This sweep reports only means and standard deviations over seeds. It includes no paired tests, so it does not establish equivalence among these temperatures. These are the development-set scores summarized in Section~4.5 of the main paper.

\subsection{Stability checks after locking}
\label{sec:supp_sel_stability}

\paragraph{Leave-one-seed-out selection.}
We held out one seed at a time and applied the selection rule to the development results of the remaining two seeds. We also re-estimated the gate thresholds for each fold using only those two seeds. This assesses selection stability on the same development set rather than serving as an independent validation.

\begin{table}[t]
\centering
\caption{Leave-one-seed-out selection (development set). ``Best on held-out seed'' is the candidate that
the rule selects when it is run on the held-out seed alone.}
\label{tab:supp_sel_loso}
\footnotesize
\setlength{\tabcolsep}{4pt}
\begin{tabular}{@{}cllcl@{}}
\toprule
\makecell{Held-out\\seed} & \makecell[l]{Selected by the\\other two seeds} & \makecell[l]{Same as\\full lock} &
  \makecell{Mean rank on\\held-out seed} & \makecell[l]{Best on held-out seed\\(mean rank)} \\
\midrule
3 & $\kappa=0.5$ & Yes & 2.33 & Top-5 expectation (1.33) \\
4 & $\kappa=0.5$ & Yes & 2.67 & Gate 0.2, Exp. $\rightarrow$ top-5 (2.00) \\
5 & Top-5 expectation & No & 3.67 & $\kappa=0.5$ (2.00) \\
\bottomrule
\end{tabular}
\end{table}

Two of the three folds select the same \(\kappa=0.5\) as the full lock. The third fold selects top-5 (Table~\ref{tab:supp_sel_loso}). For the held-out seed, the best candidate also varies within this small subset. This aligns with the three-way tie in Table~\ref{tab:supp_sel_candidates}. The development set cannot reliably distinguish between temperature expectation, top-5 expectation, and the low-quantile gate. All three are readouts that sharpen the spatial distribution.

\paragraph{Seeds trained after locking.}
We trained Reg+Map seeds 6 to 12 after locking \(\kappa\), and they did not participate in any selection. On the COCO-TP test set (\(\tau=0.5\)), these seeds yield a ScanMatch gain of \(+0.0315\) \([0.0255, 0.0379]\) for sharpening over plain expectation. This gain is positive for all 7 seeds. The gain over all 10 seeds is \(+0.031\) \([0.026, 0.036]\) (Section~4.2). These results demonstrate robustness to training randomness. 

\paragraph{Transfer to other settings.}
We directly apply the locked \(\kappa=0.5\) to TCSD, Gazeformer-Aux, COCO-TA, and AiR-D. Results are reported in Sections~4.3 and~4.4 of the main paper, and in Sections~\ref{sec:supp_baselines}, \ref{sec:supp_replication}, and~\ref{sec:supp_readout_variants} of this supplement.

We did not optimize $\kappa$ separately for other models or datasets.

\section{Full result tables}
\label{sec:supp_full_results}

These tables provide expanded results for Tables~1 to~3 and the category breakdown in Section~4.5 of the main paper. Means and paired intervals are computed as described in Section~\ref{sec:supp_evaluation}. Note that the COCO-TP readout results are exploratory.

\subsection{Full results on COCO-TP}
\label{sec:supp_full_tp}

Table~\ref{tab:supp_full_tp} reports the main comparison together with the
additional baselines and the fixed-center reference.

\begin{table*}[t]
\centering
\caption{COCO-TP test set, mean $\pm$ sample standard deviation over seeds (full version of Table~1 of the
main paper). The SR rows reuse the base checkpoints and are exploratory. External baselines use their
native stopping rules. Gazeformer-Aux uses its own seeds 1 to 5. Latency is in
Section~\ref{sec:supp_eval_efficiency}.}
\label{tab:supp_full_tp}
\footnotesize
\setlength{\tabcolsep}{5pt}
\begin{tabular}{@{}lc cccccc@{}}
\toprule
Method & Seeds & MM $\uparrow$ & SM $\uparrow$ & SED $\downarrow$ & KLD $\downarrow$ & CC $\uparrow$ & NSS $\uparrow$ \\
\midrule
Reg & 5 & $0.781\pm0.006$ & $0.253\pm0.031$ & $4.60\pm1.26$ & $19.56\pm0.41$ & $0.094\pm0.018$ & $0.50\pm0.10$ \\
Reg+Map & 10 & $0.816\pm0.006$ & $0.364\pm0.028$ & $3.64\pm0.33$ & $16.66\pm1.20$ & $0.137\pm0.027$ & $0.74\pm0.17$ \\
Reg+Map + SR ($\kappa=1$) & 10 & $0.833\pm0.001$ & $0.473\pm0.013$ & $3.20\pm0.34$ & $11.43\pm0.55$ & $0.328\pm0.026$ & $2.05\pm0.19$ \\
Reg+Map + SR & 10 & $0.836\pm0.002$ & $0.504\pm0.009$ & $3.03\pm0.32$ & $9.50\pm0.21$ & $0.461\pm0.015$ & $3.03\pm0.13$ \\
Reg+Map + SR + TC & 10 & $0.837\pm0.002$ & $0.519\pm0.007$ & $2.57\pm0.13$ & $9.68\pm0.23$ & $0.458\pm0.018$ & $3.00\pm0.14$ \\
TCSD & 10 & $0.830\pm0.002$ & $0.456\pm0.016$ & $3.78\pm0.53$ & $11.32\pm0.40$ & $0.319\pm0.022$ & $1.98\pm0.18$ \\
TCSD + SR + TC & 10 & $0.837\pm0.002$ & $0.518\pm0.005$ & $2.55\pm0.12$ & $9.63\pm0.19$ & $0.455\pm0.020$ & $2.97\pm0.17$ \\
Gazeformer & 3 & $0.811\pm0.004$ & $0.416\pm0.006$ & $2.60\pm0.08$ & $12.98\pm0.30$ & $0.306\pm0.012$ & $2.14\pm0.08$ \\
Gazeformer-Aux (native) & 5 & $0.815\pm0.007$ & $0.416\pm0.012$ & $2.60\pm0.10$ & $12.77\pm0.20$ & $0.308\pm0.015$ & $2.13\pm0.13$ \\
HAT-SF & 3 & $0.832\pm0.003$ & $0.515\pm0.010$ & $2.12\pm0.04$ & $9.40\pm0.36$ & $0.530\pm0.027$ & $3.68\pm0.17$ \\
IRL & 3 & $0.802\pm0.001$ & $0.409\pm0.003$ & $3.01\pm0.03$ & $8.17\pm0.11$ & $0.460\pm0.005$ & $3.26\pm0.02$ \\
ScanDiff & 3 & $0.777\pm0.012$ & $0.364\pm0.004$ & $2.96\pm0.07$ & $6.90\pm0.25$ & $0.296\pm0.036$ & $1.64\pm0.25$ \\
\midrule
Fixed center & n/a & 0.767 & 0.246 & 2.91 & 17.94 & 0.341 & 2.70 \\
Human LOO & n/a & 0.847 & 0.539 & 2.13 & 6.79 & 0.470 & 2.98 \\
\bottomrule
\end{tabular}
\end{table*}

\paragraph{Regression-head control.}
Table~\ref{tab:supp_full_mlp} lists two models that replace the linear coordinate head with a two-layer MLP. We show only their native regression outputs. These models are excluded from Table~1 of the main paper.

\begin{table*}[t]
\centering
\caption{Regression-head control (COCO-TP test set, native regression output, mean $\pm$ standard
deviation over seeds).}
\label{tab:supp_full_mlp}
\footnotesize
\setlength{\tabcolsep}{5pt}
\begin{tabular}{@{}lc cccccc@{}}
\toprule
Model & Seeds & MM $\uparrow$ & SM $\uparrow$ & SED $\downarrow$ & KLD $\downarrow$ & CC $\uparrow$ & NSS $\uparrow$ \\
\midrule
Reg (MLP head) & 3 to 5 & $0.790\pm0.002$ & $0.280\pm0.016$ & $4.80\pm0.84$ & $17.17\pm0.41$ & $0.183\pm0.026$ & $1.05\pm0.15$ \\
Reg+Map (MLP head) & 3 to 7 & $0.823\pm0.003$ & $0.394\pm0.020$ & $3.92\pm0.80$ & $14.18\pm0.46$ & $0.219\pm0.021$ & $1.31\pm0.17$ \\
\bottomrule
\end{tabular}
\end{table*}

Reg+Map (MLP head) corresponds to Section~4.2 of the main paper. The main paper compares SR (\(\kappa=1\)) on the linear-head Reg+Map against native regression on the MLP-head Reg+Map. This yields a ScanMatch difference of \(+0.075\) \([0.055, 0.101]\) (seeds 3 to 7). This comparison does not read the spatial map of the MLP checkpoint. Reg (MLP head) serves as a reference only.

\subsection{Full results on COCO-TA and AiR-D}
\label{sec:supp_full_transfer}

Tables~\ref{tab:supp_full_ta} and~\ref{tab:supp_full_aird} provide extended versions of the transfer results from Table~2 of the main paper. Table~\ref{tab:supp_full_tcsd} presents the direct comparison of training pathways.

\begin{table*}[t]
\centering
\caption{COCO-TA test set, mean $\pm$ sample standard deviation over seeds (full version of Table~2 of the
main paper). All six metrics are first averaged within each image, then across images, and then over
seeds. The HAT-SF + TC row is re-aggregated from its score files in the same way. Length is the mean
decoded length (human mean on the development set 5.6).}
\label{tab:supp_full_ta}
\footnotesize
\setlength{\tabcolsep}{4pt}
\begin{tabular}{@{}lc cccccc c@{}}
\toprule
Method & Seeds & MM $\uparrow$ & SM $\uparrow$ & SED $\downarrow$ & KLD $\downarrow$ & CC $\uparrow$ & NSS $\uparrow$ & Length \\
\midrule
Reg+Map & 5 & $0.799\pm0.005$ & $0.137\pm0.016$ & $13.93\pm0.50$ & $16.86\pm0.65$ & $0.155\pm0.027$ & $0.56\pm0.11$ & 15.46 \\
Reg+Map + SR & 5 & $0.800\pm0.007$ & $0.165\pm0.014$ & $13.35\pm0.42$ & $14.35\pm0.54$ & $0.163\pm0.030$ & $0.64\pm0.12$ & 15.46 \\
Reg+Map + SR + TC & 5 & $0.815\pm0.003$ & $0.320\pm0.007$ & $6.14\pm0.21$ & $15.09\pm0.35$ & $0.262\pm0.011$ & $1.09\pm0.05$ & 6.13 \\
TCSD & 5 & $0.798\pm0.003$ & $0.157\pm0.015$ & $13.36\pm0.60$ & $15.58\pm0.15$ & $0.160\pm0.025$ & $0.62\pm0.10$ & 15.18 \\
TCSD + SR & 5 & $0.803\pm0.003$ & $0.172\pm0.015$ & $13.11\pm0.58$ & $14.29\pm0.20$ & $0.179\pm0.022$ & $0.70\pm0.10$ & 15.18 \\
TCSD + SR + TC & 5 & $0.815\pm0.001$ & $0.317\pm0.005$ & $6.19\pm0.24$ & $15.03\pm0.12$ & $0.266\pm0.008$ & $1.11\pm0.04$ & 6.21 \\
HAT-SF & 3 & $0.821\pm0.002$ & $0.328\pm0.007$ & $5.41\pm0.18$ & $11.94\pm0.16$ & $0.362\pm0.007$ & $1.61\pm0.03$ & 6.45 \\
HAT-SF + TC & 3 & $0.820\pm0.002$ & $0.343\pm0.003$ & $5.01\pm0.02$ & $12.42\pm0.05$ & $0.370\pm0.007$ & $1.67\pm0.02$ & 5.67 \\
\bottomrule
\end{tabular}
\end{table*}

With TC alone and no SR, Reg+Map reaches a ScanMatch of 0.275 (Section~4.5 of the main paper). The setup of
HAT-SF + TC is in Section~\ref{sec:supp_base_hat}. An undetected difference does not imply equivalence (Section~\ref{sec:supp_equivalence}).

\begin{table*}[t]
\centering
\caption{AiR-D test set, mean $\pm$ standard deviation over seeds (seeds 3 to 5, full version of Table~2
of the main paper). Length is the mean decoded length (human mean on the development set 10.1).}
\label{tab:supp_full_aird}
\footnotesize
\setlength{\tabcolsep}{4pt}
\begin{tabular}{@{}l cccccc c@{}}
\toprule
Method & MM $\uparrow$ & SM $\uparrow$ & SED $\downarrow$ & KLD $\downarrow$ & CC $\uparrow$ & NSS $\uparrow$ & Length \\
\midrule
Reg+Map & $0.786\pm0.003$ & $0.252\pm0.006$ & $10.13\pm0.15$ & $10.18\pm0.38$ & $0.366\pm0.017$ & $0.96\pm0.08$ & 12.46 \\
Reg+Map + SR & $0.789\pm0.001$ & $0.292\pm0.005$ & $9.57\pm0.11$ & $7.73\pm0.33$ & $0.447\pm0.013$ & $1.25\pm0.04$ & 12.46 \\
Reg+Map + TC & $0.789\pm0.004$ & $0.280\pm0.007$ & $8.54\pm0.09$ & $10.24\pm0.38$ & $0.367\pm0.018$ & $0.97\pm0.08$ & 10.08 \\
Reg+Map + SR + TC & $0.793\pm0.001$ & $0.327\pm0.006$ & $7.97\pm0.05$ & $7.70\pm0.35$ & $0.456\pm0.011$ & $1.27\pm0.04$ & 10.08 \\
TCSD & $0.780\pm0.001$ & $0.268\pm0.006$ & $10.08\pm0.68$ & $8.72\pm0.36$ & $0.413\pm0.014$ & $1.14\pm0.04$ & 12.80 \\
TCSD + SR & $0.785\pm0.001$ & $0.282\pm0.008$ & $9.92\pm0.66$ & $7.68\pm0.38$ & $0.433\pm0.015$ & $1.21\pm0.04$ & 12.80 \\
TCSD + TC & $0.786\pm0.003$ & $0.307\pm0.008$ & $8.15\pm0.04$ & $8.70\pm0.36$ & $0.421\pm0.015$ & $1.16\pm0.05$ & 10.11 \\
TCSD + SR + TC & $0.792\pm0.005$ & $0.323\pm0.006$ & $8.00\pm0.05$ & $7.66\pm0.36$ & $0.443\pm0.017$ & $1.23\pm0.04$ & 10.11 \\
\bottomrule
\end{tabular}
\end{table*}

\begin{table*}[t]
\centering
\caption{TCSD + SR + TC minus Reg+Map + SR + TC (paired difference with 95\% interval). Section~4.3 of the
main paper states that the intervals of all six metrics contain zero in both transfer settings.}
\label{tab:supp_full_tcsd}
\footnotesize
\setlength{\tabcolsep}{4pt}
\renewcommand{\arraystretch}{1.1}
\begin{tabular}{@{}l cccccc@{}}
\toprule
Setting & MM & SM & SED & KLD & CC & NSS \\
\midrule
COCO-TA (seeds 3 to 7)
  & \makecell{$-0.000^\dagger$\\{\scriptsize$[-0.003,+0.002]$}}
  & \makecell{$-0.003^\dagger$\\{\scriptsize$[-0.013,+0.007]$}}
  & \makecell{$+0.059^\dagger$\\{\scriptsize$[-0.305,+0.403]$}}
  & \makecell{$-0.066^\dagger$\\{\scriptsize$[-0.363,+0.242]$}}
  & \makecell{$+0.004^\dagger$\\{\scriptsize$[-0.012,+0.019]$}}
  & \makecell{$+0.020^\dagger$\\{\scriptsize$[-0.054,+0.098]$}} \\
\addlinespace[2pt]
AiR-D (seeds 3 to 5)
  & \makecell{$-0.001^\dagger$\\{\scriptsize$[-0.006,+0.003]$}}
  & \makecell{$-0.004^\dagger$\\{\scriptsize$[-0.012,+0.005]$}}
  & \makecell{$+0.002^\dagger$\\{\scriptsize$[-0.106,+0.112]$}}
  & \makecell{$-0.009^\dagger$\\{\scriptsize$[-0.565,+0.510]$}}
  & \makecell{$-0.012^\dagger$\\{\scriptsize$[-0.035,+0.010]$}}
  & \makecell{$-0.030^\dagger$\\{\scriptsize$[-0.103,+0.037]$}} \\
\bottomrule
\end{tabular}
\end{table*}

\subsection{Contributions of the two components}
\label{sec:supp_full_components}

\begin{table*}[t]
\centering
\caption{Paired effects of the two components on Reg+Map (full version of Table~3 of the main paper).
``+ SR'' is SR minus native regression at $\tau=0.5$. ``+ TC'' is SR + TC minus SR. On COCO-TP, ``+ SR''
uses seeds 3 to 5 and ``+ TC'' uses 10 seeds, so the two are not additive. COCO-TA uses 5 seeds and AiR-D
uses 3 seeds.}
\label{tab:supp_full_components}
\footnotesize
\setlength{\tabcolsep}{4pt}
\renewcommand{\arraystretch}{1.1}
\begin{tabular}{@{}ll cccccc@{}}
\toprule
Setting & Step & MM & SM & SED & KLD & CC & NSS \\
\midrule
COCO-TP & + SR
  & \makecell{$+0.019^{\checkmark}$\\{\scriptsize$[+0.014,+0.025]$}}
  & \makecell{$+0.136^{\checkmark}$\\{\scriptsize$[+0.108,+0.171]$}}
  & \makecell{$-0.63^{\checkmark}$\\{\scriptsize$[-0.79,-0.48]$}}
  & \makecell{$-7.28^{\checkmark}$\\{\scriptsize$[-8.84,-5.91]$}}
  & \makecell{$+0.329^{\checkmark}$\\{\scriptsize$[+0.286,+0.370]$}}
  & \makecell{$+2.34^{\checkmark}$\\{\scriptsize$[+2.06,+2.61]$}} \\
COCO-TP & + TC
  & \makecell{$+0.001^\dagger$\\{\scriptsize$[-0.001,+0.002]$}}
  & \makecell{$+0.015^{\checkmark}$\\{\scriptsize$[+0.008,+0.022]$}}
  & \makecell{$-0.46^{\checkmark}$\\{\scriptsize$[-0.71,-0.25]$}}
  & \makecell{$+0.18^{\times}$\\{\scriptsize$[+0.10,+0.28]$}}
  & \makecell{$-0.002^\dagger$\\{\scriptsize$[-0.006,+0.001]$}}
  & \makecell{$-0.02^\dagger$\\{\scriptsize$[-0.05,+0.00]$}} \\
\addlinespace[2pt]
COCO-TA & + SR
  & \makecell{$+0.001^\dagger$\\{\scriptsize$[-0.002,+0.004]$}}
  & \makecell{$+0.029^{\checkmark}$\\{\scriptsize$[+0.025,+0.033]$}}
  & \makecell{$-0.58^{\checkmark}$\\{\scriptsize$[-0.70,-0.47]$}}
  & \makecell{$-2.51^{\checkmark}$\\{\scriptsize$[-2.91,-2.18]$}}
  & \makecell{$+0.008^\dagger$\\{\scriptsize$[-0.006,+0.022]$}}
  & \makecell{$+0.08^{\checkmark}$\\{\scriptsize$[+0.02,+0.14]$}} \\
COCO-TA & + TC
  & \makecell{$+0.014^{\checkmark}$\\{\scriptsize$[+0.010,+0.019]$}}
  & \makecell{$+0.155^{\checkmark}$\\{\scriptsize$[+0.142,+0.169]$}}
  & \makecell{$-7.21^{\checkmark}$\\{\scriptsize$[-7.74,-6.67]$}}
  & \makecell{$+0.74^{\times}$\\{\scriptsize$[+0.55,+0.94]$}}
  & \makecell{$+0.099^{\checkmark}$\\{\scriptsize$[+0.079,+0.119]$}}
  & \makecell{$+0.46^{\checkmark}$\\{\scriptsize$[+0.36,+0.55]$}} \\
\addlinespace[2pt]
AiR-D & + SR
  & \makecell{$+0.002^\dagger$\\{\scriptsize$[-0.001,+0.006]$}}
  & \makecell{$+0.037^{\checkmark}$\\{\scriptsize$[+0.027,+0.049]$}}
  & \makecell{$-0.52^{\checkmark}$\\{\scriptsize$[-0.73,-0.33]$}}
  & \makecell{$-2.50^{\checkmark}$\\{\scriptsize$[-3.11,-1.93]$}}
  & \makecell{$+0.076^{\checkmark}$\\{\scriptsize$[+0.045,+0.108]$}}
  & \makecell{$+0.25^{\checkmark}$\\{\scriptsize$[+0.13,+0.38]$}} \\
AiR-D & + TC
  & \makecell{$+0.004^{\checkmark}$\\{\scriptsize$[+0.002,+0.006]$}}
  & \makecell{$+0.034^{\checkmark}$\\{\scriptsize$[+0.030,+0.039]$}}
  & \makecell{$-1.61^{\checkmark}$\\{\scriptsize$[-1.80,-1.44]$}}
  & \makecell{$-0.03^{\checkmark}$\\{\scriptsize$[-0.05,-0.01]$}}
  & \makecell{$+0.009^{\checkmark}$\\{\scriptsize$[+0.006,+0.012]$}}
  & \makecell{$+0.02^{\checkmark}$\\{\scriptsize$[+0.01,+0.03]$}} \\
\bottomrule
\end{tabular}
\end{table*}

Table~\ref{tab:supp_full_components} extends Table~3 of the main paper
by reporting MultiMatch and NSS. The ``+ SR'' row compares spatial
readout with native regression at the default termination threshold.
SR improves NSS in all three settings, but a MultiMatch improvement
is detectable only on COCO-TP. The ``+ TC'' row compares SR + TC
with SR at the default threshold, measuring the incremental effect
of termination calibration while holding the spatial readout fixed.
Adding TC improves both MultiMatch and NSS on COCO-TA and AiR-D;
on COCO-TP, we detect no difference in either metric.

\subsection{Breakdown by target category}
\label{sec:supp_full_category}

Section~4.5 of the main paper shows that Reg+Map + SR + TC yields positive ScanMatch gains over native regression across all 18 target categories. Table~\ref{tab:supp_full_category} provides the full results.

\begin{table*}[t]
\centering
\caption{Gain by target category: Reg+Map + SR + TC minus native regression (COCO-TP test set, seeds 3 to
5). Within each category, seeds and physical images are resampled jointly with 10,000 replicates. Each
95\% interval is reported separately, without correction for multiple comparisons. SED is a raw
difference, so a negative value means that SR + TC is better. Rows are sorted by the ScanMatch gain.}
\label{tab:supp_full_category}
\footnotesize
\setlength{\tabcolsep}{6pt}
\begin{tabular}{@{}lc ccc@{}}
\toprule
Category & Images & ScanMatch & MultiMatch & SED \\
\midrule
fork & 23 & $+0.234$ $[+0.177,+0.301]$ & $+0.027$ $[+0.006,+0.046]$ & $-1.74$ $[-3.27,-0.77]$ \\
microwave & 16 & $+0.191$ $[+0.143,+0.239]$ & $+0.030$ $[+0.016,+0.042]$ & $-1.22$ $[-2.27,-0.71]$ \\
potted plant & 17 & $+0.183$ $[+0.119,+0.252]$ & $+0.021$ $[+0.011,+0.032]$ & $-2.31$ $[-4.22,-0.99]$ \\
car & 12 & $+0.174$ $[+0.084,+0.272]$ & $+0.015^\dagger$ $[-0.005,+0.033]$ & $-1.17$ $[-2.35,-0.46]$ \\
clock & 13 & $+0.169$ $[+0.100,+0.243]$ & $+0.028$ $[+0.019,+0.040]$ & $-0.60$ $[-0.83,-0.36]$ \\
stop sign & 13 & $+0.163$ $[+0.105,+0.219]$ & $+0.027$ $[+0.012,+0.045]$ & $-0.83$ $[-2.01,-0.22]$ \\
tv & 29 & $+0.163$ $[+0.118,+0.212]$ & $+0.020$ $[+0.010,+0.030]$ & $-1.43$ $[-2.31,-0.80]$ \\
cup & 28 & $+0.162$ $[+0.112,+0.226]$ & $+0.023$ $[+0.011,+0.037]$ & $-1.42$ $[-2.94,-0.55]$ \\
laptop & 13 & $+0.162$ $[+0.107,+0.231]$ & $+0.017$ $[+0.008,+0.025]$ & $-1.58$ $[-4.44,-0.22]$ \\
oven & 11 & $+0.161$ $[+0.097,+0.234]$ & $+0.024$ $[+0.005,+0.045]$ & $-1.97$ $[-4.07,-0.66]$ \\
bottle & 17 & $+0.161$ $[+0.094,+0.236]$ & $+0.022$ $[+0.011,+0.035]$ & $-2.10$ $[-4.08,-0.68]$ \\
mouse & 12 & $+0.157$ $[+0.098,+0.219]$ & $+0.010^\dagger$ $[-0.008,+0.025]$ & $-0.90$ $[-2.23,-0.29]$ \\
toilet & 17 & $+0.155$ $[+0.069,+0.251]$ & $+0.019$ $[+0.007,+0.033]$ & $-0.96$ $[-2.18,-0.22]$ \\
knife & 15 & $+0.150$ $[+0.091,+0.220]$ & $+0.017$ $[+0.001,+0.029]$ & $-0.76$ $[-1.07,-0.46]$ \\
chair & 26 & $+0.128$ $[+0.070,+0.195]$ & $+0.019$ $[+0.007,+0.034]$ & $-1.14$ $[-2.29,-0.44]$ \\
sink & 29 & $+0.125$ $[+0.069,+0.183]$ & $+0.019^\dagger$ $[-0.002,+0.038]$ & $-0.43$ $[-0.66,-0.17]$ \\
bowl & 15 & $+0.118$ $[+0.050,+0.197]$ & $+0.013^\dagger$ $[-0.003,+0.028]$ & $-2.89$ $[-5.40,-0.91]$ \\
keyboard & 20 & $+0.085$ $[+0.056,+0.117]$ & $+0.014$ $[+0.003,+0.025]$ & $-0.43$ $[-1.16,-0.11]$ \\
\bottomrule
\end{tabular}
\end{table*}

ScanMatch and SED intervals exclude zero in favor of SR + TC across all 18 categories. For MultiMatch, 14 categories exclude zero, and we detect no difference in the remaining four (car, mouse, sink, and bowl). These results indicate a consistent direction across the 18 tested categories, though they do not establish generalization to new ones. Each category contains only 11 to 29 images, so we do not compare gain magnitudes between them. Section~\ref{sec:supp_qualitative} provides the source for the main paper's claim that 92\% of the groups improve.


\section{External baselines}
\label{sec:supp_baselines}

This section provides extended results for the baseline comparisons in Sections~4.1 to~4.4 of the main paper. Between-model comparisons use the common seeds 3 to 5. Same-checkpoint readout comparisons for Gazeformer-Aux use seeds 1 to 5. We report IRL and ScanDiff descriptively. Paired intervals elsewhere are computed as described in Section~\ref{sec:supp_eval_stats}.

\subsection{Training and evaluation details}
\label{sec:supp_base_training}

Within each setting, all models share the same data splits and evaluator. Tables~\ref{tab:supp_base_protocol} and~\ref{tab:supp_base_convergence} detail the training settings and checkpoint selection for COCO-TP. IRL employs DCB belief maps rather than the shared grid.

\begin{table*}[t]
\centering
\caption{Training protocol on COCO-TP. Parameter counts are trainable parameters, taken from the
inference benchmark (RTX 4060 Laptop). ``Shared grid'' is the frozen Mask R-CNN R50-FPN layer4 feature
grid of size $24\times32$.}
\label{tab:supp_base_protocol}
\footnotesize
\setlength{\tabcolsep}{4pt}
\renewcommand{\arraystretch}{1.15}
\begin{tabular}{@{}P{1.75cm}P{2.75cm}P{2.75cm}P{2.75cm}P{2.3cm}P{2.3cm}@{}}
\toprule
& Reg+Map / TCSD & Gazeformer & HAT-SF & ScanDiff & IRL \\
\midrule
Code & This work & Official, \texttt{d26a40a} & Official, \texttt{557205d}, shared-feature adaptation
  & Official, \texttt{9eff26c}, shared-feature adaptation & Official, \texttt{36efbcc} \\
Visual input & Shared grid & Shared grid (official 6+6-layer Transformer, Gaussian coordinate and duration heads)
  & Shared grid, replaces Detectron2 and the MSDeformAttn pixel decoder & Shared grid, replaces DINOv2
  & Official 134-channel DCB belief maps \\
Decoding & Parallel, 16 slots & Parallel, 7 slots & Autoregressive, at most 7, starts at the center
  & DDIM-50, 16 & Policy roll-out, at most 6 \\
Parameters & 4.6M (of 128.9M, RoBERTa frozen) & 27.7M & 14.3M & 37.8M & 1.1M \\
Optimizer & AdamW, 1e-4, wd 5e-5, clip 1.0, fp16 & Official per-head rates 1e-6 / 1e-4 (coordinates) /
  2e-6 (duration), clip 1.0, fp16 & AdamW, 1e-4, wd 5e-5, clip 1.0, fp16 & AdamW, 1e-4, wd 1e-2, clip 1.0, fp16
  & GAIL 5e-5 / PPO 1e-5 (official) \\
Batch & 16 & $8\times2$ (accumulation) & $8\times2$ (accumulation) & 16 & 128 \\
Budget & At most 30 epochs & 30 epochs & At most 30k steps, evaluated every 5k & At most 30 epochs & 30 epochs \\
Selection & Combined development loss, patience 8 epochs & Native total development loss,
  patience 8 & Native total development loss, patience 4 evaluations & Native total development loss,
  patience 8 & Development probability mismatch \\
\bottomrule
\end{tabular}
\end{table*}

\paragraph{Seeds.}
All external baselines use seeds 3, 4, and 5. Cross-model paired comparisons use only these seeds. Reg+Map and TCSD use 10 seeds. Gazeformer-Aux uses seeds 1 to 5 and serves only for the same-checkpoint readout comparison.

\begin{table*}[t]
\centering
\caption{Training runs on the development set (best checkpoint and stopping point for each seed).}
\label{tab:supp_base_convergence}
\footnotesize
\setlength{\tabcolsep}{4pt}
\renewcommand{\arraystretch}{1.15}
\begin{tabular}{@{}lP{2.2cm}P{2.2cm}P{2.2cm}P{6.6cm}@{}}
\toprule
Method & Seed 3 & Seed 4 & Seed 5 & Note \\
\midrule
Reg+Map & Best epoch 8, stopped at 17 & Best 9, stopped at 18 & Best 5, stopped at 14
  & Early stopping triggered. The best checkpoint lies before the end of the budget. \\
TCSD & Best 8, stopped at 17 & Best 11, stopped at 20 & Best 8, stopped at 17
  & Early stopping triggered. The best checkpoint lies before the end of the budget. \\
HAT-SF & Best 10k, stopped at 30k & Best 5k, stopped at 25k & Best 5k, stopped at 25k
  & The development loss rises after 5k to 10k steps while the training loss keeps falling. We keep the
  early checkpoint. \\
ScanDiff & Best epoch 27 of 30 & Best 23 of 30 & Best 10, stopped at 19
  & The best checkpoint lies before the end of the budget. \\
IRL & Best epoch 24 of 30 & Best 23 of 30 & Best 22 of 30
  & The best checkpoint lies before the end of the budget. \\
Gazeformer & Best epoch 29 of 30 & Best 29 of 30 & Best 29 of 30
  & Limited by the budget. The total development loss still falls by 8\% to 10\% over the last 5 epochs.
  The coordinate term is flat ($\pm3\%$) from about epoch 24. The remaining decrease comes from the
  duration term. \\
\bottomrule
\end{tabular}
\end{table*}

Gazeformer-Aux (seeds 1 to 5) exhibits the same behavior. Its coordinate loss is flat, while the duration loss still decreases at epoch 30. Because our readout comparisons focus solely on fixation locations, the plateaued location loss is sufficient. Longer training might only marginally improve durations.

\paragraph{Adaptations.}
\begin{itemize}
\item HAT-SF. We replace HAT's~\cite{yang2024unifying} Detectron2 ResNet/FPN and MSDeformAttn pixel decoder with the shared frozen feature grid. We retain the sequential task-query decoding, fixation history memory, dense spatial map, termination head, 7-step native horizon, and modified focal loss. Since HAT does not predict durations, we assign each fixation the training-set median duration during export.
\item ScanDiff~\cite{cartella2025modeling}. We replace DINOv2 with the shared grid and retain the official DiT and diffusion objectives.
\item Gazeformer~\cite{mondal2023gazeformer}. We retain the official architecture and losses. We encode task text using the frozen RoBERTa CLS token, as the released code expects precomputed language features.
\item IRL~\cite{yang2020predicting}. We use the official DCBs and GAIL/PPO. We only add a uint8-to-bool mask conversion to support PyTorch 2.
\end{itemize}

\paragraph{Relation to published numbers.}
Published scores may rely on different splits, length limits, aggregation methods, or metric definitions. For example, HAT reports SemSS, SS, cIG, cNSS, and cAUC~\cite{yang2024unifying}. GazeXplain~\cite{chen2024gazexplain} samples fixation coordinates from its spatial maps. Therefore, it does not serve as a direct retraining-fidelity reference for Gazeformer. In contrast, our baseline comparisons use a shared evaluation protocol.

\subsection{IRL and ScanDiff on COCO-TP}
\label{sec:supp_base_full}

Table~\ref{tab:supp_full_tp} includes IRL and ScanDiff. Both underperform HAT-SF and Reg+Map + SR + TC on sequence metrics but yield lower KLD. We did not test these comparisons in pairs. ScanDiff lacks the per-fixation grid distribution required by SR. We did not apply SR to IRL (Section~3.5 of the main paper).

\subsection{First-Fixation Handling in the HAT-SF Comparison}
\label{sec:supp_base_first}

\paragraph{Why the first fixation matters.}
HAT-SF follows the official setting and starts every scanpath at the image center. Consequently, its first fixation aligns exactly with the center on the test set. In contrast, the first recorded human fixation lies 80.4 pixels from the center on average. Our parallel models predict this first fixation directly. Specifically, it lies 36.2 pixels from the center for Reg+Map + SR and 112.7 pixels for native regression. The mean length of HAT-SF is only 2.93, which drops to 1.93 after removing the first fixation. Therefore, the first fixation has a disproportionate impact on HAT-SF, and keeping or removing it changes the sequence metrics. We evaluate two protocols. The default protocol retains the first fixation, following the main paper. The alternative protocol removes it from all predicted and human sequences. Neither protocol requires retraining or regenerating predictions.

\begin{table}[t]
\centering
\caption{Means under the two first-fixation protocols (COCO-TP test set). Reg+Map rows use 10 seeds, and
HAT-SF uses 3 seeds.}
\label{tab:supp_base_first_means}
\footnotesize
\setlength{\tabcolsep}{4pt}
\begin{tabular}{@{}ll cccccc@{}}
\toprule
Protocol & Method & MM $\uparrow$ & SM $\uparrow$ & SED $\downarrow$ & KLD $\downarrow$ & CC $\uparrow$ & NSS $\uparrow$ \\
\midrule
Keep first & Fixed center      & 0.767 & 0.246 & 2.909 & 17.941 & 0.341 & 2.696 \\
Keep first & Human LOO         & 0.847 & 0.539 & 2.126 & 6.793  & 0.470 & 2.977 \\
Keep first & HAT-SF            & 0.832 & 0.515 & 2.116 & 9.396  & 0.530 & 3.681 \\
Keep first & Reg+Map + SR + TC & 0.837 & 0.519 & 2.569 & 9.676  & 0.458 & 3.005 \\
\midrule
Remove first & Fixed center      & 0.705 & 0.274 & 2.711 & 22.516 & 0.016 & 0.082 \\
Remove first & Human LOO         & 0.766 & 0.539 & 2.042 & 8.247  & 0.414 & 3.374 \\
Remove first & HAT-SF            & 0.728 & 0.552 & 2.052 & 10.698 & 0.471 & 3.726 \\
Remove first & Reg+Map + SR + TC & 0.754 & 0.522 & 2.482 & 11.690 & 0.407 & 3.340 \\
\bottomrule
\end{tabular}
\end{table}

\begin{table*}[t]
\centering
\caption{Reg+Map + SR + TC minus HAT-SF on COCO-TP (seeds 3 to 5). Paired difference with 95\%
interval.}
\label{tab:supp_base_first_paired}
\footnotesize
\setlength{\tabcolsep}{4pt}
\renewcommand{\arraystretch}{1.1}
\begin{tabular}{@{}l cccccc@{}}
\toprule
Protocol & MM & SM & SED & KLD & CC & NSS \\
\midrule
Keep first
  & \makecell{$+0.005^\dagger$\\{\scriptsize$[-0.000,+0.010]$}}
  & \makecell{$+0.006^\dagger$\\{\scriptsize$[-0.008,+0.021]$}}
  & \makecell{$+0.375^{\times}$\\{\scriptsize$[+0.183,+0.590]$}}
  & \makecell{$+0.361^\dagger$\\{\scriptsize$[-0.280,+1.014]$}}
  & \makecell{$-0.073^{\times}$\\{\scriptsize$[-0.112,-0.031]$}}
  & \makecell{$-0.664^{\times}$\\{\scriptsize$[-0.937,-0.354]$}} \\
\addlinespace[2pt]
Remove first
  & \makecell{$+0.024^{\checkmark}$\\{\scriptsize$[+0.008,+0.040]$}}
  & \makecell{$-0.023^{\times}$\\{\scriptsize$[-0.037,-0.007]$}}
  & \makecell{$+0.352^{\times}$\\{\scriptsize$[+0.160,+0.563]$}}
  & \makecell{$+1.056^{\times}$\\{\scriptsize$[+0.085,+2.062]$}}
  & \makecell{$-0.067^{\times}$\\{\scriptsize$[-0.136,-0.001]$}}
  & \makecell{$-0.406^\dagger$\\{\scriptsize$[-1.026,+0.211]$}} \\
\bottomrule
\end{tabular}
\end{table*}

When the first fixation is kept, we detect no difference in MultiMatch, ScanMatch, or KLD. However, HAT-SF performs better on SED, CC, and NSS (Table~\ref{tab:supp_base_first_paired}). When the first fixation is removed, HAT-SF leads on ScanMatch, while we lead on MultiMatch. KLD also favors HAT-SF, and we detect no difference in NSS. These findings align with Section~4.2 of the main paper. For the fixed-center reference, removing the first fixation raises ScanMatch from 0.246 to 0.274, but drops CC from 0.341 to 0.016 (Table~\ref{tab:supp_base_first_means}). Center bias thus affects these two types of metrics in opposite directions.

\paragraph{Early target hits.}
Table~\ref{tab:supp_base_hits} reports the Hit@\(k\) metric, which indicates whether any of the first \(k\) fixations fall inside the target box. These results use seeds 3 to 5 and exclude TC. When retaining the first fixation, Hit@1 is close to zero for both HAT-SF and our models. This occurs because the first fixation always lies at or near the center.

\begin{table}[t]
\centering
\caption{Target hits on the COCO-TP test set (seeds 3 to 5, without TC). ``First hit'' is the index of
the first hit, capped at the limit when there is no hit. SR without $\kappa$ means $\kappa=0.5$.}
\label{tab:supp_base_hits}
\footnotesize
\setlength{\tabcolsep}{4pt}
\begin{tabular}{@{}ll ccccc@{}}
\toprule
Protocol & Method & Hit@1 & Hit@3 & Hit@6 & Hit AUC & First hit \\
\midrule
Keep first & Human                         & 0.008 & 0.721 & 0.875 & 0.618 & 3.29 \\
Keep first & HAT-SF                        & 0.000 & 0.892 & 0.920 & 0.721 & 2.68 \\
Keep first & Reg+Map (native)              & 0.011 & 0.237 & 0.244 & 0.184 & 5.89 \\
Keep first & Reg+Map + SR ($\kappa=1$)     & 0.000 & 0.550 & 0.576 & 0.421 & 4.48 \\
Keep first & Reg+Map + SR                  & 0.000 & 0.758 & 0.782 & 0.597 & 3.42 \\
Keep first & TCSD (native)                 & 0.000 & 0.576 & 0.603 & 0.442 & 4.35 \\
\midrule
Remove first & Human                       & 0.398 & 0.832 & 0.879 & 0.762 & 2.43 \\
Remove first & HAT-SF                      & 0.680 & 0.913 & 0.921 & 0.874 & 1.75 \\
Remove first & Reg+Map (native)            & 0.127 & 0.239 & 0.243 & 0.221 & 5.67 \\
Remove first & Reg+Map + SR ($\kappa=1$)   & 0.250 & 0.574 & 0.576 & 0.517 & 3.90 \\
Remove first & Reg+Map + SR                & 0.482 & 0.778 & 0.782 & 0.728 & 2.64 \\
Remove first & TCSD (native)               & 0.271 & 0.601 & 0.603 & 0.543 & 3.75 \\
\bottomrule
\end{tabular}
\end{table}

Under both protocols, the readout affects Hit@3 and Hit@6 consistently. Native regression ranks lowest, plain expectation is intermediate, and sharpened expectation is highest. Under both protocols, HAT-SF achieves higher Hit@3 and Hit@6 than all parallel readouts and humans. Hit@1 behaves differently. When keeping the first fixation, HAT-SF scores 0.000 on Hit@1. This falls below humans (0.008) and native regression (0.011) because HAT-SF fixes its first fixation at the image center.

\subsection{HAT-SF with the same termination calibration and spatial readout}
\label{sec:supp_base_hat}

\paragraph{Setup.}
HAT-SF uses its own termination head. We apply the same TC rule as the main paper. We select a threshold on the development set to minimize the difference between predicted and human mean lengths. HAT-SF has a maximum length limit of 7 fixations. This constitutes a post-hoc analysis.

\begin{table}[t]
\centering
\caption{TC thresholds and lengths of HAT-SF (seeds 3 to 5). Ranges are over seeds.}
\label{tab:supp_base_hat_tau}
\footnotesize
\setlength{\tabcolsep}{4pt}
\begin{tabular}{@{}lcccc@{}}
\toprule
Setting & TC threshold & \makecell{Development length\\native $\rightarrow$ TC} & \makecell{Development\\human mean}
  & \makecell{Test length\\native $\rightarrow$ TC} \\
\midrule
COCO-TP & 0.67 to 0.76 & 2.92 to 3.18 $\rightarrow$ 3.88 to 3.91 & 3.89 & 2.93 $\rightarrow$ 3.85 \\
COCO-TA & 0.36 to 0.44 & 6.07 to 6.50 $\rightarrow$ 5.54 to 5.57 & 5.61 & 6.45 $\rightarrow$ 5.67 \\
\bottomrule
\end{tabular}
\end{table}

On COCO-TP, TC lengthens the overly short native predictions of HAT-SF. Conversely, on COCO-TA, TC shortens its overly long native predictions (Table~\ref{tab:supp_base_hat_tau}).

\begin{table*}[t]
\centering
\caption{Paired comparisons on COCO-TP (seeds 3 to 5, first fixation kept). Paired difference with 95\%
interval.}
\label{tab:supp_base_hat_tp}
\footnotesize
\setlength{\tabcolsep}{4pt}
\renewcommand{\arraystretch}{1.1}
\begin{tabular}{@{}l cccccc@{}}
\toprule
Comparison & MM & SM & SED & KLD & CC & NSS \\
\midrule
\makecell[l]{HAT-SF + TC\\$-$ HAT-SF}
  & \makecell{$+0.0059^{\checkmark}$\\{\scriptsize$[+0.0019,+0.0098]$}}
  & \makecell{$-0.0108^{\times}$\\{\scriptsize$[-0.0201,-0.0005]$}}
  & \makecell{$+0.316^{\times}$\\{\scriptsize$[+0.237,+0.397]$}}
  & \makecell{$-0.714^{\checkmark}$\\{\scriptsize$[-1.001,-0.441]$}}
  & \makecell{$+0.0185^{\checkmark}$\\{\scriptsize$[+0.0062,+0.0307]$}}
  & \makecell{$+0.057^\dagger$\\{\scriptsize$[-0.025,+0.145]$}} \\
\addlinespace[2pt]
\makecell[l]{Reg+Map + SR + TC\\$-$ HAT-SF + TC}
  & \makecell{$-0.0015^\dagger$\\{\scriptsize$[-0.0047,+0.0016]$}}
  & \makecell{$+0.0168^{\checkmark}$\\{\scriptsize$[+0.0055,+0.0280]$}}
  & \makecell{$+0.072^\dagger$\\{\scriptsize$[-0.130,+0.288]$}}
  & \makecell{$+1.064^{\times}$\\{\scriptsize$[+0.611,+1.529]$}}
  & \makecell{$-0.0914^{\times}$\\{\scriptsize$[-0.1224,-0.0567]$}}
  & \makecell{$-0.720^{\times}$\\{\scriptsize$[-0.950,-0.460]$}} \\
\bottomrule
\end{tabular}
\end{table*}

Under the same TC rule, we outperform HAT-SF on ScanMatch (\(+0.017\)). We detect no differences in MultiMatch or SED, while HAT-SF performs better on KLD, CC, and NSS (Table~\ref{tab:supp_base_hat_tp}). This aligns with the main paper. Notably, TC decreases the ScanMatch of HAT-SF by 0.011 but improves its MultiMatch, KLD, and CC. Therefore, this comparison evaluates both models under a shared termination rule rather than comparing their individual optimal configurations. Table~\ref{tab:supp_base_first_paired} presents the comparison with native HAT-SF.

\begin{table*}[t]
\centering
\caption{Paired comparisons on COCO-TA (seeds 3 to 5, image level). Paired difference with 95\%
interval.}
\label{tab:supp_base_hat_ta}
\footnotesize
\setlength{\tabcolsep}{4pt}
\renewcommand{\arraystretch}{1.1}
\begin{tabular}{@{}l cccccc@{}}
\toprule
Comparison & MM & SM & SED & KLD & CC & NSS \\
\midrule
\makecell[l]{HAT-SF + TC\\$-$ HAT-SF}
  & \makecell{$-0.000^\dagger$\\{\scriptsize$[-0.001,+0.001]$}}
  & \makecell{$+0.016^{\checkmark}$\\{\scriptsize$[+0.007,+0.024]$}}
  & \makecell{$-0.401^{\checkmark}$\\{\scriptsize$[-0.575,-0.199]$}}
  & \makecell{$+0.479^{\times}$\\{\scriptsize$[+0.261,+0.689]$}}
  & \makecell{$+0.008^{\checkmark}$\\{\scriptsize$[+0.002,+0.014]$}}
  & \makecell{$+0.055^{\checkmark}$\\{\scriptsize$[+0.023,+0.093]$}} \\
\addlinespace[2pt]
\makecell[l]{Reg+Map + SR + TC\\$-$ HAT-SF}
  & \makecell{$-0.008^{\times}$\\{\scriptsize$[-0.011,-0.004]$}}
  & \makecell{$-0.011^\dagger$\\{\scriptsize$[-0.023,+0.001]$}}
  & \makecell{$+0.772^{\times}$\\{\scriptsize$[+0.386,+1.171]$}}
  & \makecell{$+3.359^{\times}$\\{\scriptsize$[+2.888,+3.797]$}}
  & \makecell{$-0.106^{\times}$\\{\scriptsize$[-0.125,-0.086]$}}
  & \makecell{$-0.540^{\times}$\\{\scriptsize$[-0.632,-0.446]$}} \\
\addlinespace[2pt]
\makecell[l]{Reg+Map + SR + TC\\$-$ HAT-SF + TC}
  & \makecell{$-0.007^{\times}$\\{\scriptsize$[-0.011,-0.004]$}}
  & \makecell{$-0.026^{\times}$\\{\scriptsize$[-0.039,-0.013]$}}
  & \makecell{$+1.173^{\times}$\\{\scriptsize$[+0.765,+1.594]$}}
  & \makecell{$+2.880^{\times}$\\{\scriptsize$[+2.529,+3.225]$}}
  & \makecell{$-0.113^{\times}$\\{\scriptsize$[-0.133,-0.094]$}}
  & \makecell{$-0.595^{\times}$\\{\scriptsize$[-0.689,-0.501]$}} \\
\addlinespace[2pt]
\makecell[l]{TCSD + SR + TC\\$-$ HAT-SF + TC}
  & \makecell{$-0.006^{\times}$\\{\scriptsize$[-0.009,-0.003]$}}
  & \makecell{$-0.023^{\times}$\\{\scriptsize$[-0.035,-0.011]$}}
  & \makecell{$+1.060^{\times}$\\{\scriptsize$[+0.659,+1.475]$}}
  & \makecell{$+2.601^{\times}$\\{\scriptsize$[+2.282,+2.922]$}}
  & \makecell{$-0.102^{\times}$\\{\scriptsize$[-0.120,-0.083]$}}
  & \makecell{$-0.551^{\times}$\\{\scriptsize$[-0.641,-0.460]$}} \\
\bottomrule
\end{tabular}
\end{table*}

On COCO-TA, HAT-SF already outperforms our models under its native stopping rule. We detect no ScanMatch difference, while HAT-SF remains better on the other five metrics (Table~\ref{tab:supp_base_hat_ta}). Under the same TC rule, HAT-SF outperforms on all six metrics (ScanMatch \(-0.026\)), consistent with the main paper. TCSD exhibits the same behavior.

At each step, HAT-SF takes the argmax of its spatial map. Following Section~3.5 of the main paper, we select its readout on the development set from argmax and SR at \(\kappa=1\), 0.5, and 0.25. We use the same mean rank over MultiMatch, ScanMatch, and SED as the selection criterion. We break ties in favor of the native argmax.

\begin{table}[t]
\centering
\caption{Readout selection for HAT-SF on the development set (COCO-TP, mean over seeds 3 to 5).}
\label{tab:supp_base_hat_readout}
\footnotesize
\setlength{\tabcolsep}{5pt}
\begin{tabular}{@{}lcccc@{}}
\toprule
Readout & MM & SM & SED & Mean rank \\
\midrule
Argmax (native) & 0.833 & 0.505 & 2.30 & 1.67 \\
$\kappa=0.25$   & 0.843 & 0.495 & 2.59 & 1.67 \\
$\kappa=0.5$    & 0.834 & 0.398 & 3.67 & 2.67 \\
$\kappa=1$      & 0.814 & 0.214 & 5.57 & 4.00 \\
\bottomrule
\end{tabular}
\end{table}

Argmax and \(\kappa=0.25\) tie in mean rank, so the predefined rule retains argmax (Table~\ref{tab:supp_base_hat_readout}). Consequently, HAT-SF keeps its native readout. Forcing \(\kappa=0.5\) on COCO-TA degrades all six metrics. Relative to argmax, \(\kappa=0.5\) yields differences of MultiMatch \(-0.008\) \([-0.010, -0.005]\), ScanMatch \(-0.049\) \([-0.058, -0.041]\), SED \(+0.333\) \([+0.219, +0.453]\), KLD \(+4.235\) \([+3.921, +4.558]\), CC \(-0.132\) \([-0.152, -0.112]\), and NSS \(-0.604\) \([-0.695, -0.515]\). Because HAT-SF's current fixation influences subsequent steps, this test differs from replacing coordinates on fixed parallel outputs.

\subsection{Full readout results of Gazeformer-Aux}
\label{sec:supp_base_gazeformer}

\paragraph{Setup.}
Gazeformer-Aux adds a supervised auxiliary spatial head to Gazeformer. The remaining architecture, losses, and native stopping rule remain unchanged. We train it with seeds 1 to 5 and evaluate on the COCO-TP test set. This set contains 315 images and 3,258 scanpaths per seed. Each batch runs one forward pass. The four readouts share identical first fixations, stopping positions, and sampled durations. Only the coordinates differ. The native readout samples from each slot's Gaussian. The Gaussian mean readout uses the expectation of this distribution and retains the original adapter's non-negative coordinate transform. SR with \(\kappa=1\) and \(\kappa=0.5\) computes plain and sharpened expectations over the auxiliary spatial map, respectively. We do not apply TC.

\begin{table}[t]
\centering
\caption{Four readouts of Gazeformer-Aux on the COCO-TP test set (mean over seeds 1 to 5).}
\label{tab:supp_base_gaux_means}
\footnotesize
\setlength{\tabcolsep}{4pt}
\begin{tabular}{@{}l cccccc@{}}
\toprule
Readout & MM $\uparrow$ & SM $\uparrow$ & SED $\downarrow$ & KLD $\downarrow$ & CC $\uparrow$ & NSS $\uparrow$ \\
\midrule
Native Gaussian sampling & 0.8153 & 0.4155 & 2.6028 & 12.7666 & 0.3079 & 2.1334 \\
Gaussian mean            & 0.8153 & 0.4156 & 2.6036 & 12.8215 & 0.3076 & 2.1317 \\
SR ($\kappa=1$)          & 0.8229 & 0.4214 & 2.5614 & 13.3848 & 0.3002 & 2.0733 \\
SR ($\kappa=0.5$)        & 0.8221 & 0.4564 & 2.3990 & 11.4203 & 0.4051 & 2.8063 \\
\bottomrule
\end{tabular}
\end{table}

\begin{table*}[t]
\centering
\caption{Paired differences for Gazeformer-Aux (seeds $\times$ images, 20,000 replicates, images weighted
equally). Sampling: native Gaussian sampling. Mean: Gaussian mean. SR$_1$ and SR$_{0.5}$: SR with
$\kappa=1$ and $\kappa=0.5$.}
\label{tab:supp_base_gaux_paired}
\footnotesize
\setlength{\tabcolsep}{3pt}
\renewcommand{\arraystretch}{1.1}
\begin{tabular}{@{}l cccccc@{}}
\toprule
Comparison & MM & SM & SED & KLD & CC & NSS \\
\midrule
Mean $-$ Sampling
  & \makecell{$-0.0000^\dagger$\\{\scriptsize$[-0.0001,+0.0001]$}}
  & \makecell{$+0.0001^\dagger$\\{\scriptsize$[-0.0002,+0.0004]$}}
  & \makecell{$+0.0009^\dagger$\\{\scriptsize$[-0.0026,+0.0044]$}}
  & \makecell{$+0.0547^{\times}$\\{\scriptsize$[+0.0420,+0.0695]$}}
  & \makecell{$-0.0002^{\times}$\\{\scriptsize$[-0.0004,-0.0001]$}}
  & \makecell{$-0.0017^{\times}$\\{\scriptsize$[-0.0038,-0.0001]$}} \\
\addlinespace[2pt]
SR$_1$ $-$ Mean
  & \makecell{$+0.0071^{\checkmark}$\\{\scriptsize$[+0.0043,+0.0103]$}}
  & \makecell{$+0.0050^\dagger$\\{\scriptsize$[-0.0088,+0.0177]$}}
  & \makecell{$-0.0387^\dagger$\\{\scriptsize$[-0.1095,+0.0301]$}}
  & \makecell{$+0.5770^{\times}$\\{\scriptsize$[+0.0444,+1.1341]$}}
  & \makecell{$-0.0080^\dagger$\\{\scriptsize$[-0.0310,+0.0143]$}}
  & \makecell{$-0.0600^\dagger$\\{\scriptsize$[-0.2254,+0.1022]$}} \\
\addlinespace[2pt]
SR$_{0.5}$ $-$ Mean
  & \makecell{$+0.0061^{\checkmark}$\\{\scriptsize$[+0.0031,+0.0093]$}}
  & \makecell{$+0.0399^{\checkmark}$\\{\scriptsize$[+0.0296,+0.0503]$}}
  & \makecell{$-0.2021^{\checkmark}$\\{\scriptsize$[-0.2662,-0.1434]$}}
  & \makecell{$-1.3796^{\checkmark}$\\{\scriptsize$[-1.7982,-0.9598]$}}
  & \makecell{$+0.0971^{\checkmark}$\\{\scriptsize$[+0.0756,+0.1185]$}}
  & \makecell{$+0.6743^{\checkmark}$\\{\scriptsize$[+0.5149,+0.8378]$}} \\
\addlinespace[2pt]
SR$_{0.5}$ $-$ Sampling
  & \makecell{$+0.0061^{\checkmark}$\\{\scriptsize$[+0.0031,+0.0093]$}}
  & \makecell{$+0.0400^{\checkmark}$\\{\scriptsize$[+0.0298,+0.0504]$}}
  & \makecell{$-0.2012^{\checkmark}$\\{\scriptsize$[-0.2652,-0.1429]$}}
  & \makecell{$-1.3249^{\checkmark}$\\{\scriptsize$[-1.7498,-0.8982]$}}
  & \makecell{$+0.0969^{\checkmark}$\\{\scriptsize$[+0.0753,+0.1183]$}}
  & \makecell{$+0.6726^{\checkmark}$\\{\scriptsize$[+0.5130,+0.8364]$}} \\
\addlinespace[2pt]
SR$_{0.5}$ $-$ SR$_1$
  & \makecell{$-0.0010^\dagger$\\{\scriptsize$[-0.0026,+0.0006]$}}
  & \makecell{$+0.0349^{\checkmark}$\\{\scriptsize$[+0.0278,+0.0424]$}}
  & \makecell{$-0.1634^{\checkmark}$\\{\scriptsize$[-0.2088,-0.1190]$}}
  & \makecell{$-1.9566^{\checkmark}$\\{\scriptsize$[-2.2869,-1.6342]$}}
  & \makecell{$+0.1051^{\checkmark}$\\{\scriptsize$[+0.0880,+0.1226]$}}
  & \makecell{$+0.7343^{\checkmark}$\\{\scriptsize$[+0.5989,+0.8724]$}} \\
\bottomrule
\end{tabular}
\end{table*}

The Gaussian mean and native Gaussian sampling yield nearly identical
ScanMatch scores (Tables~\ref{tab:supp_base_gaux_means}
and~\ref{tab:supp_base_gaux_paired}). The sharpened readout
($\kappa=0.5$) significantly improves all six metrics relative to
both baselines. For plain expectation ($\kappa=1$), the paired
comparison against the Gaussian mean shows a significant improvement
only in MultiMatch and a significant increase in KLD (worse).
Section~4.4 of the main paper reports the same direction of change
in KLD, but uses native Gaussian sampling as the reference.
Compared with plain expectation, sharpening significantly improves
ScanMatch, SED, KLD, CC, and NSS.

As a descriptive reference, both the original Gazeformer (seeds 3 to 5) and Gazeformer-Aux (seeds 1 to 5) achieve a native ScanMatch of approximately 0.416 (Table~\ref{tab:supp_full_tp}). Because they use different seeds, we do not run a paired test. Therefore, this comparison cannot determine whether the auxiliary spatial head itself affects the native result. The conclusions in this subsection derive solely from the same-checkpoint readout comparison.

\paragraph{Limitations.}
We do not strictly match model capacity, optimization, or visual frontends. Consequently, the remaining gap cannot be attributed solely to generation order. Among external baselines, we test TC only on HAT-SF. AiR-D lacks a retrained external baseline.

\section{Equivalence analysis}
\label{sec:supp_equivalence}

Section~4.2 of the main paper detects no TCSD advantage under matched readout and termination. Beyond this null result, we test whether the per-metric paired difference falls within an exploratory equivalence bound derived from the test set.

\subsection{Method}
\label{sec:supp_eq_method}

\paragraph{Two one-sided tests (TOST).}
For each metric, we define an equivalence bound \(\pm\Delta\). If the 90\% bootstrap interval of the paired difference lies entirely inside \((-\Delta, +\Delta)\), we consider the bound met at the 5\% level. This is equivalent to rejecting, via two one-sided 5\% tests, the hypothesis that the difference is at least \(\Delta\) in either direction. We use the bootstrap procedure from the main paper, which jointly resamples seeds and images with 20,000 replicates. We evaluate all 10 shared seeds (3 to 12) between Reg+Map and TCSD on the COCO-TP test set using the default threshold \(\tau=0.5\).

\paragraph{Equivalence bounds.}
We define \(\Delta\) as 20\% of the gain of Reg+Map + SR (\(\kappa=1\)) over native regression (seeds 3 to 5). The bounds use differences between the reported means. For ScanMatch,
$\Delta=0.2(0.4725-0.3649)=0.02152$. These exploratory bounds derive from existing COCO-TP test gains. We use the magnitude of each gain to ensure a positive bound for every metric. Table~\ref{tab:supp_eq_bounds} lists these bounds.

\begin{table}[t]
\centering
\caption{Equivalence bounds $\Delta$.}
\label{tab:supp_eq_bounds}
\footnotesize
\setlength{\tabcolsep}{5pt}
\begin{tabular}{@{}lcccccc@{}}
\toprule
& MM & SM & SED & KLD & CC & NSS \\
\midrule
$\Delta$ & 0.0032 & 0.02152 & 0.09108 & 1.08022 & 0.0399 & 0.2787 \\
\bottomrule
\end{tabular}
\end{table}

\paragraph{Comparisons.}
We run TOST for two comparisons at the default threshold.
\begin{itemize}
\item Native TCSD (the \(\kappa=1\) expectation) against Reg+Map + SR (\(\kappa=1\)).
\item TCSD + SR (\(\kappa=0.5\)) against Reg+Map + SR (\(\kappa=0.5\)).
\end{itemize}
These are the only two formal TOST comparisons. Section~4.2 of the
main paper compares TCSD (native expectation, $\kappa=1$) + TC with
Reg+Map + SR ($\kappa=1$) + TC. Both use the same termination
calibration rule, with a threshold selected separately for each
checkpoint on the development set. This comparison is not part of the
formal TOST analysis; we examine it post hoc in
Section~\ref{sec:supp_eq_discussion}.

\subsection{Results}
\label{sec:supp_eq_results}

\begin{table*}[t]
\centering
\caption{TCSD minus Reg+Map (COCO-TP test set, seeds 3 to 12, $\tau=0.5$). Differences are raw. For SED
and KLD, a negative value means that TCSD is better. ``Met'' means that the 90\% interval lies inside
$\pm\Delta$.}
\label{tab:supp_eq_results}
\footnotesize
\setlength{\tabcolsep}{5pt}
\begin{tabular}{@{}llccccc@{}}
\toprule
Comparison & Metric & Difference & 95\% interval & 90\% interval & $\Delta$ & Bound \\
\midrule
\multirow{6}{*}{\makecell[l]{TCSD\\$-$ Reg+Map + SR ($\kappa=1$)}}
 & MM  & $-0.0026$ & $[-0.0047,-0.0005]$ & $[-0.0043,-0.0008]$ & 0.0032 & Not met \\
 & SM  & $-0.0178$ & $[-0.0310,-0.0056]$ & $[-0.0287,-0.0077]$ & 0.0215 & Not met \\
 & SED & $+0.585$  & $[+0.232,+0.994]$   & $[+0.282,+0.920]$   & 0.091  & Not met \\
 & KLD & $-0.106$  & $[-0.395,+0.200]$   & $[-0.351,+0.148]$   & 1.080  & Met \\
 & CC  & $-0.0085$ & $[-0.0230,+0.0059]$ & $[-0.0207,+0.0037]$ & 0.0399 & Met \\
 & NSS & $-0.062$  & $[-0.182,+0.054]$   & $[-0.161,+0.037]$   & 0.279  & Met \\
\midrule
\multirow{6}{*}{\makecell[l]{TCSD + SR ($\kappa=0.5$)\\$-$ Reg+Map + SR ($\kappa=0.5$)}}
 & MM  & $-0.0019$ & $[-0.0040,+0.0002]$ & $[-0.0037,-0.0001]$ & 0.0032 & Not met \\
 & SM  & $-0.0191$ & $[-0.0317,-0.0075]$ & $[-0.0295,-0.0094]$ & 0.0215 & Not met \\
 & SED & $+0.573$  & $[+0.229,+0.973]$   & $[+0.279,+0.900]$   & 0.091  & Not met \\
 & KLD & $-0.106$  & $[-0.322,+0.104]$   & $[-0.286,+0.071]$   & 1.080  & Met \\
 & CC  & $-0.0061$ & $[-0.0184,+0.0064]$ & $[-0.0165,+0.0045]$ & 0.0399 & Met \\
 & NSS & $-0.052$  & $[-0.158,+0.060]$   & $[-0.142,+0.041]$   & 0.279  & Met \\
\bottomrule
\end{tabular}
\end{table*}

\paragraph{Saliency metrics.}
In both comparisons, the 90\% intervals of KLD, CC, and NSS fall entirely within the chosen bounds (Table~\ref{tab:supp_eq_results}). These metrics therefore satisfy the equivalence criterion under these exploratory bounds.

\paragraph{Sequence metrics.}
MultiMatch, ScanMatch, and SED do not meet the bounds. For ScanMatch and SED, the 95\% intervals exclude zero. In both cases, TCSD performs worse. TCSD also performs worse on MultiMatch in the first comparison. In the second comparison, the 95\% MultiMatch interval contains zero. However, its 90\% interval extends beyond \(-\Delta\). Although we detect no difference here, the bound is not met. The SED bound (0.091) is much smaller than the observed difference (about 0.58). This occurs because the spatial readout gain on SED is inherently small at the default threshold.

\subsection{Interpretation and limitations}
\label{sec:supp_eq_discussion}

At the default threshold, TCSD and Reg+Map meet the bounds on saliency metrics. However, on sequence metrics, TCSD either performs worse or yields inconclusive results.

\paragraph{Relation to the TC comparison of the main paper.}
The main paper reports that all six 95\% intervals contain zero when both models use \(\kappa=1\) with identical termination calibration (ScanMatch \(-0.001\) \([-0.008, 0.006]\)). This specific setting is not part of the formal TOST. We perform a post-hoc check using the same bounds. The 90\% intervals fall within the bounds for MultiMatch, ScanMatch, KLD, CC, and NSS. However, SED does not meet its bound. Its difference is \(-0.010\), and its 90\% interval \([-0.118, +0.099]\) exceeds the \(\pm0.091\) limit at both ends. Therefore, even when all six 95\% intervals contain zero, we cannot describe these metrics as equivalent.

\paragraph{Limitations.}
\begin{itemize}
\item Our bounds derive from existing gains on the same test set, using a chosen ratio of 20\%. Alternative bounds could alter the conclusions. This entire analysis is exploratory.
\item Our formal TOST evaluates only the two comparisons at the default threshold. The \(\kappa=1\) check with TC remains post hoc. We do not perform an equivalence analysis for \(\kappa=0.5\) with TC.
\item We conduct this analysis only on COCO-TP. For COCO-TA and AiR-D, the TCSD and Reg+Map comparison reports only difference tests (Section~4.3 of the main paper).
\end{itemize}

\section{Termination and the stop loss}
\label{sec:supp_termination}

Termination calibration (TC) chooses a threshold on the development set to
bring the mean predicted length as close as possible to the human mean
(Section~3.4 of the main paper). It does not match the full length distribution.
To examine the benefit of predicting a length for each input, we compare the
calibrated lengths with fixed and shuffled lengths
(Section~\ref{sec:supp_term_length}). We then test all-slot stop supervision
(Section~\ref{sec:supp_term_stoploss}) and compare it with inference-time
calibration (Section~\ref{sec:supp_term_cross}).

\paragraph{Conventions.}
SR denotes the spatial expectation readout with \(\kappa=0.5\), the setting locked in the main paper. Means and paired intervals are computed as described in Section~\ref{sec:supp_evaluation}.

\subsection{Length analysis on COCO-TA}
\label{sec:supp_term_length}

\paragraph{Setup.}
We use Reg+Map and TCSD with seeds 3 to 7. We keep the SR coordinates fixed and vary only the scanpath lengths. We compare three length settings.
\begin{itemize}
\item Calibrated length. The length generated by the model at the TC threshold (mean 6.13).
\item Fixed length. We truncate every scanpath to a fixed integer length \(N_0=6\). We select \(N_0\) on the development set (mean 6.00).
\item Shuffled length. We randomly permute the calibrated lengths across task-image groups. This preserves the group-level length distribution, but the lengths no longer correspond to their original images. Because groups contain varying numbers of scanpaths, the overall scanpath-level distribution closely matches the calibrated distribution, though they are not identical (mean 6.13).
\end{itemize}

\begin{table*}[t]
\centering
\caption{Calibrated length minus a baseline length on COCO-TA, SR coordinates fixed, seeds 3 to 7. Each
cell gives the paired difference in raw units and its 95\% interval. For every interval that excludes zero,
all 5 seeds point in the same direction. Lower is better for SED and KLD, so a positive SED or KLD difference favors
the baseline.}
\label{tab:supp_term_baselines}
\footnotesize
\setlength{\tabcolsep}{5pt}
\renewcommand{\arraystretch}{1.1}
\begin{tabular}{ll cccccc}
\toprule
Model & Baseline & MultiMatch & ScanMatch & SED & KLD & CC & NSS \\
\midrule
Reg+Map & Fixed
  & \makecell{$-0.000^\dagger$\\{\scriptsize$[-0.003,+0.002]$}}
  & \makecell{$+0.007^\dagger$\\{\scriptsize$[-0.003,+0.017]$}}
  & \makecell{$+0.807^{\times}$\\{\scriptsize$[+0.437,+1.193]$}}
  & \makecell{$+0.582^{\times}$\\{\scriptsize$[+0.482,+0.688]$}}
  & \makecell{$+0.016^{\checkmark}$\\{\scriptsize$[+0.006,+0.025]$}}
  & \makecell{$+0.098^{\checkmark}$\\{\scriptsize$[+0.056,+0.140]$}} \\
\addlinespace[2pt]
Reg+Map & Shuffled
  & \makecell{$+0.001^\dagger$\\{\scriptsize$[-0.001,+0.003]$}}
  & \makecell{$+0.015^{\checkmark}$\\{\scriptsize$[+0.006,+0.023]$}}
  & \makecell{$-0.242^\dagger$\\{\scriptsize$[-0.650,+0.177]$}}
  & \makecell{$+0.006^\dagger$\\{\scriptsize$[-0.098,+0.104]$}}
  & \makecell{$+0.010^{\checkmark}$\\{\scriptsize$[+0.002,+0.019]$}}
  & \makecell{$+0.053^{\checkmark}$\\{\scriptsize$[+0.013,+0.094]$}} \\
\addlinespace[2pt]
TCSD & Fixed
  & \makecell{$+0.000^\dagger$\\{\scriptsize$[-0.003,+0.003]$}}
  & \makecell{$+0.004^\dagger$\\{\scriptsize$[-0.006,+0.015]$}}
  & \makecell{$+0.879^{\times}$\\{\scriptsize$[+0.491,+1.283]$}}
  & \makecell{$+0.538^{\times}$\\{\scriptsize$[+0.379,+0.704]$}}
  & \makecell{$+0.019^{\checkmark}$\\{\scriptsize$[+0.010,+0.028]$}}
  & \makecell{$+0.110^{\checkmark}$\\{\scriptsize$[+0.067,+0.153]$}} \\
\addlinespace[2pt]
TCSD & Shuffled
  & \makecell{$+0.002^{\checkmark}$\\{\scriptsize$[+0.000,+0.004]$}}
  & \makecell{$+0.014^{\checkmark}$\\{\scriptsize$[+0.005,+0.023]$}}
  & \makecell{$-0.247^\dagger$\\{\scriptsize$[-0.661,+0.177]$}}
  & \makecell{$-0.035^\dagger$\\{\scriptsize$[-0.136,+0.063]$}}
  & \makecell{$+0.012^{\checkmark}$\\{\scriptsize$[+0.004,+0.020]$}}
  & \makecell{$+0.061^{\checkmark}$\\{\scriptsize$[+0.022,+0.103]$}} \\
\bottomrule
\end{tabular}
\end{table*}

\paragraph{Calibrated versus shuffled.}
Compared to shuffled lengths, calibrated lengths yield higher ScanMatch, CC, and NSS (Table~\ref{tab:supp_term_baselines}). We detect no difference in SED or KLD. Since both share identical group-level length distributions, the observed gains stem primarily from the sample-to-length assignment. This indicates that the model's per-sample lengths encode information specific to both the image and the task.

\paragraph{Calibrated versus fixed.}
Compared to the fixed length, we detect no ScanMatch difference for calibrated lengths. Calibrated lengths yield higher CC and NSS, but noticeably worse SED and KLD. Therefore, a single suitable constant length can reproduce most of the ScanMatch gain that TC provides on COCO-TA. Calibrated lengths provide additional benefits in CC and NSS, though they incur a cost in SED and KLD.

\begin{table*}[t]
\centering
\caption{Per-sample length error on COCO-TA, mean over seeds 3 to 7. The constant here is the real value
5.61, the human mean length on the development set. It is used only for this error analysis and is not
the integer fixed length 6 of Table~\ref{tab:supp_term_baselines}. A group is one (task, image) pair, and
the human length of a group is the mean over its scanpaths. For the shuffled Spearman correlation we give
the mean of the permutation distribution and its 95th percentile in brackets. Cal.: calibrated. Def.\ $\tau$:
default threshold. Const.: constant. Shuf.: shuffled.}
\label{tab:supp_term_mae}
\footnotesize
\setlength{\tabcolsep}{4pt}
\begin{tabular}{l c cccc ccc cc}
\toprule
& & \multicolumn{4}{c}{MAE per scanpath} & \multicolumn{3}{c}{MAE per group} & \multicolumn{2}{c}{Spearman per group} \\
\cmidrule(lr){3-6}\cmidrule(lr){7-9}\cmidrule(lr){10-11}
Model & \makecell{Mean length\\pred / human} & Cal. & Def.\ $\tau$ & Const. & Shuf.
  & Cal. & Const. & Shuf. & Cal. & Shuf. \\
\midrule
Reg+Map & 6.13 / 5.89 & 3.79 & 9.66 & 2.67 & 4.32 & 3.46 & 1.31 & 4.00 & 0.494 & 0.000 (0.091) \\
TCSD    & 6.21 / 5.89 & 3.89 & 9.48 & 2.67 & 4.45 & 3.57 & 1.31 & 4.15 & 0.497 & 0.000 (0.091) \\
\bottomrule
\end{tabular}
\end{table*}

\paragraph{Length error.}
The Spearman correlation between calibrated and human group lengths is approximately 0.49 (Table~\ref{tab:supp_term_mae}). This is far above the permutation distribution of shuffled lengths (mean near 0, 95th percentile 0.091). Group error is also lower than for shuffled lengths. However, the absolute error is higher than that of the constant 5.61. 

This discrepancy arises from the shape of the length distribution. For Reg+Map with seed 3, the predicted length percentiles (10th, 25th, 50th, 75th, 90th) are 3, 3, 4, 5, and 16. In contrast, the human percentiles are 3, 3, 5, 7, and 11. Furthermore, 22.4\% of predicted scanpaths reach the maximum limit of 16 (19.6\% for TCSD). TC aligns the mean predicted length with the human mean. This occurs because most paths are somewhat short while a few hit the limit. Consequently, TC does not bring each individual path close to its corresponding human length.

\subsection{Stop-loss ablation}
\label{sec:supp_term_stoploss}

\paragraph{Original and all-slot stop loss.}
In the original training, the stop head receives one positive label at the true scanpath end. Slots after the end are excluded from the stop loss. We introduce an all-slot variant. This variant computes binary cross-entropy over every slot, assigning label 1 to the end slot and all subsequent slots. The other loss terms and training configuration remain identical. The stop term in the development-set selection metric also adopts this all-slot definition. 

We retrain in three settings: COCO-TP (seeds 3 to 5), AiR-D (seeds 3 to 5), and COCO-TA (seeds 3 to 7). The reference in each setting is Reg+Map with the original stop loss under identical configuration. We evaluate every checkpoint using both the native readout and SR, applying \(\tau=0.5\) and its respective TC threshold.

\begin{table}[t]
\centering
\caption{Mean predicted length and TC threshold. TC targets the human mean length on the development set
(COCO-TP 3.89, COCO-TA 5.61). Thresholds give the range over seeds.}
\label{tab:supp_term_lengths}
\small
\setlength{\tabcolsep}{4pt}
\begin{tabular}{ll ccc}
\toprule
Setting & Stop loss & \makecell{Length\\$\tau=0.5$} & \makecell{Length\\TC} & \makecell{TC\\threshold} \\
\midrule
COCO-TP & Original & 4.49  & 3.56  & 0.37 to 0.47 \\
COCO-TP & All-slot & 3.44  & 3.80  & 0.61 to 0.68 \\
\addlinespace[2pt]
AiR-D   & Original & 12.46 & 10.08 & 0.28 to 0.32 \\
AiR-D   & All-slot & 9.87  & 10.05 & 0.50 to 0.59 \\
\addlinespace[2pt]
COCO-TA & Original & 15.46 & 6.13  & 0.24 to 0.28 \\
COCO-TA & All-slot & 4.86  & 5.75  & 0.56 to 0.68 \\
\bottomrule
\end{tabular}
\end{table}

With the original stop loss, the default threshold yields overly long sequences. Consequently, TC must lower the threshold below 0.5 (Table~\ref{tab:supp_term_lengths}). The gap is largest on COCO-TA (15.46 vs. 5.61). With the all-slot loss, lengths at the default threshold are already close to the TC target or slightly shorter. TC selects thresholds of 0.5 or higher, resulting in only minor path lengthening.

\begin{table*}[t]
\centering
\caption{ScanMatch gains of TC and SR on the same checkpoint (paired difference with 95\% interval). The
TC gain is the TC threshold minus $\tau=0.5$ for the same readout. The SR gain is SR minus the native
readout at the same threshold. For COCO-TA with the original loss, the two SR
gains correspond to Table~3 and Section~4.3 of the main paper, respectively.}
\label{tab:supp_term_gains}
\small
\setlength{\tabcolsep}{5pt}
\renewcommand{\arraystretch}{1.1}

\begin{tabular}{ll cccc}
\toprule
Setting & Stop loss & TC gain (native) & TC gain (SR) & SR gain, $\tau=0.5$ & SR gain, TC \\
\midrule
COCO-TP & Original
  & \makecell{$+0.0178^{\checkmark}$\\{\scriptsize$[+0.0063,+0.0284]$}}
  & \makecell{$+0.0204^{\checkmark}$\\{\scriptsize$[+0.0064,+0.0342]$}}
  & \makecell{$+0.1347^{\checkmark}$\\{\scriptsize$[+0.1055,+0.1683]$}}
  & \makecell{$+0.1373^{\checkmark}$\\{\scriptsize$[+0.1084,+0.1726]$}} \\
COCO-TP & All-slot
  & \makecell{$-0.0113^{\times}$\\{\scriptsize$[-0.0141,-0.0085]$}}
  & \makecell{$-0.0161^{\times}$\\{\scriptsize$[-0.0219,-0.0100]$}}
  & \makecell{$+0.1896^{\checkmark}$\\{\scriptsize$[+0.1571,+0.2242]$}}
  & \makecell{$+0.1848^{\checkmark}$\\{\scriptsize$[+0.1547,+0.2180]$}} \\
\addlinespace[2pt]
AiR-D & Original
  & \makecell{$+0.0272^{\checkmark}$\\{\scriptsize$[+0.0238,+0.0310]$}}
  & \makecell{$+0.0343^{\checkmark}$\\{\scriptsize$[+0.0302,+0.0388]$}}
  & \makecell{$+0.0375^{\checkmark}$\\{\scriptsize$[+0.0269,+0.0486]$}}
  & \makecell{$+0.0446^{\checkmark}$\\{\scriptsize$[+0.0328,+0.0570]$}} \\
AiR-D & All-slot
  & \makecell{$-0.0012^\dagger$\\{\scriptsize$[-0.0023,+0.0000]$}}
  & \makecell{$-0.0014^\dagger$\\{\scriptsize$[-0.0027,+0.0000]$}}
  & \makecell{$+0.0503^{\checkmark}$\\{\scriptsize$[+0.0370,+0.0647]$}}
  & \makecell{$+0.0501^{\checkmark}$\\{\scriptsize$[+0.0369,+0.0644]$}} \\
\addlinespace[2pt]
COCO-TA & Original
  & \makecell{$+0.1384^{\checkmark}$\\{\scriptsize$[+0.1281,+0.1486]$}}
  & \makecell{$+0.155^{\checkmark}$\\{\scriptsize$[+0.142,+0.169]$}}
  & \makecell{$+0.029^{\checkmark}$\\{\scriptsize$[+0.025,+0.033]$}}
  & \makecell{$+0.045^{\checkmark}$\\{\scriptsize$[+0.038,+0.053]$}} \\
COCO-TA & All-slot
  & \makecell{$-0.0171^{\times}$\\{\scriptsize$[-0.0211,-0.0127]$}}
  & \makecell{$-0.017^{\times}$\\{\scriptsize$[-0.022,-0.013]$}}
  & \makecell{$+0.053^{\checkmark}$\\{\scriptsize$[+0.044,+0.061]$}}
  & \makecell{$+0.052^{\checkmark}$\\{\scriptsize$[+0.045,+0.060]$}} \\
\bottomrule
\end{tabular}

\end{table*}

\paragraph{The effect of TC depends on the stop loss.}
With the original stop loss, TC raises ScanMatch across all three settings, most notably on COCO-TA (Table~\ref{tab:supp_term_gains}). With the all-slot loss, we detect no TC effect on AiR-D. TC even slightly lowers ScanMatch on COCO-TP and COCO-TA. On the all-slot COCO-TA checkpoints, TC improves only KLD (\(-0.195\), interval excludes zero), while all other metrics degrade. 

A plausible explanation is that the original loss provides only one positive label. Consequently, the stop probability receives no supervision after the sequence end. This produces overly long lengths at the default threshold, which TC compensates for during inference. Once the stop head receives per-slot supervision during training, this correction is no longer necessary. The COCO-TA comparison in Table~\ref{tab:supp_term_gains} extends the main paper's seeds 3 to 5 to seeds 3 to 7 and maintains the same effect directions. 

\paragraph{The gain of SR remains after the stop supervision is replaced.}
Across both stop losses and all tested thresholds, SR consistently improves ScanMatch. Therefore, the SR benefit persists regardless of the stop supervision. However, the gain magnitude still depends on the termination method and varies with both the checkpoint and the threshold. Since the native readout also changes on the all-slot checkpoints (Section~\ref{sec:supp_term_cross}), we do not directly compare the gain magnitudes between the two checkpoint types.

\subsection{Training-side fix versus inference-side calibration}
\label{sec:supp_term_cross}

Here we compare different checkpoints using their recommended stop configurations. We evaluate the all-slot checkpoint at the default threshold and the original checkpoint at its TC threshold. Each configuration reflects the recommended use of its respective stop mechanism.

\begin{table}[t]
\centering
\caption{Cross-checkpoint comparison, all-slot minus original. ScanMatch, paired difference with 95\%
interval. The columns ``All-slot'' and ``Original'' give the threshold used for each checkpoint. The last
row is a separate control in which both sides use TC.}
\label{tab:supp_term_cross}
\small
\setlength{\tabcolsep}{4pt}
\renewcommand{\arraystretch}{1.1}
\begin{adjustbox}{max width=\linewidth}
\begin{tabular}{ll cc c}
\toprule
Setting & Readout & All-slot & Original & Difference \\
\midrule
COCO-TP & SR     & $\tau=0.5$ & TC & \makecell{$-0.0051^\dagger$\\{\scriptsize$[-0.0214,+0.0109]$}} \\
COCO-TP & Native & $\tau=0.5$ & TC & \makecell{$-0.0575^{\times}$\\{\scriptsize$[-0.0763,-0.0357]$}} \\
\addlinespace[2pt]
AiR-D   & SR     & $\tau=0.5$ & TC & \makecell{$+0.0037^\dagger$\\{\scriptsize$[-0.0044,+0.0124]$}} \\
AiR-D   & Native & $\tau=0.5$ & TC & \makecell{$-0.0021^\dagger$\\{\scriptsize$[-0.0092,+0.0051]$}} \\
\addlinespace[2pt]
COCO-TA & SR     & $\tau=0.5$ & TC & \makecell{$+0.0223^{\checkmark}$\\{\scriptsize$[+0.0152,+0.0295]$}} \\
COCO-TA & Native & $\tau=0.5$ & TC & \makecell{$+0.0148^{\checkmark}$\\{\scriptsize$[+0.0033,+0.0245]$}} \\
\midrule
COCO-TA & SR     & TC         & TC & \makecell{$+0.005^\dagger$\\{\scriptsize$[-0.003,+0.013]$}} \\
\bottomrule
\end{tabular}
\end{adjustbox}
\end{table}

\paragraph{SR readout.}
With SR, we detect no ScanMatch difference on COCO-TP or AiR-D between the all-slot checkpoint at the default threshold and the original checkpoint with TC (Table~\ref{tab:supp_term_cross}). On COCO-TP, the all-slot checkpoint achieves better SED (\(-0.341\), \([-0.547, -0.156]\)). On COCO-TA, the all-slot checkpoint yields higher ScanMatch (0.343 vs. 0.320, image-level means). Because these comparisons involve complete setups with different predicted lengths (4.86 vs. 6.13 on COCO-TA), they do not prove that the coordinate regression branch itself is more accurate. Applying TC to both setups on COCO-TA yields no detected difference.

\paragraph{Native readout.}
With the native readout, the direction depends on the setting. On COCO-TP, the all-slot checkpoint performs clearly worse (native ScanMatch 0.325 vs. 0.383). On AiR-D, we detect no difference. On COCO-TA, the all-slot checkpoint achieves a higher score (0.290 vs. 0.275). Thus, replacing the stop loss alters the native readout of the complete setup. 

\paragraph{Spatial map NLL.}
The all-slot version yields a slightly higher development-set negative log-likelihood (NLL) for the spatial map. The scores are 4.795 vs. 4.682 on COCO-TP and 5.623 vs. 5.612 on AiR-D. This constitutes a single descriptive result derived from the selected checkpoints.


\section{Mechanism analysis: regression output, map mean, and map mode}
\label{sec:supp_mechanism}

This section examines potential sources of the readout gap, building on the explanation in Section~5 of the main paper. For normalized coordinates, Smooth-L1 with \(\beta=1\) acts as a squared loss. Under ideal training, its optimum is the conditional mean of the supervision target. Plain map expectation (SR, \(\kappa=1\)) yields the mean of the model's predicted distribution. However, these two quantities need not coincide. The spatial map and the coordinate head are trained with different objectives. Additionally, finite capacity, optimization, and spatial target smoothing can all influence their outputs. Sharpening (\(\kappa=0.5\)) assigns more weight to high-probability cells.

Sections~\ref{sec:supp_mech_meanmode} to \ref{sec:supp_mech_valley} present post-hoc analyses of existing checkpoints. They examine the relationship between the regression output and the map mean or mode. They also investigate how map shape relates to readout gains. Section~\ref{sec:supp_mech_figure} provides examples. Section~\ref{sec:supp_mech_mdn} uses a mixture-density regression head as a control to isolate the roles of output parameterization and training objective.

\paragraph{Setup.}
Unless stated otherwise, we use the COCO-Search18 target-present (TP) test set, seeds 3 to 5, and the default threshold \(\tau=0.5\). Consequently, all readouts of a single checkpoint share identical lengths and durations. We compare four controlled models: Reg+Map and the three controls from Section~4.5 of the main paper. These controls provide the regression pathway with more task or spatial information.
\begin{itemize}
\item Equal access. The decoder attends directly to the task-fused features.
\item Spatial Matched Regression (SMR). The regression head also receives features pooled with the spatial map.
\item Full-map regression (FullMap). The regression head also receives the full spatial distribution of the slot.
\end{itemize}
For each decoded slot, we store its spatial map \(p_i\) and four coordinate sets. These include the native regression output, SR (\(\kappa=1\), the map mean), SR (\(\kappa=0.5\)), and the argmax (the map mode). We compute three statistics for each slot's map.
\begin{itemize}
\item Normalized entropy \(H(p_i)/\log G\). This value ranges from 0 to 1. A value of 0 indicates all probability mass on a single cell, while 1 represents a uniform distribution across all cells.
\item Number of peaks. We define a peak as a local maximum within a \(3\times3\) neighborhood. Its probability must be at least 0.1 times the slot's maximum.
\item Peak ratio. The ratio of the second-highest peak to the highest peak.
\end{itemize}
A slot is \emph{multimodal} if it has at least two peaks and a peak ratio of at least 0.3. All other slots are \emph{unimodal}. We measure distances in pixels relative to the \(1680\times1050\) image canvas. Unless stated otherwise, the 95\% intervals in Sections~\ref{sec:supp_mech_meanmode} to \ref{sec:supp_mech_valley} resample physical images with 10,000 replicates and pool seeds within each image. 

\subsection{The regression output is closer to the map mean than to the mode}
\label{sec:supp_mech_meanmode}

In the four Smooth-L1 models, the regression output lies closer to the mean in 81\% to 85\% of slots. For multimodal slots, this proportion rises to 86\% to 91\%. The distance to the mode is 1.8 to 2.9 times the distance to the mean. 

As the regression head gains more spatial information, its output aligns closer to the mean. The mean distance drops from 120 px for Reg+Map to 50 px for FullMap. Conversely, the mode distance does not shrink proportionally, decreasing only from 211 px to 147 px. This aligns with the behavior of a squared-loss estimator. However, distances alone do not prove that the regression head estimates the map mean.

With a mixture-density regression head (FullMap-MDN, Section~\ref{sec:supp_mech_mdn}), the pattern reverses. Its output aligns closer to the mean in only 46\% of slots, and in 38\% of multimodal slots.

\begin{table*}[t]
\centering
\small
\caption{Distance (pixels) from the native regression output to the map mean (SR, $\kappa=1$) and to
the map mode (argmax). COCO-TP test set, seeds 3 to 5, $\tau=0.5$. All decoded slots that have a human
fixation with the same index. ``Closer to mean'' is the share of slots in which the regression output
is closer to the mean than to the mode. ``Multimodal'' is the share of slots classified as multimodal.}
\label{tab:supp_mean_mode}
\setlength{\tabcolsep}{4pt}
\begin{adjustbox}{max width=\linewidth}
\begin{tabular}{lc ccc ccc}
\toprule
& & \multicolumn{3}{c}{All slots} & \multicolumn{3}{c}{Multimodal slots} \\
\cmidrule(lr){3-5}\cmidrule(lr){6-8}
Model & Multimodal & To mean & To mode & Closer to mean & To mean & To mode & Closer to mean \\
\midrule
Reg+Map      & 0.41 & 120.4 & 210.7 & 0.807 & 118.3 & 272.8 & 0.883 \\
Equal access & 0.42 &  97.3 & 187.7 & 0.813 & 109.7 & 270.0 & 0.894 \\
SMR          & 0.44 &  63.9 & 147.7 & 0.819 &  76.4 & 214.7 & 0.855 \\
FullMap      & 0.51 &  50.4 & 147.4 & 0.848 &  55.2 & 201.7 & 0.905 \\
\midrule
FullMap-MDN  & 0.43 & 108.8 &  92.6 & 0.463 & 175.0 & 152.3 & 0.376 \\
\bottomrule
\end{tabular}
\end{adjustbox}
\end{table*}

For diffuse maps, the mean lies near the image center. Thus, being ``closer to the mean'' may partly reflect a center bias in the regression output. The mixture-density control uses identical data, model structure, and map supervision. Although its retrained weights and predicted maps differ, it does not exhibit this pattern. Therefore, pure center bias cannot fully explain the phenomenon.

\subsection{Readout gains and map concentration}
\label{sec:supp_mech_entropy}

We test two hypotheses. First, if the regression head fails to fully exploit the map, the readout loss (SR \(\kappa=1\) minus native) should be largest for highly concentrated maps. Second, if sharpening shifts the mean toward high-probability regions, its gain (SR \(\kappa=0.5\) minus SR \(\kappa=1\)) may grow as the map spreads. However, high entropy can arise from multiple peaks or from a flat, wide unimodal map. Thus, the second hypothesis might not hold.

We rank scanpaths by their mean slot entropy and split them into four equal quantiles. Q1 contains the most concentrated maps, while Q4 contains the most spread maps. Each group holds 25\% of the scanpaths. We then compute both ScanMatch gains for each group.

Initially, both trends appear across all models. For Reg+Map, the readout gain decreases from Q1 to Q4: \(+0.147\), \(+0.135\), \(+0.105\), and \(+0.048\). The sharpening gain increases: \(+0.017\), \(+0.027\), \(+0.036\), and \(+0.037\).

However, map entropy correlates with decoded length (Spearman \(\rho=0.50\) to \(0.61\) across models). Q4 contains many scanpaths that fail to stop, with a mean length of 8.8 versus 2.8 to 3.3 in other groups. Therefore, Table~\ref{tab:supp_entropy} restricts to scanpaths with a decoded length of 2 to 6, comprising 87\% to 90\% of all scanpaths (human mean 3.9). We re-split this subset into four groups.

In this subset, mean lengths are similar across groups (Reg+Map: 2.83, 2.99, 3.20, and 3.59). Both trends persist across all Smooth-L1 models. The entropy-length correlation drops to \(\rho=0.32\) to \(0.44\). The length restriction weakens this confounding factor but does not eliminate it.

\begin{table}[t]
\centering
\small
\caption{ScanMatch gains by quartile of the scanpath's mean map entropy (Q1 most concentrated, Q4 most
spread). Only scanpaths with a decoded length of 2 to 6. ``Read'' is SR ($\kappa=1$) minus native.
``Sharpen'' is SR ($\kappa=0.5$) minus SR ($\kappa=1$). COCO-TP test set, seeds 3 to 5, $\tau=0.5$.
$^\dagger$ marks entries whose 95\% interval (images) contains zero. All other intervals exclude zero.}
\label{tab:supp_entropy}
\footnotesize
\setlength{\tabcolsep}{4pt}
\begin{adjustbox}{max width=\linewidth}
\begin{tabular}{llcccc}
\toprule
Model & Gain & Q1 & Q2 & Q3 & Q4 \\
\midrule
\multirow{2}{*}{Reg+Map}      & Read    & $+0.150$ & $+0.145$ & $+0.110$ & $+0.075$ \\
                              & Sharpen & $+0.017$ & $+0.026$ & $+0.034$ & $+0.047$ \\
\multirow{2}{*}{Equal access} & Read    & $+0.097$ & $+0.090$ & $+0.076$ & $+0.059$ \\
                              & Sharpen & $+0.018$ & $+0.030$ & $+0.046$ & $+0.046$ \\
\multirow{2}{*}{SMR}          & Read    & $+0.039$ & $+0.033$ & $+0.030$ & $+0.012$ \\
                              & Sharpen & $+0.014$ & $+0.024$ & $+0.035$ & $+0.047$ \\
\multirow{2}{*}{FullMap}      & Read    & $+0.024$ & $+0.027$ & $+0.017$ & $+0.005^\dagger$ \\
                              & Sharpen & $+0.014$ & $+0.034$ & $+0.047$ & $+0.048$ \\
\midrule
\multirow{2}{*}{FullMap-MDN}  & Read    & $-0.005^\dagger$ & $-0.018$ & $-0.027$ & $-0.042$ \\
                              & Sharpen & $+0.012$ & $+0.024$ & $+0.039$ & $+0.050$ \\
\bottomrule
\end{tabular}
\end{adjustbox}
\end{table}

The same pattern appears for single fixations. We compute entropy quartiles per fixation index (1 to 5, and 6 or later) before pooling them. This ensures each group maintains an identical positional mix. From Q1 to Q4, reading the map (\(\kappa=1\)) reduces the coordinate error of Reg+Map by 69, 60, 50, and 37 px.

In this per-fixation view, the sharpening gain is small and does not rise monotonically (3, 7, 9, and 7 px). Therefore, compared to the read-the-map trend, the sharpening trend at the scanpath level in Table~\ref{tab:supp_entropy} is less robust.

For FullMap-MDN, the read-the-map gain is negative and becomes even more negative with entropy. Its native output outperforms the map mean, and this advantage grows as the map spreads. This aligns with a head that predicts a mode-like location rather than the mean.

\subsection{Regression outputs often fall in low-probability regions of the model's own map}
\label{sec:supp_mech_valley}

Table~\ref{tab:supp_valley} reports how often each readout lands on a low-probability cell, defined as having a probability below 0.1 times the slot's maximum. For Reg+Map, 38\% to 39\% of regression outputs fall on such cells, in both unimodal and multimodal maps. For SR (\(\kappa=0.5\)), this share drops to 0.7\% in unimodal slots and 12\% in multimodal slots.

Therefore, the regression output does more than just average between peaks. It frequently disagrees with the high-probability regions of the model's own map. However, this measure is relative to the model's map alone. It does not directly prove that the regression output lies far from human fixations. That evidence comes from the Hit@3 analysis below and the per-fixation coordinate error in Section~\ref{sec:supp_mech_entropy}.

Richer inputs to the regression head reduce this share to 19\% to 26\% for FullMap, though it remains higher than that of SR. Plain expectation falls on low-probability cells more frequently in multimodal maps (23\% to 26\%). Sharpening specifically targets this limitation.

\begin{table}[t]
\centering
\small
\caption{Share of readouts that fall on a cell with probability below 0.1 times the maximum. COCO-TP
test set, seeds 3 to 5, $\tau=0.5$. The argmax lies on the maximum by definition, so its share is 0.}
\label{tab:supp_valley}
\footnotesize
\setlength{\tabcolsep}{3pt}
\begin{adjustbox}{max width=\linewidth}
\begin{tabular}{l ccc ccc}
\toprule
& \multicolumn{3}{c}{Unimodal slots} & \multicolumn{3}{c}{Multimodal slots} \\
\cmidrule(lr){2-4}\cmidrule(lr){5-7}
Model & Native & $\kappa=1$ & $\kappa=0.5$ & Native & $\kappa=1$ & $\kappa=0.5$ \\
\midrule
Reg+Map      & 0.386 & 0.110 & 0.007 & 0.379 & 0.233 & 0.118 \\
Equal access & 0.303 & 0.110 & 0.009 & 0.407 & 0.245 & 0.151 \\
SMR          & 0.207 & 0.125 & 0.008 & 0.298 & 0.261 & 0.123 \\
FullMap      & 0.189 & 0.153 & 0.009 & 0.264 & 0.243 & 0.119 \\
\midrule
FullMap-MDN  & 0.029 & 0.106 & 0.010 & 0.128 & 0.280 & 0.156 \\
\bottomrule
\end{tabular}
\end{adjustbox}
\end{table}

The readouts also differ in early target hits. With the first fixation removed and no termination calibration applied, Hit@3 measures the probability that fixations 2 to 4 fall inside the target box. Hit@3 scores are 0.239 for the native Reg+Map output, 0.574 for SR (\(\kappa=1\)), 0.778 for SR (\(\kappa=0.5\)), and 0.875 for argmax. The human baseline is 0.832 (Section~4.5 of the main paper and Section~\ref{sec:supp_baselines}).

\subsection{Examples}
\label{sec:supp_mech_figure}

\begin{figure*}[t]
\centering
\includegraphics[width=\linewidth]{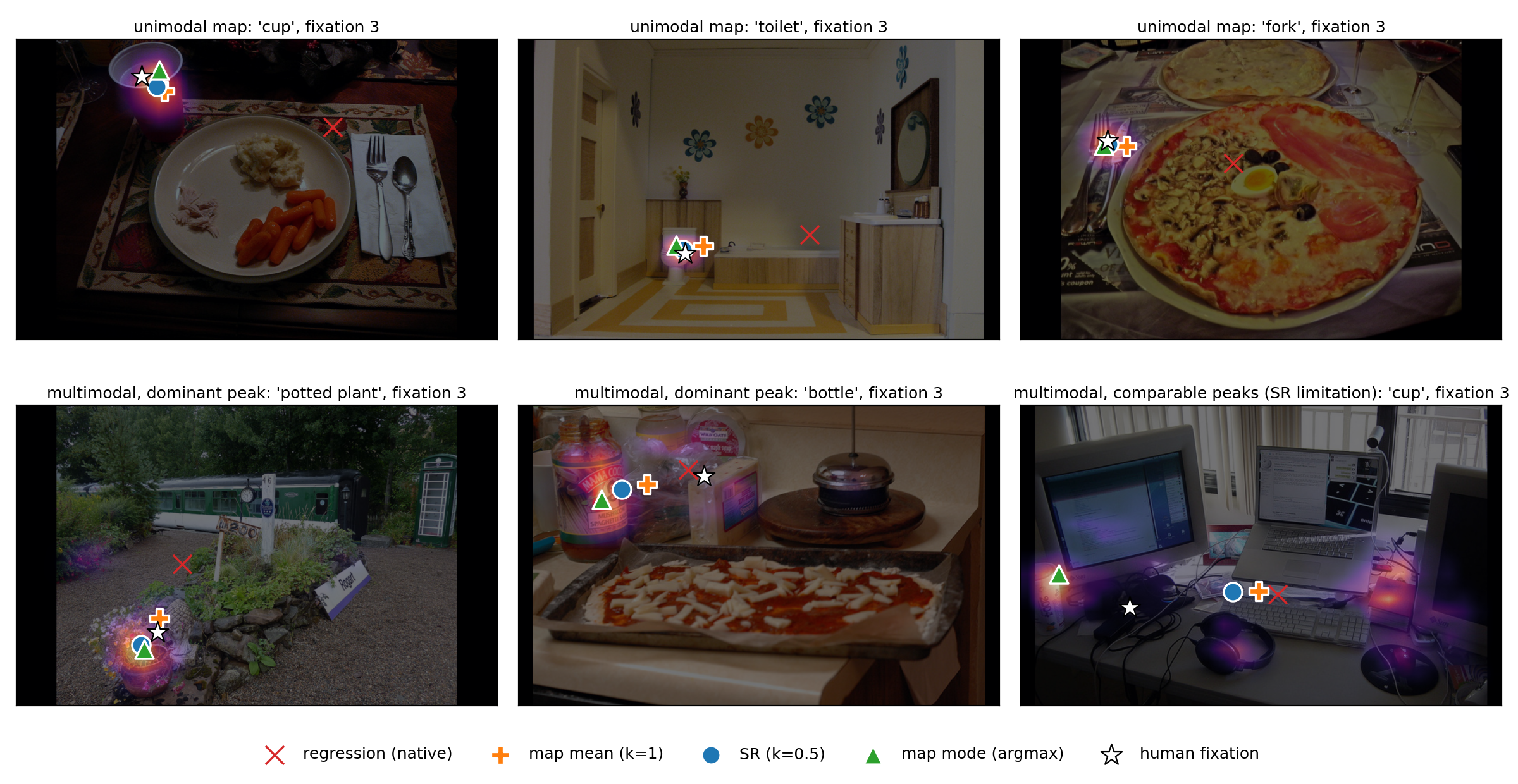}
\caption{\textbf{Readout positions on the learned spatial map.} Results are on the COCO-TP test set using Reg+Map (seed 3). The overlay shows the slot's spatial map. We select examples using fixed rules independent of human fixations. The regression output falls on a cell with a probability below 0.1 times the maximum. We rank candidates by the distance between the regression output and SR (\(\kappa=0.5\)) and select one example per image. \emph{Top:} Unimodal maps (one peak, normalized entropy below 0.55). The map mean, SR, and the mode cluster closely, while the regression output falls in a low-probability region. \emph{Bottom left and middle:} Multimodal maps with a dominant peak. The peak ratio ranges from 0.3 to 0.6, the peaks are at least 8 cells apart, and SR falls on a cell with a probability of at least 0.5 times the maximum. The regression output falls between the peaks, whereas SR moves to the dominant peak. \emph{Bottom right:} Two peaks of similar height (peak ratio at least 0.8, with SR also on a low-probability cell). Here, sharpening with \(\kappa=0.5\) proves insufficient. SR remains between the peaks along with the mean, while argmax selects one peak. We show the human fixation with the same index for reference only. Candidate counts: 770 unimodal, 51 with a dominant peak, and 38 with two similar peaks.}
\label{fig:supp_mechanism}
\end{figure*}

Figure~\ref{fig:supp_mechanism} illustrates three map geometries. In the bottom-right
example, the sharpened expectation still lies between similarly high peaks.
This illustrates why a fixed temperature need not be preferable to argmax
in every case (Section~\ref{sec:supp_readout_variants}).

\subsection{Control experiment: a mixture-density regression head}
\label{sec:supp_mech_mdn}

If this gap stems from point regression collapsing multiple locations into a single coordinate, a regression head capable of representing multiple locations should alter the native output. Consequently, it would reduce the advantage of sharpened readout.

\paragraph{Model.}
FullMap-MDN uses a mixture-density head~\cite{bishop1994mixture,chai2019multipath}. It retains the shared structure, the regression input, and the map supervision of FullMap. This regression input includes the full spatial distribution of the slot. It replaces only the output part of the regression head. The last linear layer outputs \(K=4\) diagonal Gaussian components. Each component has a weight logit, a two-dimensional mean passed through a sigmoid, and a two-dimensional log standard deviation clipped to \([-3, 0]\) in normalized coordinates.

We replace the Smooth-L1 coordinate loss with the mixture negative log-likelihood. The supervised coordinate remains the center of the cell containing the human fixation. The weight \(\lambda_x=1\) remains unchanged. The native coordinate takes the mean of the Gaussian component with the largest mixture weight. Because the components have different variances and may overlap, this rule does not guarantee the global mode of the mixture.

This control alters three elements simultaneously: the output parameterization, the training loss, and the inference-time component selection rule. We also retrain the model jointly. Gradients through the regression pathway thus affect both the map and the shared modules. The control tests whether allowing the regression head to represent several locations reduces the gap. However, it cannot disentangle the effects of the loss, the parameterization, and the readout rule.

We follow the FullMap training, checkpoint selection, and evaluation protocol on each dataset. The datasets are COCO-TP (seeds 3 to 5), AiR-D (seeds 3 to 5), and COCO-TA (seeds 3 to 5, using the readout protocol in Section~\ref{sec:supp_replication}).

\paragraph{Results.}
Table~\ref{tab:supp_mdn} compares FullMap-MDN with FullMap.
\begin{itemize}
\item \emph{Within checkpoint.} The gain of SR over the native output drops from \(+0.049\) to \(+0.008\) on COCO-TP (about 85\%). We detect no significant difference on AiR-D. On TA, the difference is also not significant after termination calibration.
\item \emph{Across checkpoints.} On COCO-TP, the native output of FullMap-MDN outperforms that of FullMap at both thresholds (ScanMatch \(+0.049\) under TC). On AiR-D, the advantage is significant only under TC (\(+0.016\)). At the default threshold, the interval crosses zero. After termination calibration, we detect no significant ScanMatch difference between the FullMap-MDN native output and FullMap + SR on TP or AiR-D. On TA, FullMap + SR performs significantly better (\(+0.016\)), and the FullMap-MDN native output does not outperform FullMap.
\item \emph{Map quality.} Current measures show no evidence of consistent map improvement. On COCO-TP, the development-set fixation-map NLL of FullMap-MDN is slightly worse (4.810 vs. 4.762). Under SR, we detect no significant ScanMatch difference between the two models on COCO-TP (\(+0.004\), \([-0.010, 0.018]\)) or TA. Consequently, the native output gain on COCO-TP is hard to attribute to a better map. On AiR-D, the NLL is slightly better (5.607 vs. 5.626), and SR on FullMap-MDN is also slightly better (\(+0.011\), \([0.001, 0.020]\)). Therefore, part of the AiR-D gain may stem from the map.
\end{itemize}

\begin{table*}[t]
\centering
\caption{Mixture-density control (ScanMatch). FullMap uses a Smooth-L1 regression head that reads the
full spatial map. FullMap-MDN is the same model with a $K=4$ mixture-density head. Seeds 3 to 5 on every
dataset. TC is the threshold calibrated on the development set for each checkpoint. Brackets give the
95\% paired interval over seeds $\times$ images (20,000 replicates). n/a means not computed. Following the protocol of each setting, COCO-TP and AiR-D means average over scanpaths and COCO-TA means average first within each image. All paired differences average first within each image. The last row
is the development-set fixation-map NLL at the selected epoch (lower is better).}
\label{tab:supp_mdn}
\small
\setlength{\tabcolsep}{5pt}
\renewcommand{\arraystretch}{1.1}
\begin{tabular}{l cc cc c}
\toprule
& \multicolumn{2}{c}{COCO-TP} & \multicolumn{2}{c}{AiR-D} & COCO-TA \\
\cmidrule(lr){2-3}\cmidrule(lr){4-5}\cmidrule(lr){6-6}
Comparison & $\tau=0.5$ & TC & $\tau=0.5$ & TC & TC \\
\midrule
FullMap native (mean)     & 0.436 & 0.453 & 0.289 & 0.306 & 0.309 \\
FullMap SR (mean)         & 0.486 & 0.505 & 0.302 & 0.321 & 0.328 \\
FullMap-MDN native (mean) & 0.483 & 0.502 & 0.296 & 0.321 & 0.312 \\
FullMap-MDN SR (mean)     & 0.490 & 0.510 & 0.303 & 0.328 & 0.319 \\
\midrule
FullMap: SR $-$ native
  & \makecell{$+0.049$\\{\scriptsize$[+0.025,+0.068]$}} & \makecell{$+0.052$\\{\scriptsize$[+0.031,+0.069]$}}
  & \makecell{$+0.012$\\{\scriptsize$[+0.005,+0.019]$}} & \makecell{$+0.013$\\{\scriptsize$[+0.006,+0.022]$}}
  & \makecell{$+0.018$\\{\scriptsize$[+0.014,+0.023]$}} \\
\addlinespace[2pt]
FullMap-MDN: SR $-$ native
  & \makecell{$+0.008$\\{\scriptsize$[+0.002,+0.013]$}} & \makecell{$+0.008$\\{\scriptsize$[+0.001,+0.014]$}}
  & \makecell{$+0.006$\\{\scriptsize$[-0.003,+0.016]$}} & \makecell{$+0.007$\\{\scriptsize$[-0.005,+0.020]$}}
  & \makecell{$+0.007$\\{\scriptsize$[-0.002,+0.015]$}} \\
\addlinespace[2pt]
MDN native $-$ FullMap native
  & \makecell{$+0.046$\\{\scriptsize$[+0.019,+0.069]$}} & \makecell{$+0.049$\\{\scriptsize$[+0.026,+0.072]$}}
  & \makecell{$+0.009$\\{\scriptsize$[-0.010,+0.028]$}} & \makecell{$+0.016$\\{\scriptsize$[+0.002,+0.032]$}}
  & \makecell{$+0.003$\\{\scriptsize$[-0.007,+0.013]$}} \\
\addlinespace[2pt]
MDN native $-$ FullMap SR
  & n/a & \makecell{$-0.003$\\{\scriptsize$[-0.015,+0.008]$}}
  & n/a & \makecell{$+0.003$\\{\scriptsize$[-0.014,+0.020]$}}
  & \makecell{$-0.016$\\{\scriptsize$[-0.026,-0.005]$}} \\
\addlinespace[2pt]
MDN SR $-$ FullMap SR
  & n/a & \makecell{$+0.004$\\{\scriptsize$[-0.010,+0.018]$}}
  & n/a & \makecell{$+0.011$\\{\scriptsize$[+0.001,+0.020]$}}
  & \makecell{$-0.009$\\{\scriptsize$[-0.019,+0.001]$}} \\
\midrule
Map NLL (FullMap / MDN) & \multicolumn{2}{c}{4.762 / 4.810} & \multicolumn{2}{c}{5.626 / 5.607} & n/a \\
\bottomrule
\end{tabular}
\end{table*}

\paragraph{Interpretation.}
Providing more spatial information to the regression head (from Reg+Map to FullMap) narrows the gap between map reading (\(\kappa=1\)) and regression. Changing the output parameterization, training loss, and component selection rule (FullMap-MDN) further reduces the sharpening advantage on COCO-TP and AiR-D. Together, these results suggest that the gap depends on two factors. The first is the regression head's ability to utilize spatial information. The second is how the model compresses this information into a single coordinate.

On target-absent search (TA), this control narrows the within-checkpoint gap but fails to improve the native output. This weaker effect on TA may stem from map multimodality. However, we do not test this explanation using cross-condition geometric statistics.

Under termination calibration, ScanMatch reveals no difference between FullMap + SR and the native FullMap-MDN output on TP and AiR-D. On TA, FullMap + SR performs significantly better. Because SR requires no retraining, it provides a practical alternative.

\section{Readout controls on AiR-D and target-absent search}
\label{sec:supp_replication}

We extend the COCO-TP controls from Section~4.5 of the main paper to AiR-D
(equal access, SMR, and FullMap) and COCO-TA (equal access and FullMap).

\subsection{Setup}
\label{sec:supp_repl_setup}

\paragraph{Models.}
We evaluate three controls defined in Section~4.5 of the main paper. They modify only the location pathway and keep all other modules and losses unchanged.
\begin{itemize}
\item Equal access. Building upon Reg+Map, the decoder attends directly to task-fused features.
\item SMR. We replace the linear coordinate head with a two-layer MLP. This MLP regresses coordinates from the slot state and task-fused features pooled with the slot's spatial map using Smooth-L1.
\item FullMap. Building upon SMR, the regression head additionally receives the full spatial distribution of the slot.
\end{itemize}
On AiR-D, we include Reg+Map and TCSD as references using their frozen checkpoints from the main paper. On COCO-TA, we train only two controls: Equal access and FullMap.

\paragraph{Training.}
Each dataset retains the training configuration of the main paper. AiR-D uses the image-disjoint split with a batch size of 8, a maximum of 12 epochs, and a patience of 5. COCO-TA uses its respective TA split and configuration. We select checkpoints on each development set using the main paper's rule. Both datasets use seeds 3 to 5.

\paragraph{Readout and termination.}
For each checkpoint, we compare the native readout, SR (\(\kappa=1\)), and the locked SR (\(\kappa=0.5\)). We evaluate these readouts under the default threshold (\(\tau=0.5\)) and the development-set threshold calibrated via the main paper's TC rule. Within a given checkpoint and threshold, all readouts share identical lengths and durations.

\paragraph{Evaluation.}
For AiR-D, we follow the evaluation conventions of the main paper. The canvas size is \(512\times384\). We compute sequence metrics per scanpath and saliency metrics per question-image group. As a sanity check, the native decoded lengths of Reg+Map and TCSD at \(\tau=0.5\) perfectly match the main paper's frozen evaluation files across all scanpaths. For COCO-TA, we adopt the main paper's TA conventions. We first average metrics within each image and then across images, retaining the first fixation.

\paragraph{Statistics.}
Paired differences are computed for the same checkpoint and the same threshold, as in
Section~\ref{sec:supp_eval_stats}.

\subsection{Summary: read-the-map gain and sharpening gain}
\label{sec:supp_repl_summary}

Table~\ref{tab:supp_repl_summary} decomposes the SR gain into two components. The read-the-map gain is SR (\(\kappa=1\)) minus the native output. It measures the added value of directly reading the spatial map over the regression head. The sharpening gain is SR (\(\kappa=0.5\)) minus SR (\(\kappa=1\)). It captures the additional benefit of shifting the readout toward high-probability regions. To ensure consistency across settings, we apply the default threshold \(\tau=0.5\) here.

Trends remain similar across the three settings but are not strictly monotonic. On COCO-TP, the read-the-map gain shrinks progressively from Reg+Map to FullMap. On AiR-D, this gain is small overall. Equal access (\(+0.003\), not detected) is lower than SMR (\(+0.007\)). For FullMap, the read-the-map gain remains detectable on COCO-TP (\(+0.016\)) but not on AiR-D or COCO-TA. 

The sharpening gain is detectable across all models and settings. It ranges from 0.027 to 0.032 on COCO-TP and 0.013 to 0.016 on AiR-D and COCO-TA. Combined with the map-only variant (\(+0.026\), Section~4.5 of the main paper), the COCO-TP values yield the 0.026 to 0.032 range reported in Section~5 of the main paper.

\begin{table*}[t]
\centering
\caption{Read-the-map gain (``Read'') and sharpening gain (``Sharpen'') in the three evaluation
settings. ScanMatch, $\tau=0.5$, seeds 3 to 5. Brackets give the 95\% paired interval over seeds
$\times$ images. $^\dagger$ marks intervals that contain zero. n/a means that this comparison was not run
for this model in this setting. The COCO-TP values are those cited in Section~5 of the main paper.}
\label{tab:supp_repl_summary}
\small
\setlength{\tabcolsep}{5pt}
\renewcommand{\arraystretch}{1.1}
\begin{tabular}{l cc cc cc}
\toprule
& \multicolumn{2}{c}{COCO-TP} & \multicolumn{2}{c}{AiR-D} & \multicolumn{2}{c}{COCO-TA} \\
\cmidrule(lr){2-3}\cmidrule(lr){4-5}\cmidrule(lr){6-7}
Model & Read & Sharpen & Read & Sharpen & Read & Sharpen \\
\midrule
Reg+Map
  & \makecell{$+0.106$\\{\scriptsize$[+0.076,+0.140]$}} & \makecell{$+0.028$\\{\scriptsize$[+0.023,+0.034]$}}
  & \makecell{$+0.021$\\{\scriptsize$[+0.014,+0.029]$}} & \makecell{$+0.016$\\{\scriptsize$[+0.010,+0.023]$}}
  & n/a & n/a \\
\addlinespace[2pt]
Equal access
  & \makecell{$+0.071$\\{\scriptsize$[+0.049,+0.098]$}} & \makecell{$+0.032$\\{\scriptsize$[+0.026,+0.037]$}}
  & \makecell{$+0.003^\dagger$\\{\scriptsize$[-0.005,+0.011]$}} & \makecell{$+0.015$\\{\scriptsize$[+0.008,+0.021]$}}
  & \makecell{$+0.012$\\{\scriptsize$[+0.009,+0.014]$}} & \makecell{$+0.015$\\{\scriptsize$[+0.013,+0.017]$}} \\
\addlinespace[2pt]
SMR
  & \makecell{$+0.028$\\{\scriptsize$[+0.019,+0.038]$}} & \makecell{$+0.027$\\{\scriptsize$[+0.021,+0.034]$}}
  & \makecell{$+0.007$\\{\scriptsize$[+0.002,+0.012]$}} & \makecell{$+0.013$\\{\scriptsize$[+0.007,+0.019]$}}
  & n/a & n/a \\
\addlinespace[2pt]
FullMap
  & \makecell{$+0.016$\\{\scriptsize$[+0.003,+0.029]$}} & \makecell{$+0.032$\\{\scriptsize$[+0.022,+0.042]$}}
  & \makecell{$-0.001^\dagger$\\{\scriptsize$[-0.005,+0.003]$}} & \makecell{$+0.013$\\{\scriptsize$[+0.007,+0.019]$}}
  & \makecell{$-0.001^\dagger$\\{\scriptsize$[-0.003,+0.001]$}} & \makecell{$+0.016$\\{\scriptsize$[+0.013,+0.018]$}} \\
\bottomrule
\end{tabular}
\end{table*}

We provide three remarks. First, on TA we do not compute SR (\(\kappa=1\)) for Reg+Map or train SMR. Consequently, these two rows are not applicable. Table~3 of the main paper reports the total Reg+Map gain on TA. Second, at \(\tau=0.5\), the decoded TA length approaches the 16-slot limit (15.3 to 15.8). Section~\ref{sec:supp_repl_ta} reports the results under calibrated termination. Third, significant gains within individual models do not imply significant differences between models. 

\subsection{AiR-D}
\label{sec:supp_repl_aird}

On AiR-D, SR (\(\kappa=0.5\)) significantly improves ScanMatch over the native output across all five models (Table~\ref{tab:supp_repl_aird}). This holds at both the default and calibrated thresholds. The gain decreases from \(+0.045\) for Reg+Map (TC) to \(+0.013\) for FullMap (TC).

\begin{table*}[t]
\centering
\caption{ScanMatch on the AiR-D test set, seeds 3 to 5. Each cell gives $\tau=0.5$ / TC. Length is the
mean decoded length (human mean on the development set 10.1). The last two columns give the paired
difference of SR ($\kappa=0.5$) minus the native output with its 95\% interval. Reg+Map and TCSD use the
frozen checkpoints of the main paper. The native output of TCSD is SR ($\kappa=1$). Following the AiR-D
convention of the main paper, the means average over scanpaths. The paired differences average first
within each image. Subtracting two mean columns therefore need not give the paired difference.}
\label{tab:supp_repl_aird}
\small
\setlength{\tabcolsep}{4pt}
\renewcommand{\arraystretch}{1.1}
\begin{tabular}{l c ccc cc}
\toprule
Model & Length & Native & SR ($\kappa=1$) & SR ($\kappa=0.5$) & SR $-$ native, $\tau=0.5$ & SR $-$ native, TC \\
\midrule
Reg+Map      & 12.46 / 10.08 & 0.252 / 0.280 & 0.275 / 0.308 & 0.292 / 0.327
  & $+0.038$ {\scriptsize$[+0.027,+0.049]$} & $+0.045$ {\scriptsize$[+0.033,+0.057]$} \\
Equal access & 11.91 / 10.07 & 0.283 / 0.302 & 0.287 / 0.311 & 0.302 / 0.327
  & $+0.017$ {\scriptsize$[+0.007,+0.028]$} & $+0.023$ {\scriptsize$[+0.012,+0.034]$} \\
SMR          & 12.82 / 10.02 & 0.262 / 0.297 & 0.269 / 0.307 & 0.281 / 0.320
  & $+0.020$ {\scriptsize$[+0.012,+0.028]$} & $+0.024$ {\scriptsize$[+0.015,+0.034]$} \\
FullMap      & 11.61 / 10.16 & 0.289 / 0.306 & 0.289 / 0.307 & 0.302 / 0.321
  & $+0.012$ {\scriptsize$[+0.005,+0.019]$} & $+0.013$ {\scriptsize$[+0.006,+0.022]$} \\
TCSD         & 12.80 / 10.11 & 0.268 / 0.307 & 0.268 / 0.307 & 0.282 / 0.323
  & $+0.013$ {\scriptsize$[+0.007,+0.020]$} & $+0.015$ {\scriptsize$[+0.008,+0.022]$} \\
\bottomrule
\end{tabular}
\end{table*}

\paragraph{Other metrics.}
Under TC, SR lowers SED over the native output across all five models (\(-0.12\) to \(-0.54\), with paired intervals excluding zero). The MultiMatch results are mixed. TCSD improves significantly (\(+0.005\)). We detect no difference for Reg+Map, SMR, or FullMap. Equal access degrades significantly (\(-0.009\)). 

For the controls of this section, we compute only the means of KLD, CC, and NSS on AiR-D, without paired intervals. For Reg+Map, Section~4.3 of the main paper reports paired intervals for all six metrics. On average, KLD decreases across all five models (e.g., from 10.24 to 7.70 for Reg+Map and from 8.69 to 7.88 for FullMap), while CC increases across all five.

\subsection{COCO-TA}
\label{sec:supp_repl_ta}

Under calibrated termination on COCO-TA, SR (\(\kappa=0.5\)) significantly improves all six metrics over the native output for both controls. All three seeds agree in direction (Table~\ref{tab:supp_repl_ta}). Specifically, ScanMatch improves by \(+0.042\) for equal access and \(+0.018\) for FullMap. For reference, the main paper reports an SR gain of \(+0.045\) for Reg+Map on TA under the same condition (5 seeds).

\begin{table*}[t]
\centering
\caption{ScanMatch on the COCO-TA test set, seeds 3 to 5. Following the TA convention of the main paper,
both the means and the paired differences are first averaged within each image and then across images.
Each cell gives $\tau=0.5$ / TC. Length is the mean decoded length (human mean on the development set
5.6). The last two columns give the paired difference of SR ($\kappa=0.5$) minus the native output with
its 95\% interval.}
\label{tab:supp_repl_ta}
\small
\setlength{\tabcolsep}{4pt}
\renewcommand{\arraystretch}{1.1}
\begin{tabular}{l c ccc cc}
\toprule
Model & Length & Native & SR ($\kappa=1$) & SR ($\kappa=0.5$) & SR $-$ native, $\tau=0.5$ & SR $-$ native, TC \\
\midrule
Equal access & 15.29 / 5.93 & 0.144 / 0.288 & 0.155 / 0.314 & 0.170 / 0.331
  & $+0.027$ {\scriptsize$[+0.023,+0.030]$} & $+0.042$ {\scriptsize$[+0.037,+0.048]$} \\
FullMap      & 15.75 / 5.90 & 0.148 / 0.309 & 0.147 / 0.311 & 0.163 / 0.328
  & $+0.015$ {\scriptsize$[+0.012,+0.018]$} & $+0.018$ {\scriptsize$[+0.014,+0.023]$} \\
\bottomrule
\end{tabular}
\end{table*}

\paragraph{The two parts of the gain.}
Under TC, the read-the-map gain is \(+0.025\) (\([+0.022, +0.029]\)) for equal access and \(+0.001\) (\([-0.002, +0.004]\), not detected) for FullMap. This matches the pattern at \(\tau=0.5\) in Section~\ref{sec:supp_repl_summary}. Once the regression head reads the full spatial map, the read-the-map gain becomes undetectable, while the sharpening gain remains.

\paragraph{Other metrics.}
Under TC, SR significantly improves MultiMatch, SED, KLD, CC, and NSS over the native output for both models. At \(\tau=0.5\), we detect no difference for CC and NSS. At this threshold, the decoded length approaches the slot limit. Consequently, the predicted saliency map is built from too many fixations. This may reduce the sensitivity of these two metrics to individual locations.

The same evaluation also compares the argmax readout, with results provided in Section~\ref{sec:supp_readout_variants}.

\subsection{Summary and limitations}
\label{sec:supp_repl_limits}

In the three evaluated settings, plain expectation yields only a small ScanMatch gain over the native output for FullMap. This gain is detectable on COCO-TP but not on AiR-D or COCO-TA. In contrast, at the default termination threshold, sharpening detectably improves ScanMatch over plain expectation across all three settings on the same checkpoint. This indicates that the location readout still affects sequence prediction even when the regression head receives the full spatial map.

\begin{itemize}
\item Each control in this section uses only 3 seeds. Consequently, the intervals provide limited insight into training randomness.
\item COCO-TP and COCO-TA are two task conditions of the same dataset rather than independent datasets.
\item We did not compute SR (\(\kappa=1\)) for Reg+Map on COCO-TA.
\item On AiR-D, the MultiMatch results are mixed. SR lowers MultiMatch for equal access. For the AiR-D controls of this section, we report only the means of KLD, CC, and NSS without paired intervals.
\item The numbers are not directly comparable across settings. TA metrics are averaged within each image. AiR-D uses a \(512\times384\) evaluation canvas. The human scanpath lengths also differ. Therefore, this section compares directions within each setting rather than values across settings.
\end{itemize}

\section{Readout variants}
\label{sec:supp_readout_variants}

Section~3.3 of the main paper notes that SR approaches the argmax as \(\kappa\rightarrow0\) when the spatial map has a unique maximum. The argmax itself is restricted to grid-cell centers. Section~\ref{sec:supp_readout_selection} describes how we selected \(\kappa=0.5\) on the development set. This section provides two additional analyses. First, we present a paired comparison between the locked SR (\(\kappa=0.5\)) and the argmax across the test sets of the three evaluation settings. Second, we report the full set of MultiMatch sub-dimensions. Section~\ref{sec:supp_base_hat} presents the readout boundary test on HAT-SF.

\paragraph{Conventions.}
The paired comparisons in Section~\ref{sec:supp_var_argmax} share the same checkpoint and threshold. Only the coordinate readout varies, while lengths and durations remain fixed. The table in Section~\ref{sec:supp_var_multimatch} lists various models and termination settings, so this constraint does not apply. Paired intervals follow Section~\ref{sec:supp_eval_stats}.

\subsection{SR ($\kappa=0.5$) and argmax}
\label{sec:supp_var_argmax}

\paragraph{Data sources.}
\begin{itemize}
\item COCO-TP. Reg+Map, seeds 3 to 5, \(\tau=0.5\). We recompute paired differences from the existing per-scanpath and per-group score cache without running any models. The six SR means in the cache (e.g., ScanMatch 0.5013) match those of the same configuration in our other reports.
\item AiR-D. Reg+Map and TCSD, seeds 3 to 5. We compare each model at \(\tau=0.5\) and with TC. We compute paired intervals only for sequence metrics.
\item COCO-TA. We do not compute the argmax readout for Reg+Map. Instead, we report the results of the two control models from Section~\ref{sec:supp_replication} (equal access and FullMap) using seeds 3 to 5 with TC.
\end{itemize}

\begin{table*}[t]
\centering
\caption{SR ($\kappa=0.5$) minus argmax (paired difference with 95\% interval). n/a means that the report
did not compute this item. Eq.\ access: equal access.}
\label{tab:supp_var_argmax}
\footnotesize
\setlength{\tabcolsep}{2.6pt}
\renewcommand{\arraystretch}{1.1}
\begin{tabular}{@{}llc cccccc@{}}
\toprule
Setting & Model & Threshold & MM & SM & SED & KLD & CC & NSS \\
\midrule
COCO-TP & Reg+Map & $\tau=0.5$
  & \makecell{$+0.0090^{\checkmark}$\\{\scriptsize$[+0.0068,+0.0113]$}}
  & \makecell{$+0.0016^\dagger$\\{\scriptsize$[-0.0038,+0.0068]$}}
  & \makecell{$-0.005^\dagger$\\{\scriptsize$[-0.041,+0.032]$}}
  & \makecell{$-0.14^\dagger$\\{\scriptsize$[-0.38,+0.10]$}}
  & \makecell{$-0.008^\dagger$\\{\scriptsize$[-0.027,+0.010]$}}
  & \makecell{$+0.08^\dagger$\\{\scriptsize$[-0.09,+0.24]$}} \\
\addlinespace[2pt]
AiR-D & Reg+Map & $\tau=0.5$
  & \makecell{$+0.0086^{\checkmark}$\\{\scriptsize$[+0.0047,+0.0124]$}}
  & \makecell{$-0.0066^\dagger$\\{\scriptsize$[-0.0181,+0.0048]$}}
  & \makecell{$+0.110^\dagger$\\{\scriptsize$[-0.109,+0.325]$}}
  & n/a & n/a & n/a \\
AiR-D & Reg+Map & TC
  & \makecell{$+0.0041^{\checkmark}$\\{\scriptsize$[+0.0007,+0.0075]$}}
  & \makecell{$-0.0055^\dagger$\\{\scriptsize$[-0.0182,+0.0072]$}}
  & \makecell{$+0.126^\dagger$\\{\scriptsize$[-0.082,+0.332]$}}
  & n/a & n/a & n/a \\
AiR-D & TCSD & $\tau=0.5$
  & \makecell{$+0.0108^{\checkmark}$\\{\scriptsize$[+0.0069,+0.0146]$}}
  & \makecell{$-0.0003^\dagger$\\{\scriptsize$[-0.0135,+0.0133]$}}
  & \makecell{$-0.039^\dagger$\\{\scriptsize$[-0.282,+0.206]$}}
  & n/a & n/a & n/a \\
AiR-D & TCSD & TC
  & \makecell{$+0.0057^{\checkmark}$\\{\scriptsize$[+0.0022,+0.0091]$}}
  & \makecell{$-0.0007^\dagger$\\{\scriptsize$[-0.0149,+0.0135]$}}
  & \makecell{$-0.038^\dagger$\\{\scriptsize$[-0.273,+0.196]$}}
  & n/a & n/a & n/a \\
\addlinespace[2pt]
COCO-TA & Eq.\ access & TC
  & \makecell{$+0.011^{\checkmark}$\\{\scriptsize$[+0.008,+0.013]$}}
  & \makecell{$-0.0060^{\times}$\\{\scriptsize$[-0.0118,-0.0001]$}}
  & \makecell{$+0.076^{\times}$\\{\scriptsize$[+0.019,+0.136]$}}
  & \makecell{$+0.84^{\times}$\\{\scriptsize$[+0.59,+1.09]$}}
  & \makecell{$-0.028^{\times}$\\{\scriptsize$[-0.044,-0.012]$}}
  & \makecell{$-0.11^{\times}$\\{\scriptsize$[-0.19,-0.03]$}} \\
COCO-TA & FullMap & TC
  & \makecell{$+0.012^{\checkmark}$\\{\scriptsize$[+0.008,+0.015]$}}
  & \makecell{$-0.000^\dagger$\\{\scriptsize$[-0.007,+0.006]$}}
  & \makecell{$+0.036^\dagger$\\{\scriptsize$[-0.026,+0.098]$}}
  & \makecell{$+0.72^{\times}$\\{\scriptsize$[+0.45,+1.02]$}}
  & \makecell{$-0.018^{\times}$\\{\scriptsize$[-0.033,-0.002]$}}
  & \makecell{$-0.05^\dagger$\\{\scriptsize$[-0.14,+0.03]$}} \\
\bottomrule
\end{tabular}
\end{table*}

\paragraph{Observations.}
\begin{itemize}
\item MultiMatch. SR is higher than argmax in all seven parallel-model comparisons of Table~\ref{tab:supp_var_argmax}.
\item COCO-TP and AiR-D. We detect no difference in ScanMatch or SED. We also detect no difference in the saliency metrics on COCO-TP. On AiR-D, we report only the means for saliency metrics.
\item COCO-TA, equal-access model. Argmax outperforms SR on ScanMatch, SED, KLD, CC, and NSS.
\item COCO-TA, FullMap. Argmax outperforms SR on KLD and CC. We detect no difference for ScanMatch, SED, or NSS.
\end{itemize}

SR and argmax both read locations from the same spatial map. Both yield large gains over native regression. On COCO-TP, the ScanMatch gain is \(+0.136\) for SR and \(+0.135\) for argmax (seeds 3 to 5, \(\tau=0.5\)). Their difference is much smaller than their difference from native regression and varies by metric. This aligns with Section~\ref{sec:supp_readout_selection}, where the development set fails to reliably separate the sharpened readouts.

\subsection{MultiMatch sub-dimensions}
\label{sec:supp_var_multimatch}

The MultiMatch reported in the main paper is the mean of five sub-dimensions
(Section~\ref{sec:supp_eval_metrics}). Table~\ref{tab:supp_var_multimatch} gives each of them.

\begin{table}[t]
\centering
\caption{MultiMatch sub-dimensions (COCO-TP test set, seeds 3 to 5, mean over scanpaths). Rows without TC
use $\tau=0.5$ or the native stopping rule of each model. The means over seeds 3 to 5 differ slightly from
the 10-seed means of Table~1 of the main paper.}
\label{tab:supp_var_multimatch}
\footnotesize
\setlength{\tabcolsep}{4pt}
\begin{tabular}{@{}l cccccc@{}}
\toprule
Method & Vector & Direction & Length & Position & Duration & MultiMatch \\
\midrule
Reg & 0.926 & 0.543 & 0.865 & 0.865 & 0.715 & 0.783 \\
Reg+Map & 0.925 & 0.655 & 0.888 & 0.887 & 0.724 & 0.816 \\
Reg+Map + SR & 0.931 & 0.703 & 0.899 & 0.918 & 0.725 & 0.835 \\
Reg+Map + SR + TC & 0.926 & 0.716 & 0.894 & 0.920 & 0.725 & 0.836 \\
TCSD & 0.933 & 0.688 & 0.900 & 0.913 & 0.720 & 0.831 \\
TCSD + SR + TC & 0.930 & 0.720 & 0.897 & 0.921 & 0.719 & 0.837 \\
HAT-SF & 0.911 & 0.707 & 0.882 & 0.919 & 0.739 & 0.832 \\
Gazeformer & 0.924 & 0.671 & 0.897 & 0.895 & 0.671 & 0.811 \\
IRL & 0.881 & 0.665 & 0.843 & 0.884 & 0.739 & 0.802 \\
ScanDiff & 0.892 & 0.639 & 0.860 & 0.863 & 0.632 & 0.777 \\
\bottomrule
\end{tabular}
\end{table}

\paragraph{Observations.}
\begin{itemize}
\item For Reg+Map on COCO-TP, the sub-dimension mean gains from native regression to SR stem mainly from direction (\(+0.0486\)) and position (\(+0.0303\)). Length and vector exhibit smaller changes. Because the table lacks argmax sub-dimensions, these changes cannot explain the gap between SR and argmax.
\item Both readouts use identical predicted durations, and the duration score barely changes (0.724 vs. 0.725). However, MultiMatch aligns the sequences before computing each dimension. Consequently, coordinate changes can still affect the duration score via this alignment.
\end{itemize}


\section{Qualitative results}
\label{sec:supp_qualitative}

This section documents the cases and the 92\% group-level result from Section~4.5 of the main paper. We compare Reg+Map + SR + TC against native Reg+Map on the COCO-TP test set. Plots use seed 3 with \(\kappa=0.5\), a calibrated \(\tau=0.47\), and a native \(\tau=0.5\). This comparison evaluates the joint effect of readout and termination rather than an SR-only ablation.
\subsection{Case selection rule}
\label{sec:supp_qual_rule}

The test set contains 326 groups, each corresponding to a target category and an image. Every readout produces one deterministic prediction per group. The gain is the mean ScanMatch difference (SR + TC minus native) over the group's human scanpaths. 

We select two groups nearest the median gain, two nearest the 90th percentile, and one nearest the 10th percentile. We break ties by the group identifier. We exclude groups with fewer than five human scanpaths, though no group met this criterion. Table~\ref{tab:supp_qual_cases} lists the selected groups.

\begin{table}[t]
\centering
\caption{The five cases selected by the rule (median group gain 0.129).}
\label{tab:supp_qual_cases}
\footnotesize
\setlength{\tabcolsep}{5pt}
\begin{tabular}{@{}cllcc@{}}
\toprule
Case & Quantile & Target & Image & ScanMatch gain \\
\midrule
1 & Median & bottle & 000000413247 & $+0.130$ \\
2 & Median & tv & 000000401010 & $+0.127$ \\
3 & 90th percentile & toilet & 000000241602 & $+0.291$ \\
4 & 90th percentile & keyboard & 000000055323 & $+0.287$ \\
5 & 10th percentile & potted plant & 000000228335 & $+0.009$ \\
\bottomrule
\end{tabular}
\end{table}

\subsection{Qualitative cases}
\label{sec:supp_qual_cases}

\begin{figure*}[t]
\centering
\includegraphics[width=\linewidth]{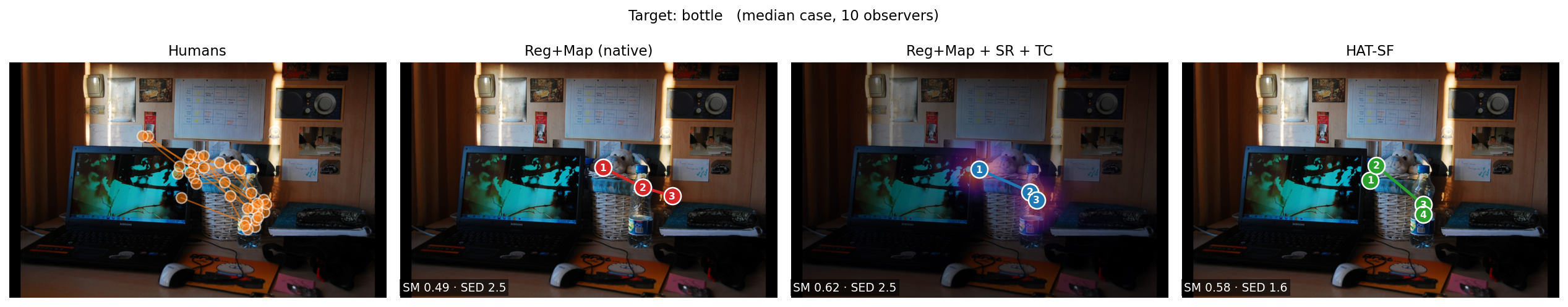}
\caption{Bottle: group near the median ScanMatch gain. Shared panel
conventions are in Section~\ref{sec:supp_qual_cases}.}
\label{fig:supp_qual_case1}
\end{figure*}

\begin{figure*}[t]
\centering
\includegraphics[width=\linewidth]{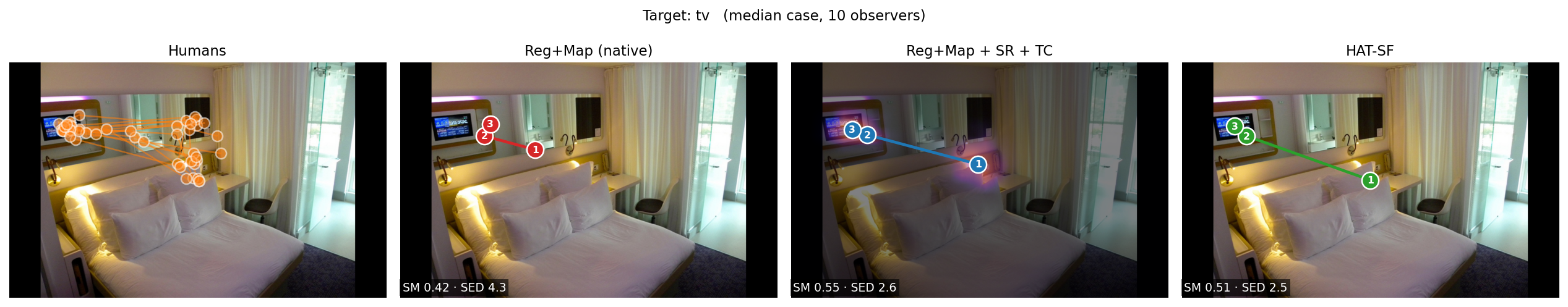}
\caption{TV: group near the median ScanMatch gain.}
\label{fig:supp_qual_case2}
\end{figure*}

\begin{figure*}[t]
\centering
\includegraphics[width=\linewidth]{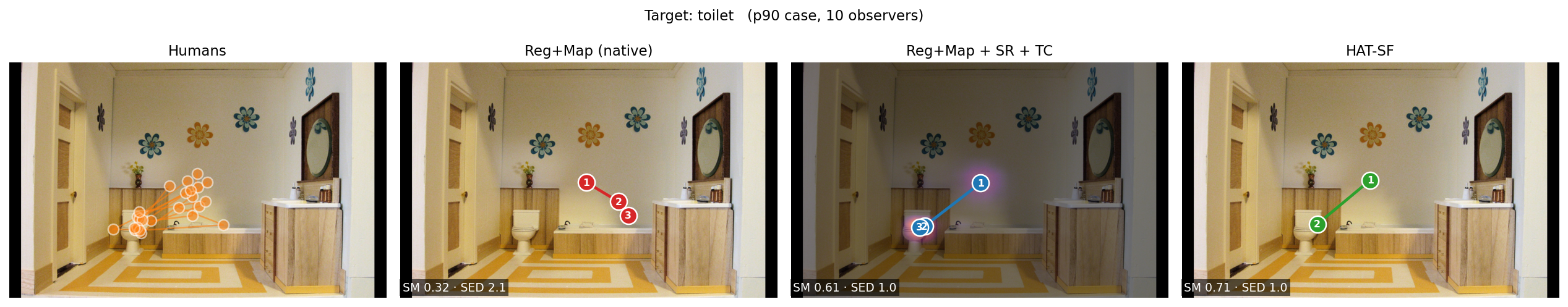}
\caption{Toilet: group near the 90th-percentile gain. SR + TC shifts
fixations towards the target without changing the decoded length.}
\label{fig:supp_qual_case3}
\end{figure*}

\begin{figure*}[t]
\centering
\includegraphics[width=\linewidth]{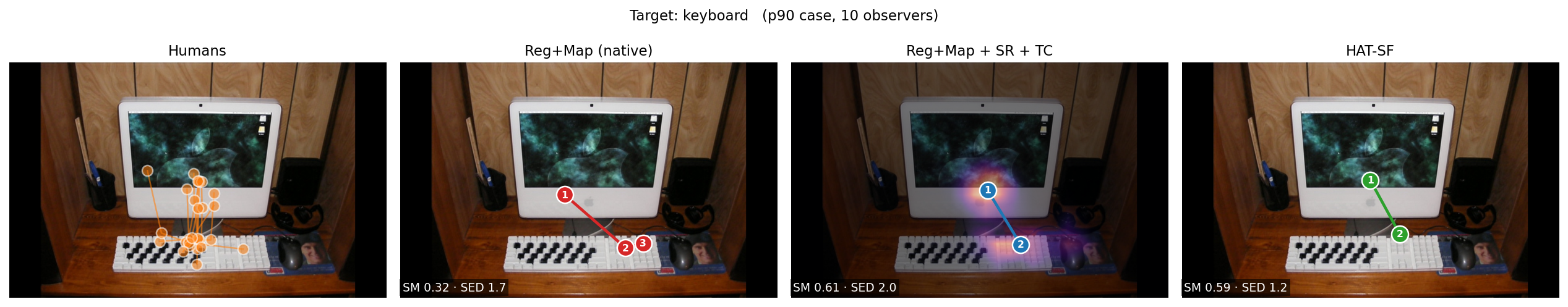}
\caption{Keyboard: group near the 90th-percentile gain. The two outputs differ
in both location and length.}
\label{fig:supp_qual_case4}
\end{figure*}

\begin{figure*}[t]
\centering
\includegraphics[width=\linewidth]{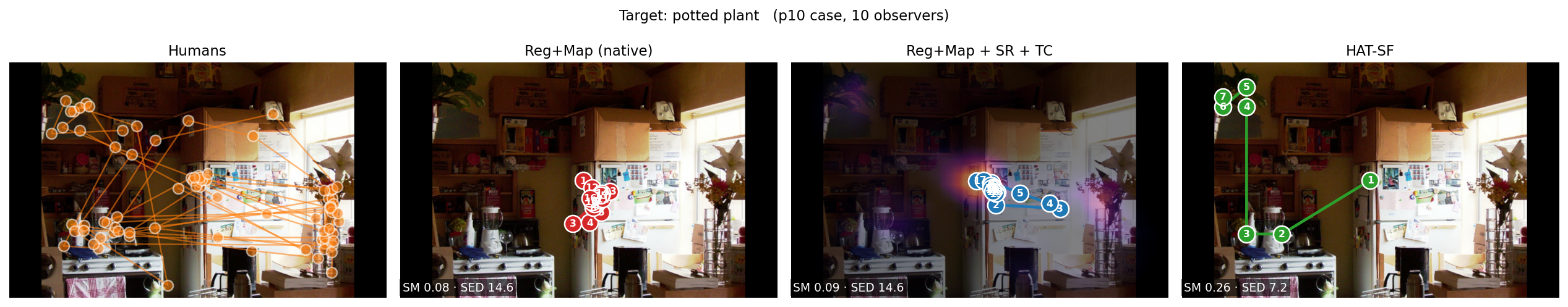}
\caption{Potted plant: group near the 10th-percentile gain. Both outputs
reach the 16-slot limit. An example of the failure mode in Section~4.5 of the main paper.}
\label{fig:supp_qual_case5}
\end{figure*}

Figures~\ref{fig:supp_qual_case1} to~\ref{fig:supp_qual_case5} show, from left to
right, the human scanpaths, native Reg+Map, Reg+Map + SR + TC, and HAT-SF
(seed~3). The model panels report group-mean ScanMatch and SED. The third
panel overlays the per-cell maximum of the unsharpened spatial probability
maps across the slots retained after calibration. This summary is for
visualization and does not identify individual slot peaks.

\paragraph{Observations.}
The toilet case (Figure~\ref{fig:supp_qual_case3}) illustrates the pattern described in the main paper. Native fixations land on the nearby background. In contrast, the second and third fixations of SR + TC fall on the target. Both outputs contain three fixations (ScanMatch \(0.32 \rightarrow 0.61\), SED \(2.1 \rightarrow 1.0\)). For the keyboard case (Figure~\ref{fig:supp_qual_case4}), SR + TC shortens the sequence from three fixations to two. Consequently, ScanMatch improves, but SED worsens. The potted plant case (Figure~\ref{fig:supp_qual_case5}) exemplifies the failure mode described in Section~4.5 of the main paper. Human search is spread over both sides of the image. However, the plotted spatial response and both model outputs concentrate in the middle. Reading the map cannot correct location errors inherent in the map itself. Furthermore, both outputs reach the 16-slot limit, yielding a minimal ScanMatch gain of only \(0.009\).

\subsection{Source and scope of the 92\% figure}
\label{sec:supp_qual_share}

On seed 3, 300 of the 326 groups improve (92.0\%), 25 decline, and one remains unchanged. The denominator counts task-image groups rather than physical images or pooled seeds. Seed 4 yields 299/326 improvements (91.7\%), with 24 declines and 3 unchanged cases. Seed 5 yields 311/326 improvements (95.4\%), with 14 declines and 1 unchanged case.


\end{document}